\documentclass[journal]{IEEEtran}

\IEEEoverridecommandlockouts 
\let\labelindent\relax  %

\makeatletter
\def\cl@chapter{\@elt {theorem}}
\makeatother

\usepackage{times}

\usepackage{textcomp}
\usepackage{lipsum}
 \usepackage{flushend}

\usepackage[utf8]{inputenc}

\usepackage{url}
\usepackage{ifpdf}
\usepackage{cite}

\ifpdf
\usepackage[table]{xcolor}
\usepackage{graphics}
\usepackage[pdftex]{graphicx}
\DeclareGraphicsExtensions{.pdf,.PDF,.png,.PNG,.jpg,.JPG}
\usepackage{hyperref}
\hypersetup{
  colorlinks,%
  citecolor=black,%
  filecolor=black,%
  linkcolor=black,%
  urlcolor=black
}
\else
\usepackage{graphicx}
\DeclareGraphicsExtensions{.eps,.EPS,.ps,.PS,}
\fi

\usepackage{amsmath}
\usepackage{amssymb} 
\usepackage{stmaryrd}           
\usepackage{mathptmx}
\usepackage[mathscr]{euscript}
\usepackage{bm}
\usepackage{nicefrac}
\usepackage[hang,flushmargin]{footmisc}

\DeclareMathAlphabet{\mathcal}{OMS}{cmsy}{m}{n} %

\newcommand\imCMsym[4][\mathord]{%
  \DeclareFontFamily{U} {#2}{}
  \DeclareFontShape{U}{#2}{m}{n}{
    <-6> #25
    <6-7> #26
    <7-8> #27
    <8-9> #28
    <9-10> #29
    <10-12> #210
    <12-> #212}{}
  \DeclareSymbolFont{CM#2} {U} {#2}{m}{n}
  \DeclareMathSymbol{#4}{#1}{CM#2}{#3}
}

\imCMsym{cmmi}{124}{\CMjmath}

\usepackage[ruled,vlined]{algorithm2e}

\usepackage{tabularx}
\usepackage{balance}
\usepackage{booktabs}
\usepackage{multirow}
\usepackage{multicol}
\usepackage{array}
\usepackage{adjustbox}
\usepackage[british]{babel}
\usepackage{dcolumn}
\usepackage{colortbl}
\usepackage{threeparttable}
\usepackage{enumitem}

\usepackage[binary-units=true]{siunitx}

\makeatletter
\if@twocolumn
  \renewcommand\normalsize{%
   \@setfontsize\normalsize\@xpt{12.5pt}%
   \abovedisplayskip=3 mm plus6pt minus 4pt
   \belowdisplayskip=3 mm plus6pt minus 4pt
   \abovedisplayshortskip=0.0 mm plus6pt
   \belowdisplayshortskip=2 mm plus4pt minus 4pt
   \let\@listi\@listI}%

  \renewcommand\small{%
   \@setfontsize\small{8.5pt}\@xpt
   \abovedisplayskip 8.5\p@ \@plus3\p@ \@minus4\p@
   \abovedisplayshortskip \z@ \@plus2\p@
   \belowdisplayshortskip 4\p@ \@plus2\p@ \@minus2\p@
   \def\@listi{\leftmargin\leftmargini
               \parsep 0\p@ \@plus1\p@ \@minus\p@
               \topsep 4\p@ \@plus2\p@ \@minus4\p@
               \itemsep0\p@}%
   \belowdisplayskip \abovedisplayskip}

\else
  \if@smallext
   \renewcommand\normalsize{%
   \@setfontsize\normalsize\@xpt\@xiipt
   \abovedisplayskip=3 mm plus6pt minus 4pt
   \belowdisplayskip=3 mm plus6pt minus 4pt
   \abovedisplayshortskip=0.0 mm plus6pt
   \belowdisplayshortskip=2 mm plus4pt minus 4pt
   \let\@listi\@listI}%

  \renewcommand\small{%
   \@setfontsize\small\@viiipt{9.5pt}%
   \abovedisplayskip 8.5\p@ \@plus3\p@ \@minus4\p@
   \abovedisplayshortskip \z@ \@plus2\p@
   \belowdisplayshortskip 4\p@ \@plus2\p@ \@minus2\p@
   \def\@listi{\leftmargin\leftmargini
               \parsep 0\p@ \@plus1\p@ \@minus\p@
               \topsep 4\p@ \@plus2\p@ \@minus4\p@
               \itemsep0\p@}%
   \belowdisplayskip \abovedisplayskip}
 \else
  \renewcommand\normalsize{%
   \@setfontsize\normalsize{9.5pt}{11.5pt}%
   \abovedisplayskip=3 mm plus6pt minus 4pt
   \belowdisplayskip=3 mm plus6pt minus 4pt
   \abovedisplayshortskip=0.0 mm plus6pt
   \belowdisplayshortskip=2 mm plus4pt minus 4pt
   \let\@listi\@listI}%

  \renewcommand\small{%
   \@setfontsize\small\@viiipt{9.25pt}%
   \abovedisplayskip 8.5\p@ \@plus3\p@ \@minus4\p@
   \abovedisplayshortskip \z@ \@plus2\p@
   \belowdisplayshortskip 4\p@ \@plus2\p@ \@minus2\p@
   \def\@listi{\leftmargin\leftmargini
               \parsep 0\p@ \@plus1\p@ \@minus\p@
               \topsep 4\p@ \@plus2\p@ \@minus4\p@
               \itemsep0\p@}%
   \belowdisplayskip \abovedisplayskip}
  \fi
\let\footnotesize\small
\makeatother

\usepackage{microtype}

\usepackage{scrtime}

\usepackage{etoolbox}

\newcolumntype{P}[1]{>{\centering\arraybackslash}p{#1}}

\usepackage[capitalise,noabbrev]{cleveref}
\usepackage{subcaption}
\usepackage{bbm}
\usepackage{pifont}
\usepackage{tikz}

\makeatletter
\newcommand{\algorithmfootnote}[2][\footnotesize]{%
  \let\old@algocf@finish\@algocf@finish%
  \def\@algocf@finish{\old@algocf@finish%
    \leavevmode\rlap{\begin{minipage}{\linewidth}
    #1#2
    \end{minipage}}%
  }%
}
\makeatother

\usepackage[nolist,nohyperlinks]{acronym}
\newacro{CNN}{Convolutional Neural Network}
\newacro{DNN}{Deep Neural Network}
\newacro{GPS}{Global Positioning System}
\newacro{GNSS}{Global Navigation Satellite System}
\newacro{NLOS}{non-line-of-sight}
\newacro{ADAS}{Advanced Driver Assistance Systems}
\newacro{LIDAR}[LiDAR]{Light Detection And Ranging}
\newacro{HD map}{High Definition map}
\newacro{EV}{Embedding Vector}
\newacro{SLAM}{Simultaneouos Localization And Mapping}
\newacro{MLP}{MultiLayer Perceptron}
\newacro{IMU}{Inertial Measurement Unit}
\newacro{ML}{Machine Learning}
\newacro{SfM}{Structure from Motion}
\newacro{PnP}{Perspective-n-Points}
\newacro{ASPP}{Atrous Spatial Pyramid Pooling}
\newacro{RANSAC}{RANdom SAmple Consensus}
\newacro{CV}{Computer Vision}
\newacro{GRU}{Gated Recurrent Unit}
\newacro{MAE}{Mean Absolute Error}
\newacro{FoV}{Field of View}

\def\etal{\emph{et al. }}
\def\eg{\emph{e.g., }}
\def\ie{\emph{i.e., }}

\def\cmrnet2{CMRNext}

\usepackage{tikz}
\usepackage{textcomp}
\usepackage{lipsum}

\newcounter{nodecount}
\newcommand\tabnode[1]{\addtocounter{nodecount}{1} \tikz \node (\arabic{nodecount}) {#1};}
\tikzstyle{every picture}+=[remember picture,baseline]
\tikzstyle{every node}+=[inner sep=0pt,anchor=base,
minimum width=1cm,align=center,text depth=.25ex,outer sep=1.5pt]
\tikzstyle{every path}+=[thick, rounded corners]

\begin{document}

\title{\LARGE \bf
\cmrnet2: Camera to LiDAR Matching in the Wild \\ for Localization and Extrinsic Calibration
}

\author{Daniele Cattaneo and Abhinav Valada%
\thanks{© 2025 IEEE. Personal use of this material is permitted. Permission from IEEE must be obtained for all other uses, in any current or future media, including reprinting/republishing this material for advertising or promotional purposes, creating new collective works, for resale or redistribution to servers or lists, or reuse of any copyrighted component of this work in other works.}
\thanks{Department of Computer Science, University of Freiburg, Germany.}%
\thanks{This work was funded by the German Research Foundation (DFG) Emmy Noether Program grant number 468878300.}%
}

\markboth{T\MakeLowercase{his paper will appear in:} IEEE Transactions on Robotics}%
{Cattaneo and Valada, {CMRNext}: Camera to LiDAR Matching in the Wild for Localization and Extrinsic Calibration}

\maketitle

\begin{abstract}
LiDARs are widely used for mapping and localization in dynamic environments. However, their high cost limits their widespread adoption. On the other hand, monocular localization in LiDAR maps using inexpensive cameras is a cost-effective alternative for large-scale deployment. Nevertheless, most existing approaches struggle to generalize to new sensor setups and environments, requiring retraining or fine-tuning. In this paper, we present CMRNext, a novel approach for camera-LIDAR matching that is independent of sensor-specific parameters, generalizable, and can be used in the wild for monocular localization in LiDAR maps and camera-LiDAR extrinsic calibration. CMRNext exploits recent advances in deep neural networks for matching cross-modal data and standard geometric techniques for robust pose estimation. We reformulate the point-pixel matching problem as an optical flow estimation problem and solve the Perspective-n-Point problem based on the resulting correspondences to find the relative pose between the camera and the LiDAR point cloud. We extensively evaluate CMRNext on six different robotic platforms, including three publicly available datasets and three in-house robots. Our experimental evaluations demonstrate that CMRNext outperforms existing approaches on both tasks and effectively generalizes to previously unseen environments and sensor setups in a zero-shot manner. We make the code and pre-trained models publicly available at \url{http://cmrnext.cs.uni-freiburg.de}.
\end{abstract}

\section{Introduction}\label{sec:introduction}

Localization is an essential task for any autonomous robot, as it is a precursor for subsequent safety-critical tasks such as path planning and navigation. The accuracy of \acp{GNSS} is not sufficient for most of these tasks, and therefore, sensors such as \ac{LIDAR} and cameras are employed to estimate the pose of the robot. While \ac{LIDAR}-based localization methods~\cite{cattaneo2022tro,xu2022fast} achieve the best performance in terms of accuracy and robustness, the comparatively high price of \ac{LIDAR} sensors hinders their widespread adoption. On the other hand, cameras are inexpensive and already widely available in consumer vehicles for \ac{ADAS} applications. However, the performance of camera-based localization methods~\cite{campos2021orb,Vodisch_2023_CVPR,vodisch2022continual,ballardini2021vehicle,Ballardini_ICRA_2019} is still not comparable to \ac{LIDAR}-based localization techniques, especially when the lighting and weather conditions during deployment differ from those recorded while mapping.

\begin{figure}
  \centering
      \includegraphics[width=1\linewidth]{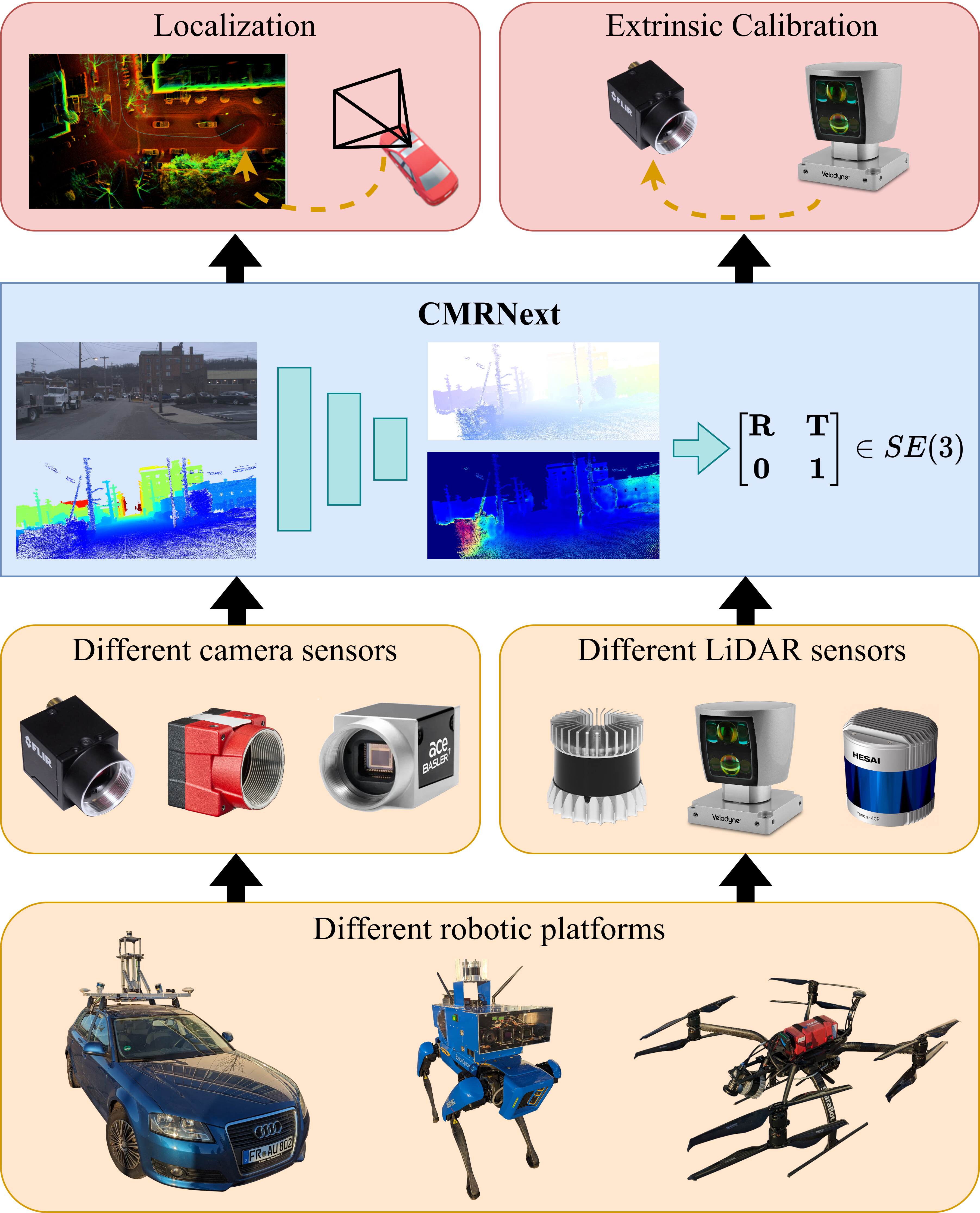}
      \caption{Our proposed \cmrnet2{} estimates the 6-DoF transformation between camera images and LiDAR scans in the wild. It can be readily employed for monocular localization in LiDAR maps and camera-LiDAR extrinsic calibration.}
    \label{fig:teaser}
\end{figure}

Consequently, monocular localization in \ac{LIDAR} maps has gained considerable attention from both academia and industry in recent years, as it combines the high accuracy and robustness of \ac{LIDAR}-based localization while only requiring a camera sensor during deployment. The increasing availability of \ac{LIDAR} maps from mapping companies such as HERE and TomTom has further contributed to the popularity of this approach. Once a \ac{LIDAR} map is available, either obtained from a mapping company or generated using recent \ac{LIDAR} \ac{SLAM} algorithms~\cite{arce2023padloc, xu2022fast, greve2023collaborative}, a vehicle can be localized within this map using a single camera. Considering the low cost of cameras, compared to \ac{LIDAR} sensors, this approach is particularly appealing for large-scale deployment, where the cost of the final system is a key factor.
While traditionally this task has been addressed by first generating a 3D representation from the camera image and then matching it against the \ac{LIDAR} map~\cite{Caselitz_2016}, the recent advancement of \ac{DNN}-based methods for multimodal representation learning~\cite{guo2019deep,nayak2024ralf,ballardini2017,mohan2024progressive} has enabled the matching to be performed directly in the feature space.

Most of the existing \ac{DNN}-based approaches for monocular localization in \ac{LIDAR} maps, however, follow the end-to-end learning paradigm, where the network directly predicts the metric pose of the camera given an input image and a representation of the \ac{LIDAR} map. While this paradigm has the advantage of being simple to train and usually achieves good performance when evaluated on images from the same camera used during training, it has the major drawback of being dependent on the specific camera's intrinsic parameters. Consequently, the network has to be retrained for every new camera and cannot generalize to different sensor setups. This is even worse when these approaches are applied to the task of extrinsic calibration, where the goal is to estimate the relative pose between the camera and the \ac{LIDAR} sensor. In this case, end-to-end approaches trained on a single camera-LiDAR pair typically tend to overfit to the specific intrinsic and extrinsic parameters of the dataset and fail to generalize even when evaluated on a different camera in the same dataset. For example, a network trained in an end-to-end manner on the left camera of the KITTI dataset~\cite{Geiger2012CVPR} and evaluated on the right camera will still predict the pose of the left camera. This shows that these networks just learn the specific extrinsic parameters of the camera and thus cannot be used for extrinsic calibration in any real-world application, suggesting that a combination of learning based-methods and geometric techniques should be preferred~\cite{petek2024mdpcalib}.

In this paper, we propose \cmrnet2{}, a novel \ac{DNN}-based approach for camera-\ac{LIDAR} matching that is independent of any sensor-specific parameter. Our approach can be employed for monocular localization in \ac{LIDAR}-maps, as well as camera-\ac{LIDAR} extrinsic calibration. Moreover, \cmrnet2{} is able to generalize to different environments and different sensor setups, including different intrinsic and extrinsic parameters. This is achieved by decoupling the matching step from the metric pose estimation step. In particular, we first train a \ac{CNN} to predict dense matches between image pixels and 3D points in the \ac{LIDAR} point clouds, together with their respective uncertainties. These matches are then used to estimate the pose of the camera using a traditional \ac{PnP} algorithm. Hence, the network only reasons at the pixel level and is thus independent of any metric information.
Although the intrinsic parameters of the camera are still required to be known, they are provided as input to the \ac{PnP} algorithm, and therefore, the network can be used with different cameras without needing any retraining.
We perform comprehensive experimental evaluations that demonstrate that \cmrnet2{} outperforms existing methods and achieves state-of-the-art performance on both monocular localization in \ac{LIDAR}-maps and extrinsic calibration on three publicly available datasets. Moreover, we evaluate our approach on three in-house robotic platforms: a self-driving perception car, a quadruped robot, and a quadcopter, and show that \cmrnet2{} can generalize to very different robotic platforms and sensor setups without any retraining or fine-tuning.

The main contributions of this work are as follows:
\begin{itemize}[topsep=0pt]
  \item We propose \cmrnet2{}, a novel two-step approach for monocular localization in \ac{LIDAR}-maps and camera-LiDAR extrinsic calibration, which is independent of any sensor-specific parameter.
  \item We propose to estimate the uncertainty of camera-\ac{LIDAR} matches as additional output for usage in downstream tasks.
  \item We demonstrate that \cmrnet2{} achieves state-of-the-art performance for both tasks on multiple real-world publicly available datasets, and we perform extensive ablation studies to analyze the contribution of each component in our method.
  \item We demonstrate the generalization ability of our approach to previously unseen environments and sensor setups on three different in-house robotic platforms.
  \item We release the code and pre-trained models of our approach to facilitate future research at \url{http://cmrnext.cs.uni-freiburg.de}.
\end{itemize}

\section{Related Works}\label{sec:related-works}

In this section, we discuss related work on monocular localization in LiDAR-maps, followed by an overview of existing methods to solve the extrinsic calibration problem between the camera and LiDAR sensors.

\subsection*{Monocular Localization in LiDAR-maps}\label{sec:related-works-localization}
Due to the intrinsically different nature of cameras and LiDAR sensors, standard approaches for monocular localization in LiDAR-maps first convert one modality into the other, such that the matching can be performed in the same space. For example, Caselitz \etal~\cite{Caselitz_2016} propose to generate a 3D point cloud from the image stream using visual odometry and local bundle adjustment, and then continuously match the reconstructed points with the LiDAR map to track the camera position. subsequent works further improved this approach by using an updatable scale estimator to reduce the scale-drift problem~\cite{sun2019scale} or by extending it to multiple camera setups~\cite{yabuuchi2021visual}. Conversely, the match can be performed in the visual space, by projecting the LiDAR map into the image plane and comparing the intensities produced by the LiDAR sensor with a grayscale image using normalized mutual information~\cite{Wolcott_2014, pandey2012automatic} or normalized information distance~\cite{Pascoe_2015_ICCV_Workshops}. The former family of approaches, however, is based on a point cloud generated from monocular visual odometry, which is noisy and requires a continuous estimation of the scale factor. The latter family of approaches, instead, is based on the assumption that LiDAR intensities and grayscale images are visually similar, which is not always true, especially considering that intensities vary greatly across different LiDAR sensor models and manufacturers~\cite{carballo2020comparison}.

Recently, \ac{DNN}-based approaches have been proposed to directly match the two modalities in the feature space, taking advantage of recent advancements in robust feature learning. Prior works on DNN-based metric monocular localization~\cite{8458420} implicitly learn the map during the training stage and thus can only be employed in the same area used for training. Our previous work CMRNet~\cite{Cattaneo_2019} was the first DNN-based method to use the LiDAR map as additional input, such that the network learns to match the camera image with the LiDAR map. Thereby, CMRNet was able to localize a camera image in any environment for which a LiDAR map is available without any retraining or fine-tuning. Based on CMRNet, HyperMap~\cite{chang2021hypermap} proposes a method to compress the LiDAR map without sacrificing localization performances, while POET~\cite{miao2023poses} replaced the fully connected layers of CMRNet with a Transformer-based pose regressor. These approaches, however, while being able to generalize to different environments, are still dependent on the specific camera used during training.
Concurrently, a parallel branch of research proposed to match keypoints across images and LiDAR maps. 2D3D-MatchNet~\cite{Mengdan2019} was the first proposed method in this category, and it leveraged DNN-based feature extractors and metric learning. Few recent works have since been proposed~\cite{Wang_2021_ICCV,Li_2023_ICCV}, and although they are intrinsically camera and map-agnostic, they either focus on indoor environments or achieve performance that are not suitable for mobile robotics applications when evaluated in outdoor environments.

More recently, we proposed CMRNet++~\cite{cattaneo2020cmrnet}, an extension of CMRNet that is independent of the camera's intrinsic parameters. CMRNet++ for the first time demonstrated that a \ac{DNN}-based method for monocular localization in LiDAR map can generalize to different sensor setups and different environments in a zero-shot setting. Inspired by CMRNet++, I2D-Loc further improves the performance by densifying the sparse depth map, and by replacing the Perspective-n-Point (PnP) algorithm with the differentiable version BPnP~\cite{Chen_2020_CVPR} during training.
In this work, we redesigned our previous approach CMRNet++~\cite{cattaneo2020cmrnet}, and propose \cmrnet2{}, with a novel architecture, faster runtime, and uncertainty estimation. Through extensive ablation studies, hyperparameter optimization, and experimental evaluations on six different robotic platforms, we demonstrate that \cmrnet2{} achieves state-of-the-art performance.

\subsection*{LiDAR-Camera Extrinsic Calibration}\label{sec:related-works-calibration}

The extrinsic calibration between LiDAR and camera sensors is typically performed using target-based methods, where an ad-hoc calibration board is placed in the environment, and the relative pose between the two sensors is estimated by detecting the board in both modalities. For example, common target boards used for this task are one or multiple checkerboards on planar boards~\cite{geiger2012automatic}, ordinary boxes~\cite{Pusztai_2017_ICCV} or a planar board with circular holes~\cite{velas2014calibration}. Although target-based methods are widely used, they require a data recording specifically for this purpose, which should be repeated periodically to account for possible changes in the extrinsic parameters due to vibrations, temperature variations, and manual repositioning of the sensors. Early attempts at targetless extrinsic calibration used visual similarity measures, such as mutual information~\cite{pandey2012automatic}, contours~\cite{5457439}, or edges~\cite{levinsonautomatic}, to estimate the relative pose between the sensors. These methods, however, are sensitive to appearance changes, such as lighting and weather conditions, and thus require a careful tuning of the parameters.

In recent years, \ac{DNN}-based approaches have been proposed to estimate the extrinsic parameters between LiDAR and camera sensors. RegNet~\cite{Schneider_2017} was a pioneering method in this field, and it used a \ac{CNN} that processed a camera image and a LiDAR projection to directly estimate the relative rotation and translation between the two sensors. Many subsequent works follow a similar approach and propose different network architectures to improve the performance~\cite{8593693,Yuan2020rggnet}. More recently, taking inspiration from CMRNet, LCCNet~\cite{lv2021lccnet} proposed to leverage the cost volume usually employed in optical flow networks~\cite{Sun_2018_CVPR} instead of the simple concatenation to aggregate features from the two modalities. These approaches, however, are intrinsically dependent on the specific camera-LiDAR pair used during training and thus cannot be used for extrinsic calibration in any other sensor setup. In fact, all the aforementioned methods are evaluated only on the same camera-LiDAR pair used during training, which is, in our opinion, an improper evaluation protocol for the extrinsic calibration task. Similarly, RGKCNet~\cite{ye2021keypoint} associates only keypoints rather than the entire point cloud, relying on a dedicated keypoint extraction network. However, this requires a sufficient number of evenly distributed keypoints in the scene to ensure parameter observability. An alternative approach is to align features produced by a deep network, followed by a subsequent optimization step that makes the method generalizable across different sensor setups and scenarios~\cite{fu2023batch}.
Following CMRNet++ for monocular localization, a few camera-agnostic approaches based on flow estimation between camera and LiDAR projection have been proposed for extrinsic calibration~\cite{lv2021cfnet,jing2022dxq}. Although these approaches are independent of the camera's intrinsic parameters, when evaluated on a different robotic platform, they still need to be fine-tuned on the new camera-LiDAR pair. While fine-tuning is computationally inexpensive compared to the full training required by camera-dependent approaches, requiring ground truth extrinsic calibration to predict the same extrinsic parameters is not a realistic assumption for real-world applications.
In this paper, we additionally extend \cmrnet2{} for extrinsic calibration, and we demonstrate that our approach can generalize to very different sensor setups in a zero-shot manner without any retraining or fine-tuning.

\begin{figure*}
  \centering
      \includegraphics[width=.99\textwidth]{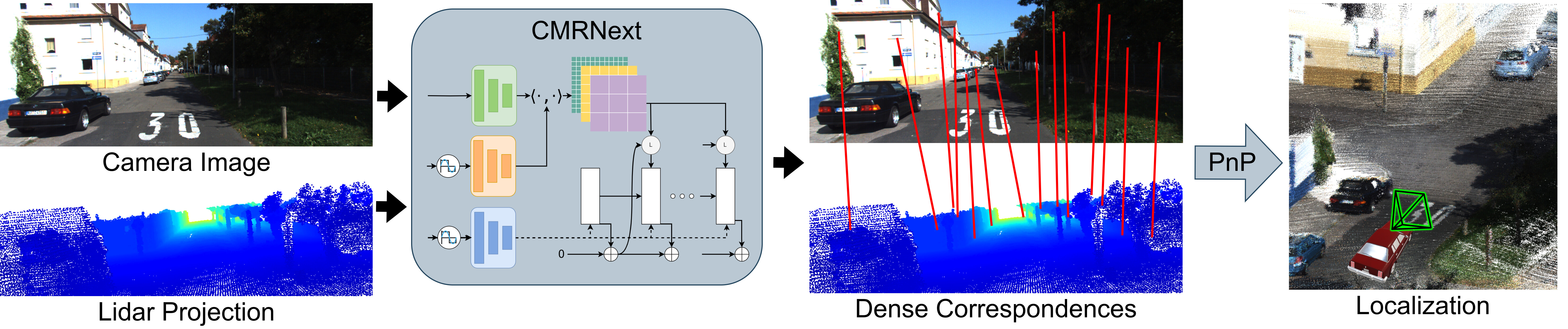}
      \caption{Overview of the proposed approach: the input camera image and LiDAR-image are fed to \cmrnet2, which predicts dense correspondences between the two inputs. The predicted matches are used to localize the camera within the LiDAR-map by solving the Perspective-n-Point problem.}
      \label{fig:pipeline}
      \vspace{0.2cm}
  \end{figure*}

\section{Technical Approach}\label{sec:proposed-approach}

In this section, we describe our proposed \cmrnet2 for generalizable camera-LiDAR matching, which can be used for tasks such as monocular localization in LiDAR-maps and extrinsic camera-LiDAR calibration. An overview of the proposed method is depicted in \cref{fig:pipeline}, and is composed of two main stages: pixel to 3D point matching, followed by 6-DoF pose regression. In the first step, the \ac{CNN} only focuses on matching at the pixel level instead of metric basis, which makes the network independent of the intrinsic parameters of the camera. These parameters are instead employed in the second step, where traditional computer vision methods are exploited to estimate the pose of the camera, given the matches from the first step. Consequently, after training, \cmrnet2 can also be used with different cameras and maps from those used while training. 
Although the method is agnostic to the camera's intrinsic parameters, these parameters still need to be provided as input, and we assume them to be accurate.
We first detail the aforementioned stages, followed by the network architecture, the loss function used to train \cmrnet2, the generation of the ground truth, and the iterative refinement approach employed to improve the predictions.

\subsection{Matching Step}
\label{sec:matching}
\begin{figure}
  \centering
  \footnotesize
  \setlength{\tabcolsep}{0.1cm}
      {\renewcommand{\arraystretch}{1}
        \begin{tabular}{p{8.6cm}}
      \includegraphics[width=1\linewidth]{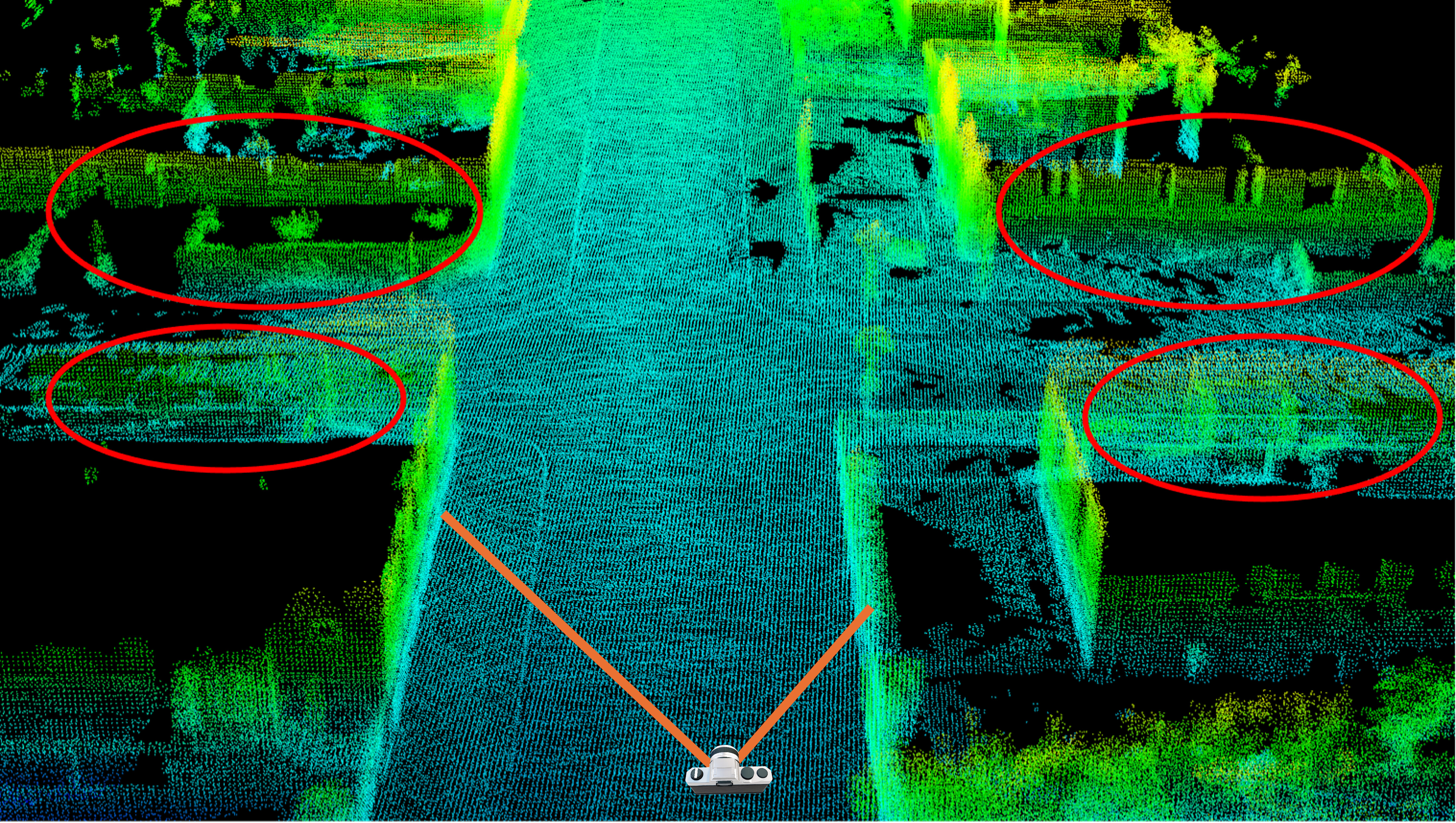} \\
        \multicolumn{1}{c}{LiDAR-map} \\
      \includegraphics[width=1\linewidth]{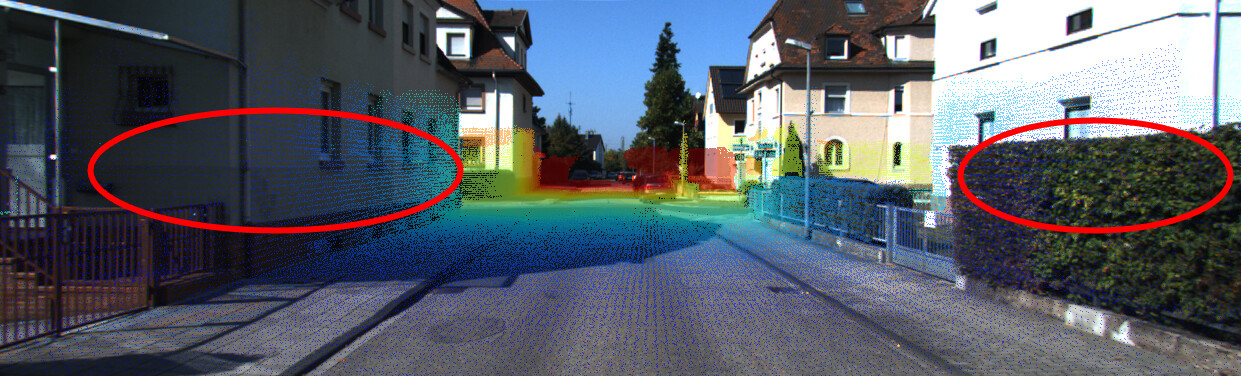} \\
        \multicolumn{1}{c}{With Occlusion Filter} \\
      \includegraphics[width=1\linewidth]{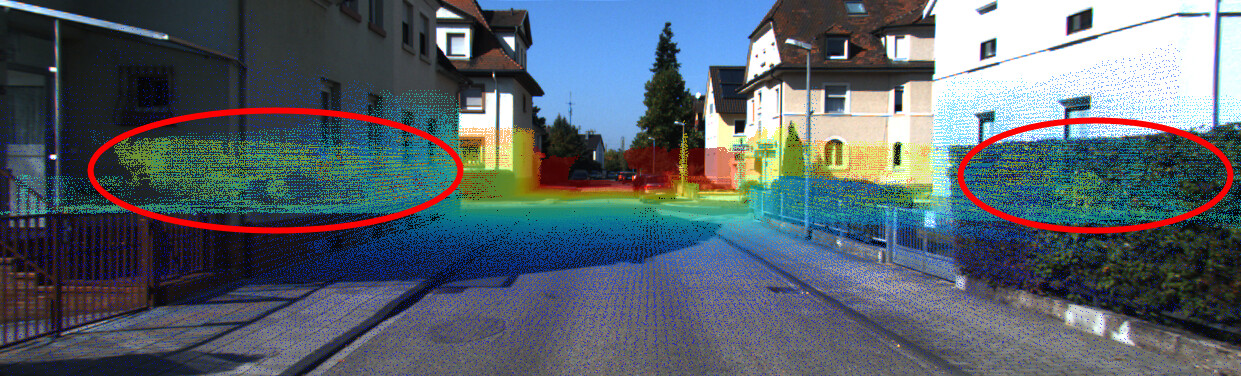} \\
      \multicolumn{1}{c}{Without Occlusion Filter} \\
      \caption{LiDAR-map (top) projected in the ground truth position with (middle) and without (bottom) the occlusion filter. Most of the points that are occluded by closer objects, highlighted in red, are correctly removed by the occlusion filter.}
    \label{fig:occlusion}
    \end{tabular}}
    \vspace{-1cm}
\end{figure}

The main intuition behind the matching step is that the projection of the LiDAR-map in the correct pose should align with the camera image, see~\cref{fig:occlusion} (middle). Therefore, given a camera image and a (misaligned) LiDAR projection, we can train a \ac{CNN} to align the two inputs by predicting pixel-level displacements.
Specifically, we generate a synthesized depth image, which we refer to as a LiDAR-image, by projecting the map into a virtual image plane placed at an initial pose estimate $H_{map}^{init}$.
This initial pose can be obtained, for example, from a \ac{GNSS} or a cross-modal place recognition method~\cite{cattaneo2020global}. Although \cmrnet2 is independent of the camera intrinsic parameters, i.e. the network's weights are not tied to a specific camera, the intrinsics are still required as input to our method, and they are used to perform the projection. \Cref{fig:pipeline}(bottom left) shows an example of a LiDAR-image. %

Due to the sparse nature of LiDAR point clouds, the projection of the map into the virtual image plane can include points that are occluded by other objects. For example, a point that lays behind a building will still be projected in the LiDAR-image if no closer points are present along the same projection line.
Therefore, to deal with occlusions, we employ a z-buffer technique followed by an occlusion estimation filter
In particular, we reimplemented the occlusion filter proposed in~\cite{pintus2011real}, which is based on the insight that a point is considered occluded if some other points obstruct the accessibility of that point. They define the accessibility of a point $P_i$ as the aperture of the maximum cone that connects the point to the pinhole of the camera without intersecting any other points. More formally, for every point $P_j$ which is projected in a neighborhood of size $K_{occ} \times K_{occ}$ centered around the projection of $P_i$, we define the normalized vector between the two points as $\vec{c}_{ij} = \frac{P_j - P_i}{\lVert P_j - P_i \rVert}$. Then, we compute the angle between this vector and the normalized vector $\vec{v}_i$ from $P_i$ to the camera pinhole as $\alpha_{ij} = \vec{v}_i \cdot \vec{c}_{ij}$. We split the $K_{occ} \times K_{occ}$ occlusion kernel into four sectors, and for each sector, we calculate the maximum $\alpha_{ij}$ value. If the sum of the maximum values of the four sectors is greater than a threshold $Thr$, then the point $P_i$ is considered as visible, otherwise it is considered as occluded. Since the occlusion filter is performed in the image space, it can be efficiently implemented as a GPU kernel. We refer the reader to~\cite{pintus2011real} for more details. The effect of the occlusion filter is shown in~\cref{fig:occlusion}.
Once the inputs to the network (RGB and LiDAR images) have been obtained, for every 3D point in the LiDAR-image, \cmrnet2{} estimates which pixel of the RGB image depicts the same world point. %

The architecture of \cmrnet2 is based on RAFT~\cite{Sun_2018_CVPR}, a state-of-the-art network for optical flow estimation between two consecutive RGB frames. More details regarding the network architecture are reported in \cref{sec:network}. The output of \cmrnet2 is a dense feature map, which consists of four channels that represent, for every pixel in the LiDAR-image, the displacement ($u$, $v$) of the pixel in the RGB image from the same world point and the uncertainties of this match ($\sigma_u$, $\sigma_v$). A visual representation of this pixel displacement is depicted in \Cref{fig:pipeline}. The predicted uncertainties can be used for any downstream task, such as active \ac{SLAM} or map change detection.

In order to train \cmrnet2, we first need to generate the ground truth pixel displacement $\Delta P = (\Delta_u, \Delta_v)$ of the LiDAR-image with respect to the RGB image. To accomplish this, we first compute the coordinates of the map's points in the initial reference frame $H_{map}^{init}$ using \Cref{eq:frame_init} and the pixel position $(\mathbf{u}^{init},\mathbf{v}^{init})$ of their projection in the LiDAR-image exploiting the intrinsic matrix $K$ of the camera using \Cref{eq:proj_init}.
\begin{align}
    [
    \mathbf{x}^{init} \
    \mathbf{y}^{init} \
    \mathbf{z}^{init} \
    \mathbf{1}
    ]^\intercal
     &=  H_{map}^{init} \cdot 
    [ 
    \mathbf{x}^{map} \
    \mathbf{y}^{map} \
    \mathbf{z}^{map} \
    \mathbf{1}
    ]^\intercal,    \label{eq:frame_init}\\
    [ 
    \mathbf{u}^{init} \
    \mathbf{v}^{init} \
    \mathbf{1}
    ]^\intercal
     &= K \cdot 
    [ 
    \mathbf{x}^{init} \
    \mathbf{y}^{init} \
    \mathbf{z}^{init} \
    \mathbf{1}
    ]^\intercal.  \label{eq:proj_init}
\end{align}

We keep track of indices of valid points in an array $\mathbf{VI}$. This is done by excluding indices of points whose projection lies behind or outside the image plane, as well as points marked occluded by the occlusion estimation filter. Subsequently, we generate the sparse LiDAR-image $\mathcal{D}$
\begin{gather}
  \mathcal{D}_{\mathbf{u}^{init}_i, \mathbf{v}^{init}_i} = \mathbf{z}^{init}_i,\  i \in \mathbf{VI}. \label{eq:lidar_img}
\end{gather}
Then, we project the points of the map into a virtual image plane placed at the ground truth pose $H_{map}^{GT}$. We then store the pixels' position of these projections as
\begin{gather}
[ 
    \mathbf{u}^{GT} \
    \mathbf{v}^{GT} \
    \mathbf{1}
    ]^\intercal
    = K \cdot H_{map}^{GT} \cdot 
    [ 
    \mathbf{x}^{map} \
    \mathbf{y}^{map} \
    \mathbf{z}^{map} \
    \mathbf{1}
    ]^\intercal. \label{eq:proj_GT}
\end{gather}

Finally, we compute the displacement ground truths $\Delta P$ by comparing the projections in the two image planes as
\begin{equation}
\label{eq:GT}
\Delta P_{\mathbf{u}^{init}_i, \mathbf{v}^{init}_i} = \left[ \mathbf{u}^{GT}_i-\mathbf{u}^{init}_i, \mathbf{v}^{GT}_i-\mathbf{v}^{init}_i \right], \ i \in \mathbf{VI}.
\end{equation}

For every pixel $[u, v]$ without an associated 3D point, we set $\mathcal{D}_{u, v}=0$ and $\Delta P_{u, v}=[0,0]$. Moreover, we generate a mask of valid pixels as follows
\begin{equation}
\label{eq:mask}
  mask_{u,v} = 
  \begin{cases}
    1, & \text{if } \mathcal{D}_{u,v}> 0 \\
    0, & \text{otherwise.}
  \end{cases} 
\end{equation}

\subsection{Localization Step}
\label{sec:localization}
Once \cmrnet2 has been trained, during inference, we have the map, \ie a set of 3D points $\mathbf{Q}$ whose coordinates are known, altogether with their projection in the initial pose $[\mathbf{u}^{init}_i, \mathbf{v}^{init}_i]$, and a set $\mathbf{p}$ of matching pixels in the RGB image that is computed using the displacements $\widehat{\Delta P}$ predicted by the \ac{CNN} given as
\begin{align}
    \mathbf{Q}_i &= [\mathbf{x}^{map}_i, \mathbf{y}^{map}_i, \mathbf{z}^{map}_i], \ i \in \mathbf{VI}, \\
    \mathbf{p}_i &= [\mathbf{u}^{init}_i, \mathbf{v}^{init}_i] + \widehat{\Delta P}_{\mathbf{u}^{init}_i,\mathbf{v}^{init}_i} , \ i \in \mathbf{VI}.
\end{align}
Although \cmrnet2{} predicts dense correspondences, even for pixels without an associated 3D LiDAR point, we first filter out invalid pixels by using the $mask$ defined in~\cref{eq:mask}

Estimating the pose of the camera given a set of 2D-3D correspondences and the camera intrinsics $K$ is known as the \ac{PnP} problem.
More formally, \ac{PnP} aims at estimating the pose of the camera $H^*$ that minimizes the reprojection errors as
\begin{gather}
  H^* = \arg\min_{H} \sum_{i \in \mathbf{VI}} \lVert \mathbf{p}_i - K \cdot H \cdot \mathbf{Q}_i \rVert^2 .
\end{gather}

While different algorithms to solve the \ac{PnP} problem have been proposed, they are usually highly sensitive to outliers due to the least-squares nature of the optimization. For this reason, we employ the robust \ac{RANSAC} estimator~\cite{fischler1981random} by randomly selecting four correspondences, estimating the camera pose based on these correspondences using the EPnP algorithm~\cite{lepetit2009epnp} and calculating how many points have a reprojection error lower than a threshold (inliers). We repeat this process for $N_R$ iterations and we select the pose with the highest number of inliers.
Since the LiDAR images can contain tens of thousands of points, computing the reprojection error for each hypothesis is computationally expensive, and thus, we implemented a GPU-based computation of the reprojection errors.

\begin{figure*}
  \centering
      \includegraphics[width=.99\textwidth]{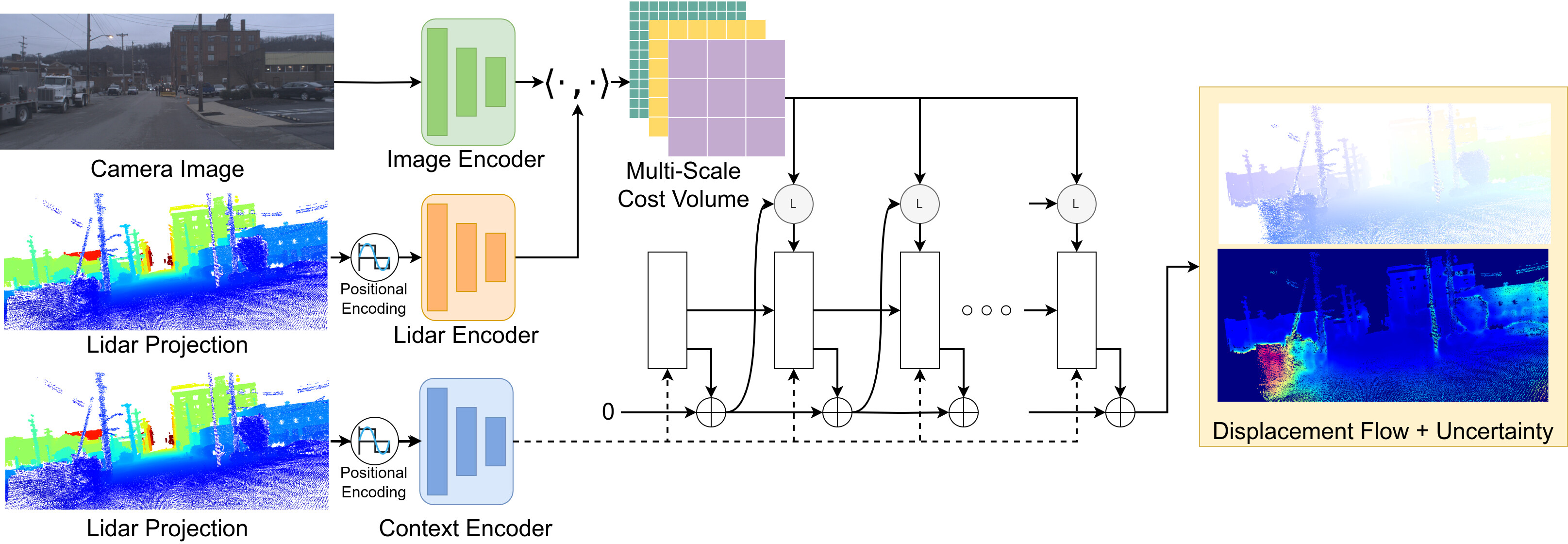}
      \caption{The network architecture of \cmrnet2{} is based upon RAFT~\cite{teed2020raft}. The camera image and the LiDAR image are processed by the image and LiDAR encoders, respectively, and their features are used to compute a multi-scale cost volume. The LiDAR image is additionally processed by the context encoder, which is then used together with the cost volume to iteratively refine the optical flow using a GRU module. The output of the network is a pixel-wise camera-LiDAR displacement map and the corresponding uncertainty map.
      The displacement flow is color-coded based on~\cite{baker2011database}, while the uncertainty is colorized based on the (normalized) sum of the component-wise uncertainties $\sigma_u + \sigma_v$.}
      \label{fig:raft}
      \vspace{0.2cm}
  \end{figure*}

\subsection{Network Architecture}
\label{sec:network}
We build our pixel-level matching network based on the RAFT architecture~\cite{teed2020raft}, originally proposed for optical flow estimation. RAFT outperformed previous methods by a large margin and demonstrated better generalization ability, and many of the most recent state-of-the-art approaches are built upon it~\cite{xu2023unifying,Mehl_2023_WACV}. Although RAFT shares some similarities with previous coarse-to-fine architectures, such as the well-known PWCNet~\cite{Sun_2018_CVPR}, it presents some key differences that allow it to achieve better performance at a lower computational cost. In coarse-to-fine architectures, a \textit{local} correlation volume (where each pixel is compared against neighboring pixels) is used to predict a coarse optical flow (\eg $1/32$-th of the original image size). The predicted flow is then upsampled by a factor of two and used to warp the features of the image. This process is repeated until the desired resolution is reached (\eg $1/4$-th of the original resolution). In contrast, RAFT computes an \textit{all-pairs} correlation volume by comparing all pixels in one feature map with every pixel in the other feature map. Although this correlation volume is more computationally and memory intensive than a local correlation volume, it is only computed once, and it does not require any warping operation. The optical flow is initialized to zero and iteratively refined via a recurrent \ac{GRU} module.

Our modified version of RAFT for camera to LiDAR matching is depicted in~\cref{fig:raft} and differs from the original architecture by using separate feature encoders for camera and LiDAR inputs, using the LiDAR image in the context encoder with added positional encoding, integrating uncertainty estimation in the update block, and replacing the L1 loss with a masked Negative Log-Likelihood loss. We will analyze all components in the following.
The image encoder first employs a convolutional layer with downsampling by a factor of 2, followed by six residual blocks and downsampling at every other block. The final image feature map has a shape of $H/8 \times W/8 \times 256$. The LiDAR and context encoders share the same architecture as the image encoder, except for an additional Fourier feature mapping module. Recent work~\cite{NEURIPS2020_55053683} has shown that Fourier features help coordinate-based \acp{DNN} learn high-frequency functions and improve their performance in a wide variety of tasks. Similar feature mapping modules have been used in seminal works such as the positional encoding employed in Transformers~\cite{vaswani2017attention} and NERFs~\cite{mildenhall2021nerf}. In particular, we define the mapping $\Phi : \mathbb{R} \to \mathbb{R}^{2m+1}$ as follows:
\begin{equation}
\label{eq:pos_encoding}
\begin{split}
   \Phi(d) = &[d, \sin(d \pi 2^0), \cos(d \pi 2^0), \dots, \\
  & \sin(d \pi 2^{m - 1}), \cos(d \pi 2^{m - 1}) ] ,
\end{split}
\end{equation}
where $d$ is the depth in the LiDAR projection and $m$ is a hyperparameter that defines the number of frequencies.

Given the image features $f^{2D} \in \mathbb{R}^{H/8 \times W/8 \times C}$ and the LiDAR features $f^{3D} \in \mathbb{R}^{H/8 \times W/8 \times C}$, where $C$ is the number of channels, the cost volume $\mathbf{c} \in \mathbb{R}^{H/8 \times W/8 \times H/8 \times W/8}$ is defined as the similarity, calculated using the dot product, between every pair of pixels $\mathbf{C}_{ijkl} = \sum_{h} f^{2D}_{ijh} \cdot f^{3D}_{klh}$. The correlation volume is further downsampled in the last two dimensions by a factor of $\{1, 2, 4, 8\}$, resulting in a multi-scale cost volume $\{\mathbf{C}^1, \mathbf{C}^2, \mathbf{C}^3, \mathbf{C}^4\}$.
The final component of our network starts from an empty flow field $\mathbf{f}_0 = \mathbf{0}$ and iteratively updates it using a \ac{GRU} unit. At the $k$-th iteration, the predicted flow $\mathbf{f}^k$ is used to retrieve, for each pixel $x = (u, v)$ of the LiDAR projection, a patch from the multi-scale correlation volume centered around the predicted corresponding image pixel $x' = x \oplus \mathbf{f}^k_{u,v}$ using a lookup operation. The next \ac{GRU} hidden state $h_{k+1}$ is then computed as follows:
\begin{align}
  &z_{k+1} = \sigma (Conv_{3 \times 3} ([ h_k, x_{k + 1}] , W_z)), \\
  &r_{k+1} = \sigma (Conv_{3 \times 3} ([ h_k, x_{k + 1}] , W_r)), \\
  &\tilde{h}_{k+1} = tanh(Conv_{3 \times 3}( [r_{k+1} \odot h_k, x_{k + 1}], W_h)), \\
  &h_{k+1} = (1 - z_{k+1}) \odot h_k + z_{k+1} \odot \tilde{h}_{k+1} ,
\end{align}
where $x_{k + 1}$ is the concatenation of the patch from the correlation volume, the output of the context encoder, and the current predicted flow $\mathbf{f}^k$. The $Conv_{3 \times 3}$ operation denotes a convolutional layer with a kernel size of $3 \times 3$, $W_z, W_r,$ and $W_h$ are the weights of the convolutional layers, and $\sigma$ is the sigmoid operation.
Afterward, the hidden state $h_{k+1}$ is used to predict the next residual flow $\Delta \mathbf{f}^{k+1}$ and the respective uncertainty $\boldsymbol{\sigma}^{k+1}$ using two convolutional layers. The residual flow is then added to the previous flow $\mathbf{f}^k$ to obtain the updated flow $\mathbf{f}^{k+1} = \mathbf{f}^k + \Delta \mathbf{f}^{k+1}$. This process is repeated $N$ times to obtain the final flow. Since the \ac{GRU} weights $W_z, W_r,$ and $W_h$ are shared across iterations, the number of update iterations can be increased during inference for improved performance.
Finally, the predicted flow at every iteration is upsampled to the original image resolution using learned convex upsampling~\cite{teed2020raft}.

\subsection{Loss Function}

Following~\cite{teed2020raft}, we supervise the predicted pixel displacements $\mathbf{f}^k$ at every iteration $k$ of the \ac{GRU} update. Since \cmrnet2 also predicts uncertainties $\boldsymbol{\sigma}^k = (\sigma_u^k, \sigma_v^k)$ in addition to the displacements, we replace the \ac{MAE} distance originally used in RAFT with the negative log-likelihood as follows
\begin{gather}
  \mathcal{L}_{reg} = \sum_{k = 1}^{N} \frac{\sum_{u,v} - \log p(\Delta P_{u,v} | \mathbf{f}^k_{u,v}, \boldsymbol{\sigma}^k_{u,v}) \cdot mask_{u,v}}{\sum_{u,v} mask_{u,v}} \gamma^{(N - k)} \label{eq:loss} ,
\end{gather}
where $\gamma < 1$ is used to exponentially increase the weight of the loss for later iterations of the GRU.

\subsection{Iterative Refinement}
\label{sec:iterative}
Due to the incorrect initial position $H_{map}^{init}$, the LiDAR-Image $\mathcal{D}$ and the camera image $\mathcal{I}$ can have a small overlapping \ac{FoV}, thus sharing only a few points that can be matched by our method.
We, therefore, employ an iterative refinement approach in which multiple instances of \cmrnet2 are independently trained using different initial error ranges. During inference, the pair ($\mathcal{I}$, $\mathcal{D}$) is fed to the instance of \cmrnet2 trained with the highest initial error range, and the predicted pose $H^*_1$ is used to generate a new synthesized LiDAR-image $\mathcal{D}_2$. The latter is subsequently fed to the second instance of \cmrnet2, which is trained with a lower initial error range. This process can be repeated multiple times, using instances of \cmrnet2 trained with increasingly small error ranges, iteratively reducing the estimated localization error.

\begin{table*}
  \centering
  \caption{Sensor setup in the evaluation datasets}
  \label{tab:datasets}
  \begin{threeparttable}
    \begin{tabular}{lcccccc}
    \toprule
     & KITTI & Argoverse & Pandaset & Freiburg - Car & Freiburg - Quadruped & Freiburg - UAV \\ 
     \midrule
     \multirow{2}{*}{\parbox[l]{1cm}{City}} & \multirow{2}{*}{\parbox[c]{1cm}{\centering Karlsruhe}} & Miami / & San Francisco / & \multirow{2}{*}{\parbox[c]{1cm}{\centering Freiburg}} & \multirow{2}{*}{\parbox[c]{1cm}{\centering Freiburg}} & \multirow{2}{*}{\parbox[c]{1cm}{\centering Freiburg}} \\
      & & Pittsburgh & El Camino Real & & & \\ \midrule
      LiDAR & HDL-64E & 2 x VLP-32 & Pandar64 / PandarGT & OS1-128 & OS1-128 & OS1-128 \\ \midrule
      RGB Cameras & 2 & 9 & 6 & 4 & 6 & 1 \\ \midrule
      Camera & $1224\times370$ / & $1920\times1200$ / & \multirow{2}{*}{\parbox[c]{2cm}{\centering $1920\times1080$}} & \multirow{2}{*}{\parbox[c]{2cm}{\centering $1920\times1200$}} & \multirow{2}{*}{\parbox[c]{2cm}{\centering $2464\times2056$}} & \multirow{2}{*}{\parbox[c]{2cm}{\centering $2048\times1536$}} \\
      Resolution & $1242\times376$ & $2056\times2464$ & & & & \\ \midrule
      Included in Training & \ding{51} & \ding{51} & \ding{51} & \ding{55} & \ding{55} & \ding{55} \\
    \bottomrule
  \end{tabular}
   \end{threeparttable}
\end{table*}

\section{Experimental Results}\label{sec:experimental-results}
In this section, we will detail and analyze the extensive experimental evaluations we performed to validate our novel \cmrnet2. First, we present the datasets we used to train and evaluate our approach, followed by the training and implementation details. We then report qualitative and quantitative results on two different tasks: monocular localization in LiDAR maps and extrinsic camera-LiDAR calibration. Although the two tasks are similar in nature, they tackle two different use cases, and they present different and complementary challenges.
Finally, we present multiple ablation studies aimed at validating the architectural choices we made when designing \cmrnet2. Unless otherwise specified, to mitigate the effect of the random initial error added to the ground truth poses, all experiments were performed using three different random seeds, and the results were averaged.

\subsection{Datasets}
\label{sec:datasets}
We evaluate \cmrnet2 on four real-world autonomous driving datasets recorded in different countries, with different sensor setups, and including different traffic scenarios. We would like to emphasize that, differently from existing learning-based methods for camera-LiDAR matching, \cmrnet2 is trained on multiple datasets simultaneously and that the same network's weights are used to evaluate our method on different datasets without any retraining or fine-tuning, demonstrating that \cmrnet2 can effectively generalize to different sensor setups and different cities. \cref{tab:datasets} summarizes the main characteristics of the datasets we used for training and evaluation.
 
\subsubsection{KITTI} The KITTI dataset~\cite{Geiger2012CVPR} was recorded in Karlsruhe, Germany, and includes both urban and interurban scenarios. Accurate localization and semantic information for the LiDAR point clouds are additionally provided for the “odometry” sequences in the SemanticKITTI dataset~\cite{behley2019iccv}. The recording vehicle was equipped with two forward-facing RGB cameras in a stereo configuration that provides images ranging from $1224\times370$ to $1242\times376$ pixels at 10 FPS. It is also equipped with a Velodyne HDL-64E that we used to generate the LiDAR-maps. We include the \textit{left} camera images from the odometry sequences 03, 05, 06, 07, 08, and 09 in the training set (total of \num{11426} frames), and we use the sequence 00 for evaluation (\num{4541} frames).

\subsubsection{Argoverse} The Argoverse dataset~\cite{Argoverse} was recorded in urban areas around the cities of Miami and Pittsburgh, USA. Precise localization and 3D bounding boxes are provided with the dataset. The recording vehicle is equipped with seven ring RGB cameras (1920$\times$1200 at 30 FPS) and two forward-facing stereo cameras (2056$\times$2464 at 5 FPS). It is also equipped with two Velodyne VLP-32, whose output we use to generate the LiDAR-maps, exploiting the ground truths provided with the dataset. We used the central forward-facing camera images from splits \textit{train1}, \textit{train2} and \textit{train3} of the ``3D tracking dataset'' as training set (\num{36347} frames), and the \textit{train4} split as the validation set (\num{2741} frames).

\subsubsection{Pandaset} The Pandaset dataset~\cite{xiao2021pandaset} was recorded in San Francisco and El Camino Real and includes six cameras (1920$\times$1080 at 10 FPS), a 64-beam Pandar64 spinning LiDAR, and a PandarGT solid-state LiDAR. The dataset additionally provides localization and semantic segmentation for LiDAR point clouds. We use the three forward-facing cameras (front left, front right, and front center) of nine randomly selected sequences as validation (\num{2160}) and the images from the other sequences as training (\num{12720}).

\subsubsection{In-House Datasets} We additionally recorded three in-house datasets by driving around the city of Freiburg, Germany, and around our university campus using three different robotic platforms, a \textbf{Self-driving Perception Car}, a \textbf{Quadruped} robot, and a quadcopter \textbf{UAV}. The three platforms are depicted in the lower box of~\cref{fig:teaser}. All platforms are equipped with an ouster OS1-128 LiDAR and a forward-facing RGB camera. Resolution and field of view of the cameras mounted on the three platforms are different and are reported in \cref{tab:datasets}. The intrinsic parameters of all cameras were computed using the Kalibr\footnote{\href{https://github.com/ethz-asl/kalibr}{https://github.com/ethz-asl/kalibr}} toolbox. The in-house datasets are used to evaluate the generalization ability of our approach to different cities and sensor setups and, therefore, are not included in the training set. In particular, the Self-driving Perception Car dataset is used to evaluate the generalization ability on the monocular localization task, while all three datasets are used to evaluate the generalization ability on the extrinsic calibration task, as the sensors on the Quadruped and UAV platforms are not time-synchronized.

\subsection{Training Details}
Unless otherwise specified, we train \cmrnet2{} on four NVIDIA GeForce RTX 3090 with a total batch size of 16, using the Adam optimizer with a base learning rate of $3*10^{-4}$ and a weight decay of $5*10^{-6}$. We use a OneCycle learning rate scheduler and we train the network in two stages. In the first stage, we train the network without uncertainty estimation for 150 epochs using the original MAE-based loss function from RAFT~\cite{teed2020raft}. In the second stage, we fine-tune the network with uncertainty estimation for 50 epochs using the loss function defined in \cref{eq:loss}. We use $m=12$ frequencies for the Fourier feature mapping module in \cref{eq:pos_encoding} and we set $\gamma=0.8$ in \cref{eq:loss}.

\subsection{LiDAR-maps Generation}

In order to generate LiDAR-maps for the four aforementioned datasets, we first aggregate single scans at their respective ground truth position, which is provided by the dataset itself (KITTI, Argoverse, and Pandaset). We then downsample the resulting maps at a resolution of \SI{0.1}{\meter} using the Open3D library~\cite{Zhou2018}. Moreover, as we would like to have only static objects in the maps (\eg no pedestrians or cars), we remove dynamic objects by exploiting the 3D bounding boxes provided with Argoverse and the semantic segmentation for SemanticKITTI and Pandaset. Alternatively, we can also use dynamic object detection methods\cite{bevsic2022dynamic} when such information is not available. For our in-house Freiburg - Car dataset, we used FAST-LIO2~\cite{xu2022fast}, a state-of-the-art LiDAR-inertial odometry method, to generate the ground truth poses and the LiDAR-map.

\subsection{Training on Multiple Datasets}

We train \cmrnet2 by combining training samples from KITTI, Argoverse, and Pandaset datasets. Training a \ac{CNN} on multiple diverse datasets creates certain challenges. First, the different cardinality of the three training datasets (\num{11426}, \num{36347}, and \num{12720}, respectively) might lead the network to perform better on one dataset than the other.
To overcome this problem, we randomly sampled a subset of Argoverse and Pandaset at the beginning of every epoch to have the same number of samples as KITTI.
As the subset is sampled every epoch, the network will eventually process every sample from the datasets.

Moreover, the three datasets have cameras with very different fields of view and resolution. To address this issue, we resize the images so to have the same resolution.
One straightforward way to accomplish this would be to just crop the Argoverse and Pandaset images. However, this would yield images with a very narrow field of view, making the matching between the RGB and LiDAR-image increasingly difficult. Therefore, we first downsample the Argoverse and Pandaset images to half the original resolution and then randomly crop images from all datasets to $960\times320$ pixels. We perform this random cropping at runtime during training in order to have different crop positions for the same image at different epochs.
Since we generate both the LiDAR-image $\mathcal{D}$ and the ground truth displacements $\Delta P$ at the original resolution, and only after we downsample and crop them, we don't need to update the intrinsic parameters of the camera. Moreover, we also halve the pixel displacements $\Delta P$ during the downsampling operation.

\subsection{Initial Pose Sampling and Data Augmentation}

We employ the iterative refinement approach presented in \Cref{sec:iterative} by training three instances of \cmrnet2. To simulate the initial pose $H_{init}$, we add uniform random noise to all components of the ground truth pose $H_{GT}$, independent for each sample. The range of the noise that we add to the first iteration is [$\pm 2$ m] for the translation components and [$\pm$ \ang{10}] for the rotation components. The maximum range for the second and third iteration are {[$\pm 0.2$ m, $\pm$ \ang{0.5}] and [$\pm 0.05$ m, $\pm$ \ang{0.1}]}, respectively.

To improve the generalization ability of our approach, we employ a data augmentation scheme. First, we apply color jittering to the images by randomly changing the contrast, saturation, and brightness. Subsequently, we randomly horizontally mirror the images %
and we modify the camera calibration by updating the principal point $(c_x, c_y)$ as follows: $c'_x = W - c_x$, where $W$ is the width of the image. We then randomly rotate the images in the range [\ang{-5}, \ang{+5}]. Finally, we transform the LiDAR point cloud to reflect these changes before generating the LiDAR-image.
The LiDAR point clouds are expressed with the camera pose as the origin, and with reference frame X-forward, Y-left, and Z-up. To reflect the horizontal mirror of the image, we multiply the Y-coordinates by $-1$, and we rotate around the X-axis by the same angle used to rotate the image.

To summarize, we train every instance of \cmrnet2 as follows:
\begin{itemize}
    \item Randomly draw a subset of the Argoverse and Pandaset datasets and shuffle them with samples from KITTI
    \item For every batch:
\begin{itemize}
    \item Apply data augmentation to the images $\mathcal{I}$ and modify the camera matrices and the point clouds accordingly
    \item Sample the initial poses $H^{init}_{map}$
    \item Generate the LiDAR-images $\mathcal{D}$, the displacement ground truths $\Delta P$ and the masks $mask$
    \item Downsample $I, \mathcal{D}, \Delta P$ and $mask$ for the Argoverse and Pandaset samples in the batch
    \item Crop $I, \mathcal{D}, \Delta P$ and $mask$ to the resolution $960\times320$.
    \item Feed the batch ($I, \mathcal{D}$) to \cmrnet2, compute the loss using \cref{eq:loss}, and update the weights
\end{itemize}
\item Repeat for $150$ epochs.
\end{itemize}

\subsection{Inference}

During inference, we process one image at a time, and since the network architecture is fully convolutional, we do not crop the images of different datasets to have the same resolution. Therefore, we feed the whole image to \cmrnet2. However, we still downsample the images of the Argoverse, Pandaset, and our in-house datasets to have a similar field of view as the images used for training, and we successively upsample the displacement flow predicted by CMRNext to the original resolution and scale it accordingly. Once we have the set of 2D-3D correspondences predicted by the network, we apply EPnP+RANSAC to estimate the pose of the camera with respect to the map, as detailed in \Cref{sec:localization}. We repeat this whole process three times using three specialized instances of \cmrnet2 to iteratively refine the estimation.

\subsection{Monocular Localization}
\label{sec:monocular_loc}
\begin{table*}
  \centering
  \caption{Comparison of median translation and rotation errors for the localization task.}
  \label{tab:localization_comparison}
  \begin{threeparttable}
    \begin{tabular}{clcccccccc}
    \toprule
     && \multicolumn{2}{c}{KITTI} & \multicolumn{2}{c}{Argoverse} & \multicolumn{2}{c}{Pandaset} & \multicolumn{2}{c}{Freiburg - Car} \\ \cmidrule(lr){3-4} \cmidrule(lr){5-6} \cmidrule(lr){7-8} \cmidrule(lr){9-10}
    && Transl. [cm] & Rot. [°] & Transl. [cm] & Rot. [°] & Transl. [cm] & Rot. [°] & Transl. [cm] & Rot. [°] \\ \midrule

    \multirow{6}{*}{\rotatebox[origin=c]{90}{\parbox[c]{1cm}{\centering\scriptsize Camera\\Dependent}}} 
    &CMRNet~\cite{Cattaneo_2019} & 45 & 1.35 & {\color{black}48} & {\color{black}0.83} & {\color{black}123} & {\color{black}1.25} & - & - \\
    &HyperMap~\cite{chang2021hypermap} & 48 & 1.42 & 58 & 0.93 & - & - & - & - \\
    &PSNet~\cite{wu2022psnet} & 25 & 0.63 & - & - & - & - & - & - \\
    &BEVLOC~\cite{chenbevloc} & 39 & 1.28 & - & - & - & - & - & - \\
    &POET~\cite{miao2023poses} & 41 & 1.39 & - & - & - & - & - & - \\
    &I2P-Net~\cite{wang2023end} & \textbf{7} & \underline{0.67} & - & - & - & - & - & - \\

    \midrule

    \multirow{3}{*}{\rotatebox[origin=c]{90}{\parbox[c]{1cm}{\centering\scriptsize Camera\\Agnostic}}}
    & CMRNet++~\cite{cattaneo2020cmrnet} & 44 & 1.13 & 61 & 1.13 & \underline{77} & \underline{1.01} & \underline{84} & \underline{2.23} \\
    &I2D-Loc~\cite{chen2022i2d} & 18 & 0.70 & \underline{47} & \underline{0.71} & 144 & 3.36 & 129 & 4.50 \\
    &\textbf{\cmrnet2{} (ours)} & \underline{10} & \textbf{0.29} & \textbf{13} & \textbf{0.20} & \textbf{12} & \textbf{0.15} & \textbf{16} & \textbf{0.42} \\
    \bottomrule
  \end{tabular}
  \begin{tablenotes}[para,flushleft]
       \footnotesize
       All methods except I2P-Net employ an iterative refinement technique. For a fair comparison, in this table we report the results after the first iteration only.
     \end{tablenotes}
   \end{threeparttable}
\end{table*}

\begin{table*}
  \centering
  \caption{Comparison after iterative refinement for the localization task.}
  \label{tab:localization_lenc}
  \begin{threeparttable}
    \begin{tabular}{clcccccccc}
    \toprule
     && \multicolumn{2}{c}{KITTI} & \multicolumn{2}{c}{Argoverse} & \multicolumn{2}{c}{Pandaset} & \multicolumn{2}{c}{Freiburg - Car} \\ \cmidrule(lr){3-4} \cmidrule(lr){5-6} \cmidrule(lr){7-8} \cmidrule(lr){9-10}
    && Transl. [cm] & Rot. [°] & Transl. [cm] & Rot. [°] & Transl. [cm] & Rot. [°] & Transl. [cm] & Rot. [°] \\ \midrule

    & Initial Pose & 188 & 9.90 & 187 & 9.87 & 193 & 9.80 & 188 & 9.87 \\ \midrule
   \multirow{3}{*}{\rotatebox[origin=c]{90}{\parbox[c]{1cm}{\centering\scriptsize Single\\Iteration}}}  & {HyperMap~\cite{chang2021hypermap}} & {48} & {1.42} & - & - & - & - &- &- \\
    &{BEVLoc~\cite{chenbevloc}} & {39} & {1.28} & - & - & - & - &- &- \\
    &I2P-Net~\cite{wang2023end} & {7} & 0.67 & - & - & - & - &- &- \\ \midrule
    \multirow{7}{*}{\rotatebox[origin=c]{90}{\parbox[c]{1cm}{\centering\scriptsize {Iterative\\Refinement}}}} &{CMRNet~\cite{Cattaneo_2019}} & {27} & {1.07} & {\color{black}14} & {\color{black}0.44} & {\color{black}26} & {\color{black}0.60} &- &- \\
    &{PSNet~\cite{wu2022psnet}} & {25} & {0.63} & - & - & - & - &- &- \\
    &{POET~\cite{miao2023poses}} & {20} & {0.79} & - & - & - & - &- &- \\
    &I2D-Loc~\cite{chen2022i2d} & 8 & {0.30} & - & - & - & - &- &- \\
    &{CMRNet++~\cite{cattaneo2020cmrnet}} & {14} & {0.43} & {{25}} & {{0.45}} & - & - & - & - \\
    &\textbf{{\cmrnet2{} (same range)}} & {6.33} & {\textbf{0.23}} & {19.09} & {0.31} & {8.39} & {0.13} & {17.73} & {\textbf{0.27}} \\ 
    &\textbf{\cmrnet2{} (ours)} & \textbf{6.21} & \textbf{0.23 }& \textbf{7.51} & \textbf{0.16} & \textbf{7.18} & \textbf{0.11} & {\textbf{12.94}} & {0.28} \\ %
    
    \bottomrule
  \end{tabular}
   \end{threeparttable}
\end{table*}

We evaluate the monocular localization performance of \cmrnet2{} on the KITTI, Argoverse, Pandaset, and our in-house datasets. We compare our approach with the state-of-the-art camera-dependent methods CMRNet~\cite{Cattaneo_2019}, HyperMap~\cite{chang2021hypermap}, PSNet~\cite{wu2022psnet}, BEVLOC~\cite{chenbevloc}, POET~\cite{miao2023poses}, and I2P-Net~\cite{wang2023end}, as well as camera-agnostic methods CMRNet++~\cite{cattaneo2020cmrnet} and I2D-Loc~\cite{chen2022i2d}.

\begin{figure*}
  \centering
  \footnotesize
  \setlength{\tabcolsep}{0.05cm}
      {\renewcommand{\arraystretch}{1}
        \begin{tabular}{p{5.9cm}p{5.9cm}p{5.9cm}}

        \multicolumn{1}{c}{Initial Pose} & \multicolumn{1}{c}{Ground Truth} & \multicolumn{1}{c}{\cmrnet2{} (ours)} \\

      \includegraphics[width=\linewidth]{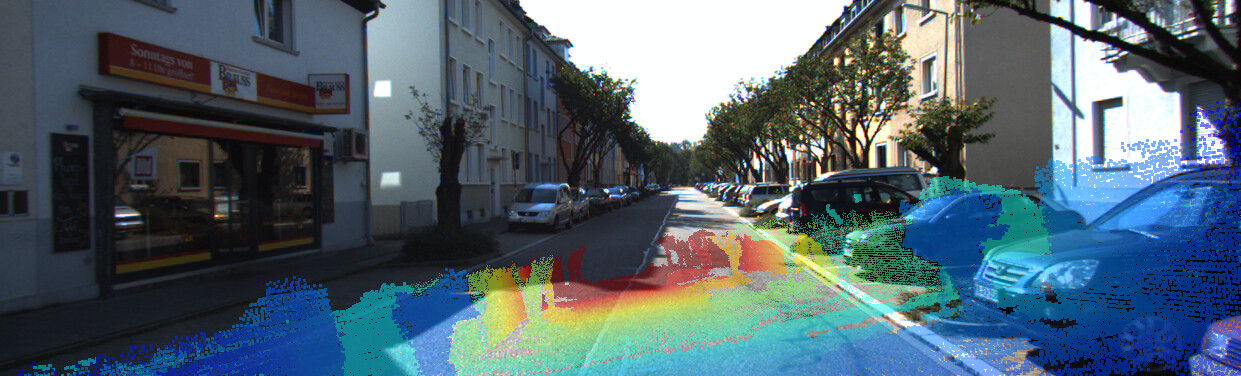} &
      \includegraphics[width=\linewidth]{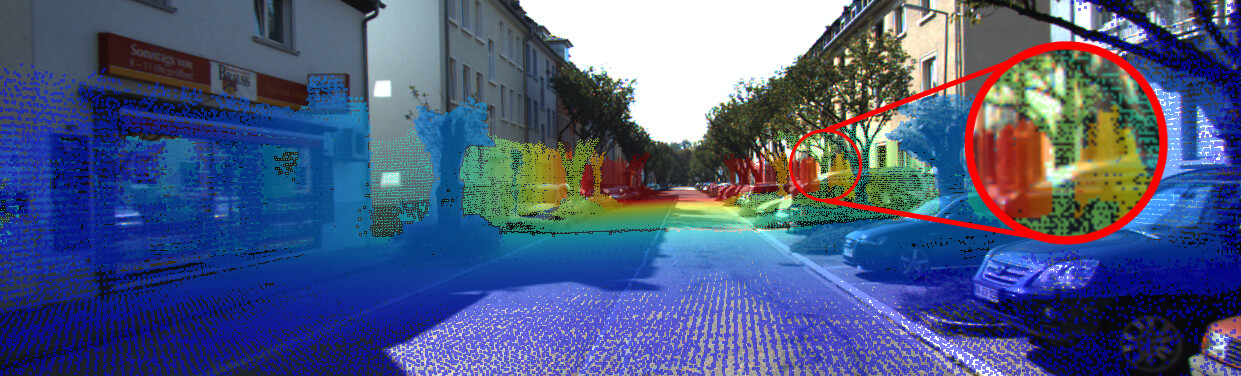} &
      \includegraphics[width=\linewidth]{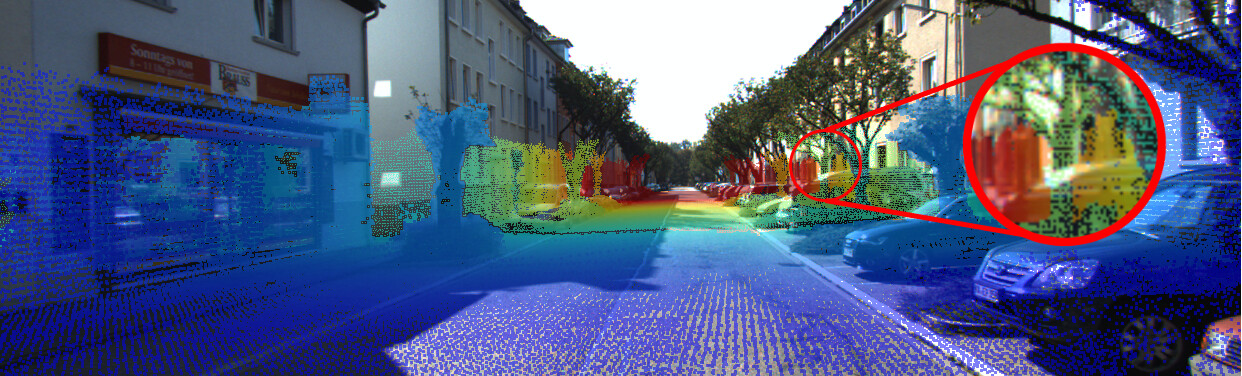} \\

      \includegraphics[width=\linewidth]{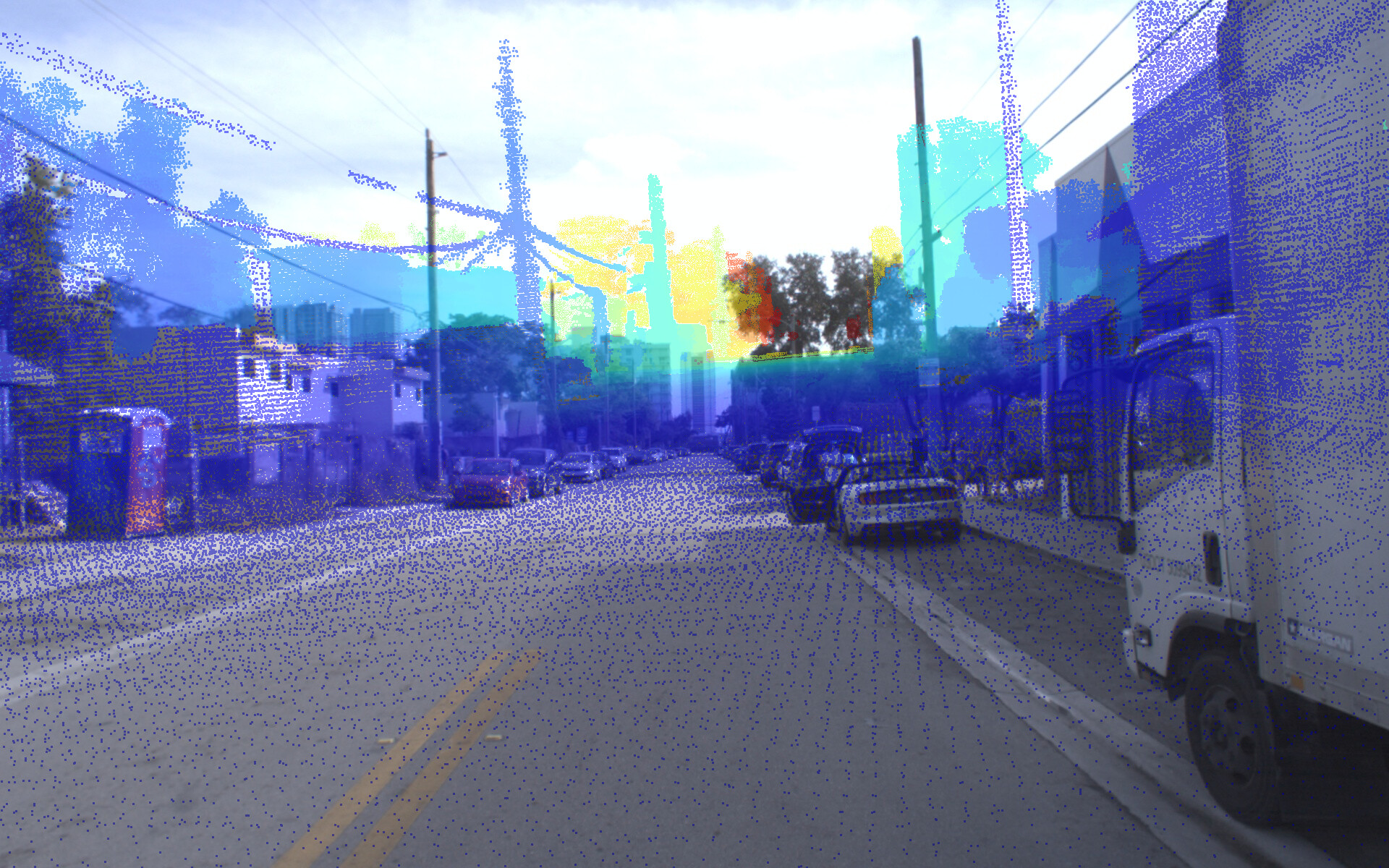} &
      \includegraphics[width=\linewidth]{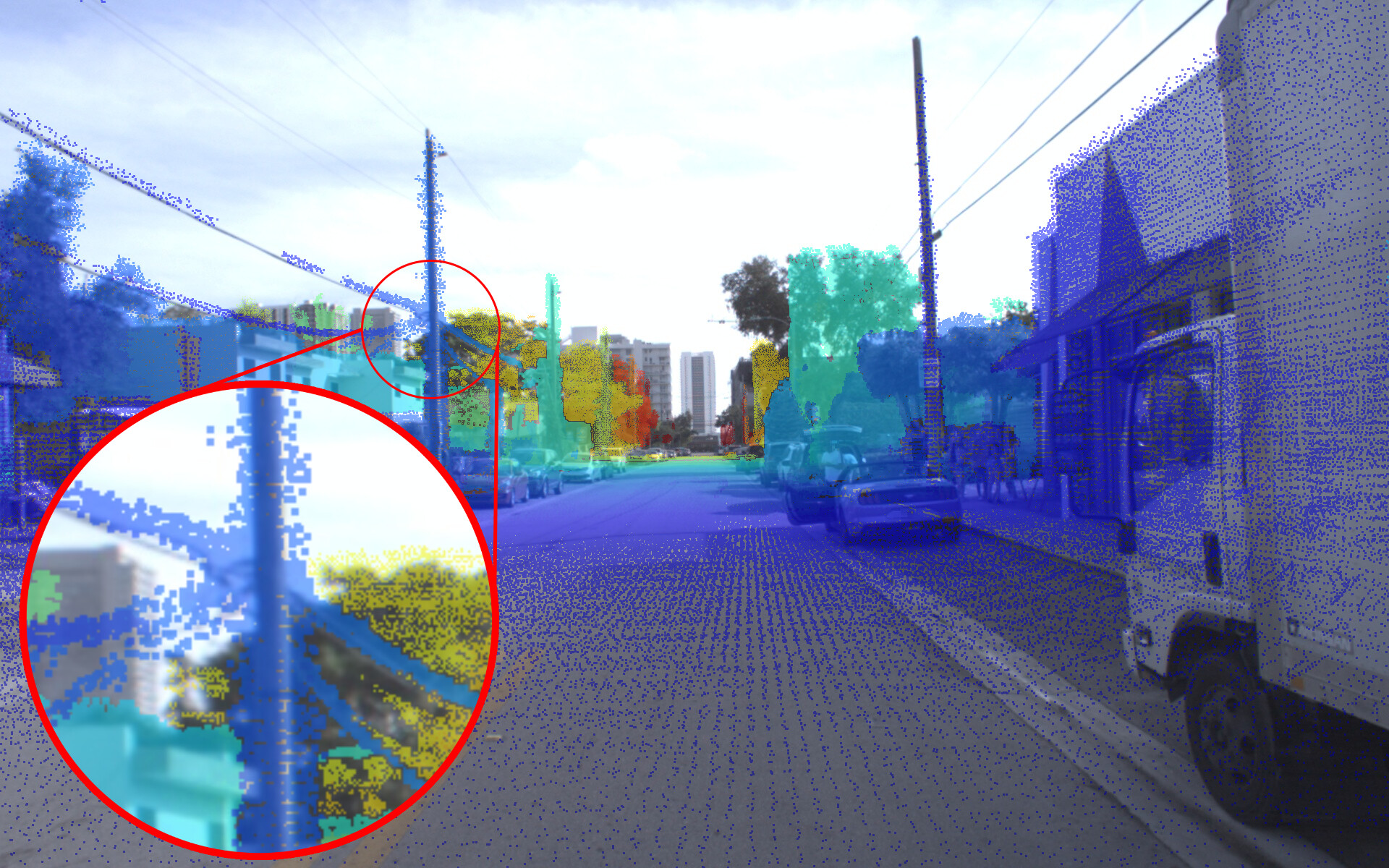} &
      \includegraphics[width=\linewidth]{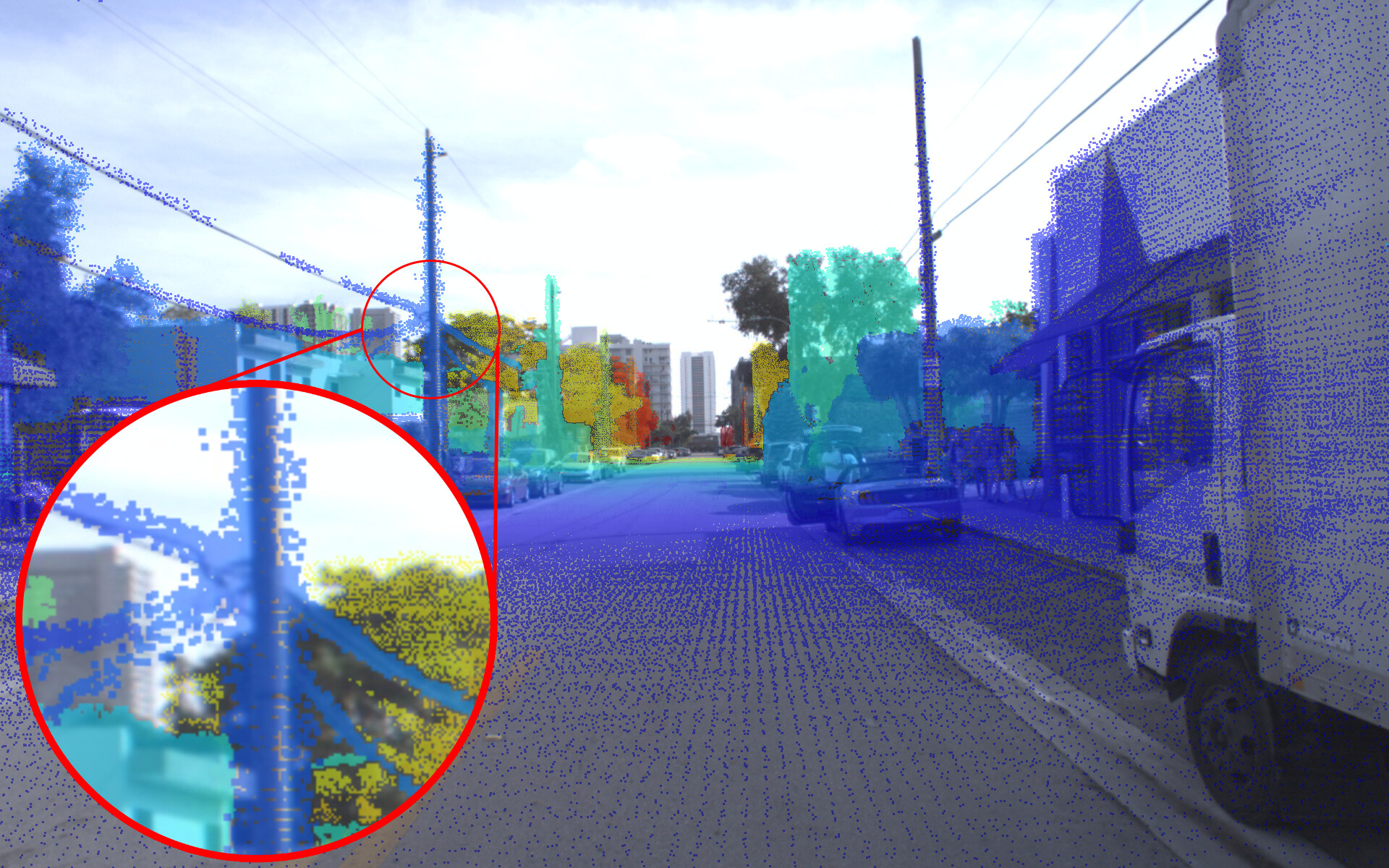} \\

      \includegraphics[width=\linewidth]{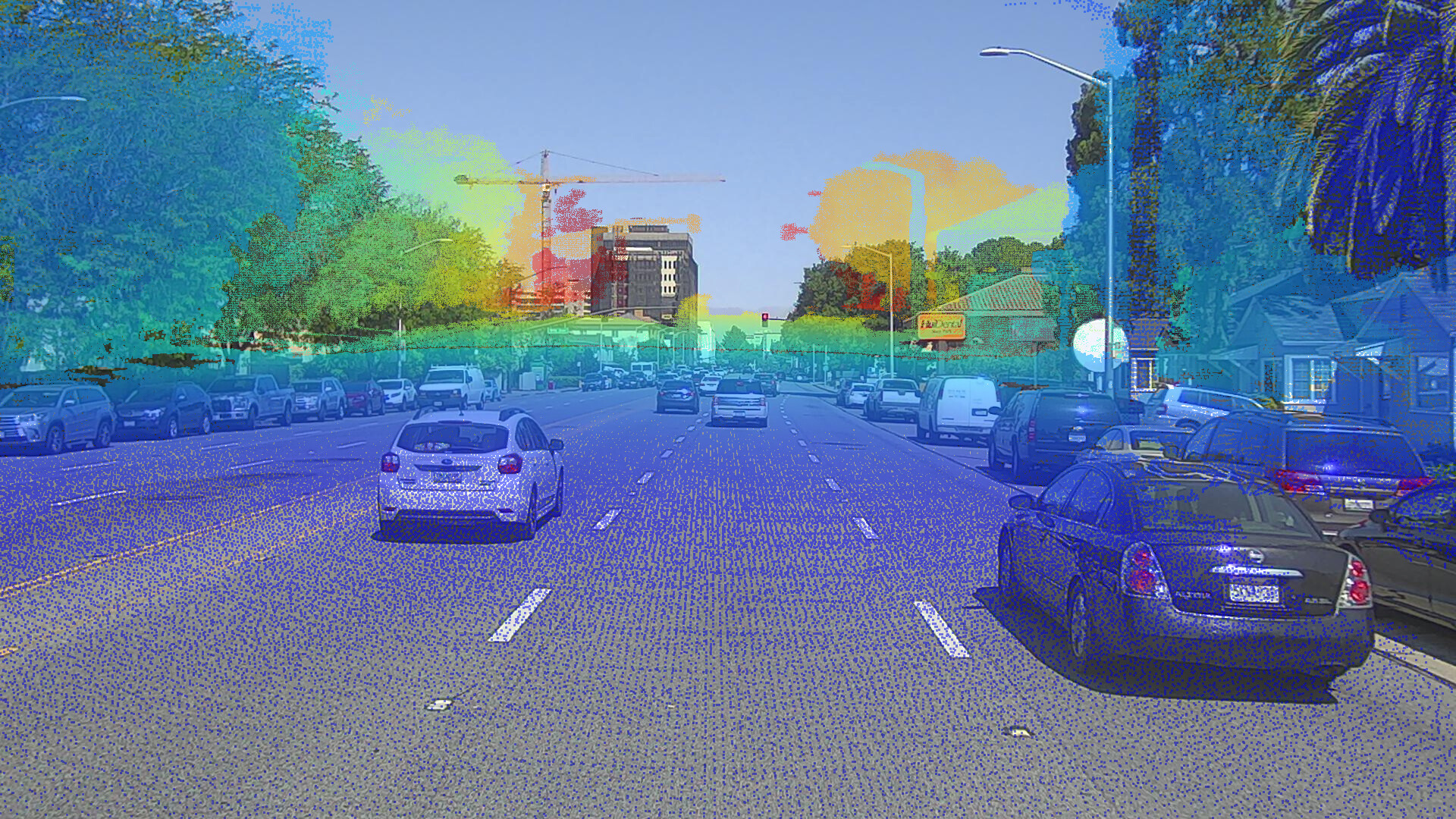} &
      \includegraphics[width=\linewidth]{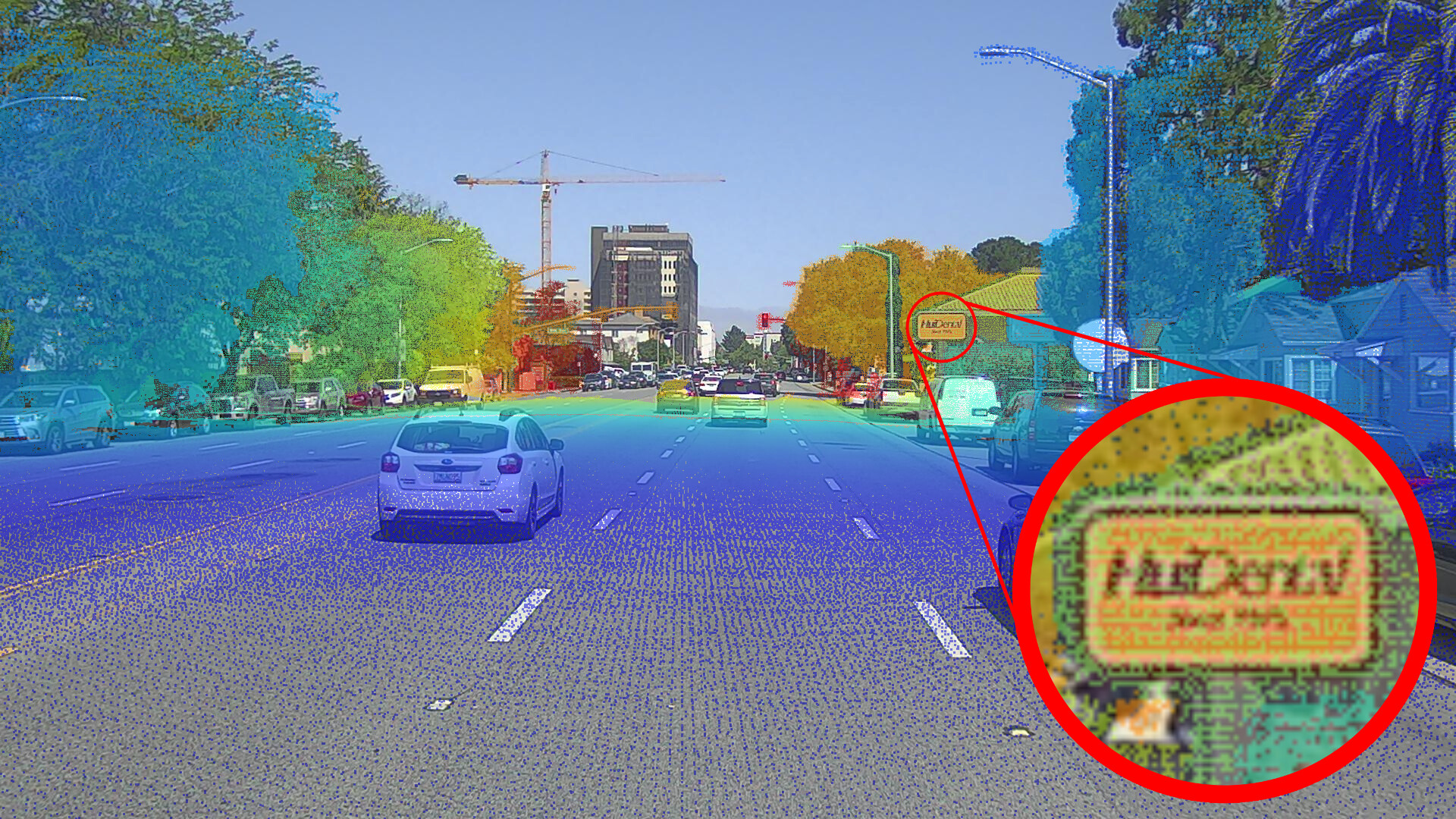} &
      \includegraphics[width=\linewidth]{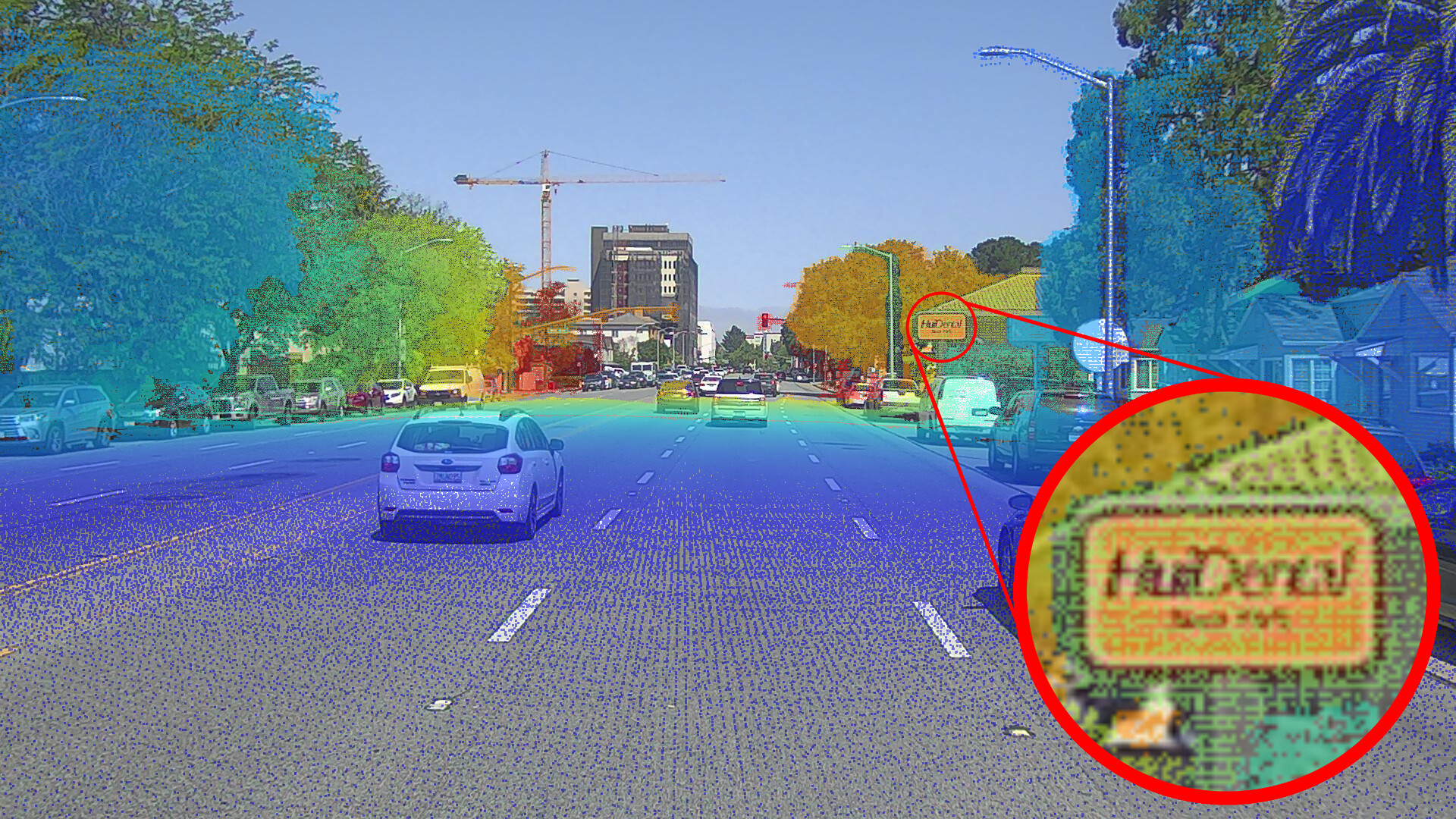} \\

      \includegraphics[width=\linewidth]{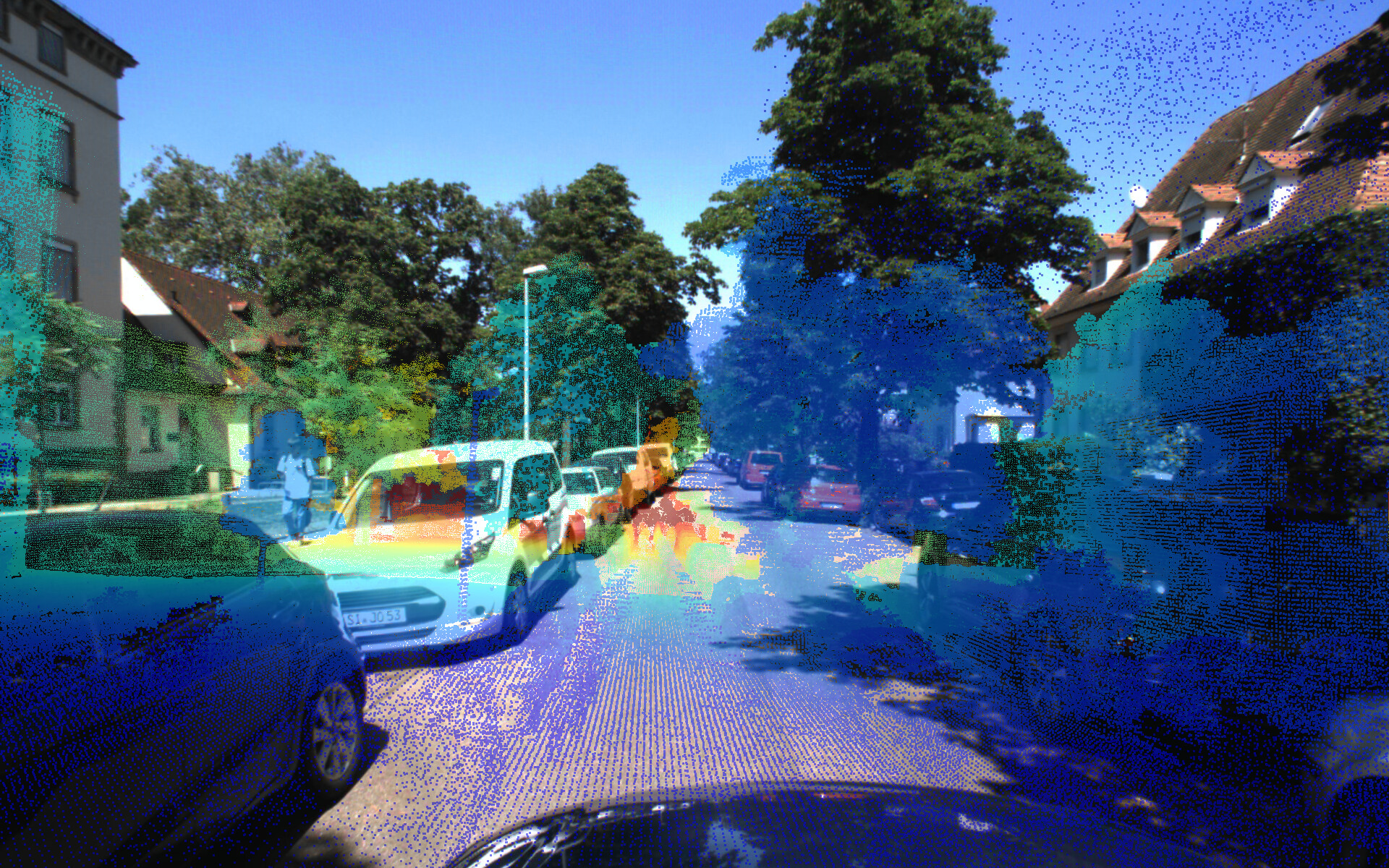} &
      \includegraphics[width=\linewidth]{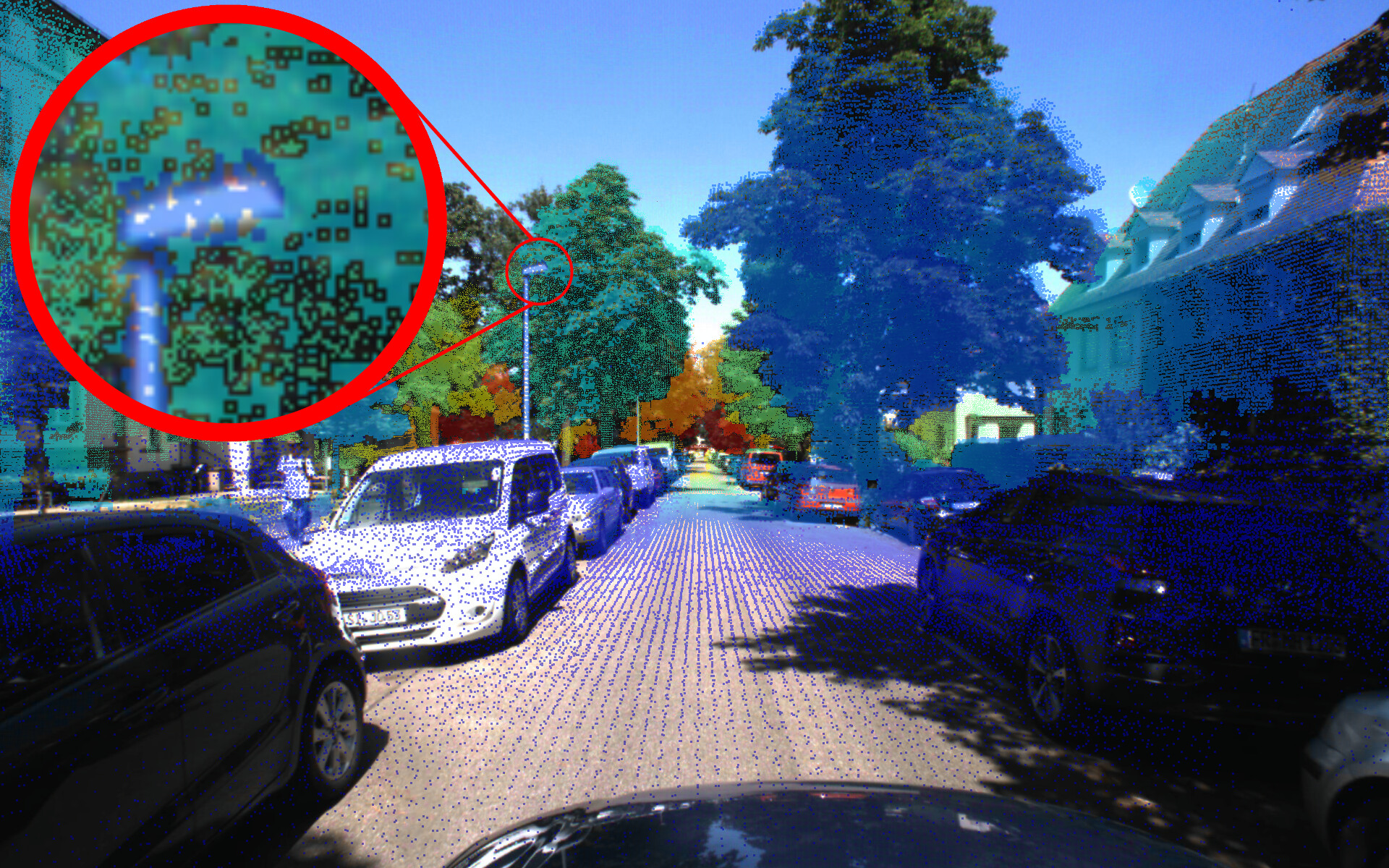} &
      \includegraphics[width=\linewidth]{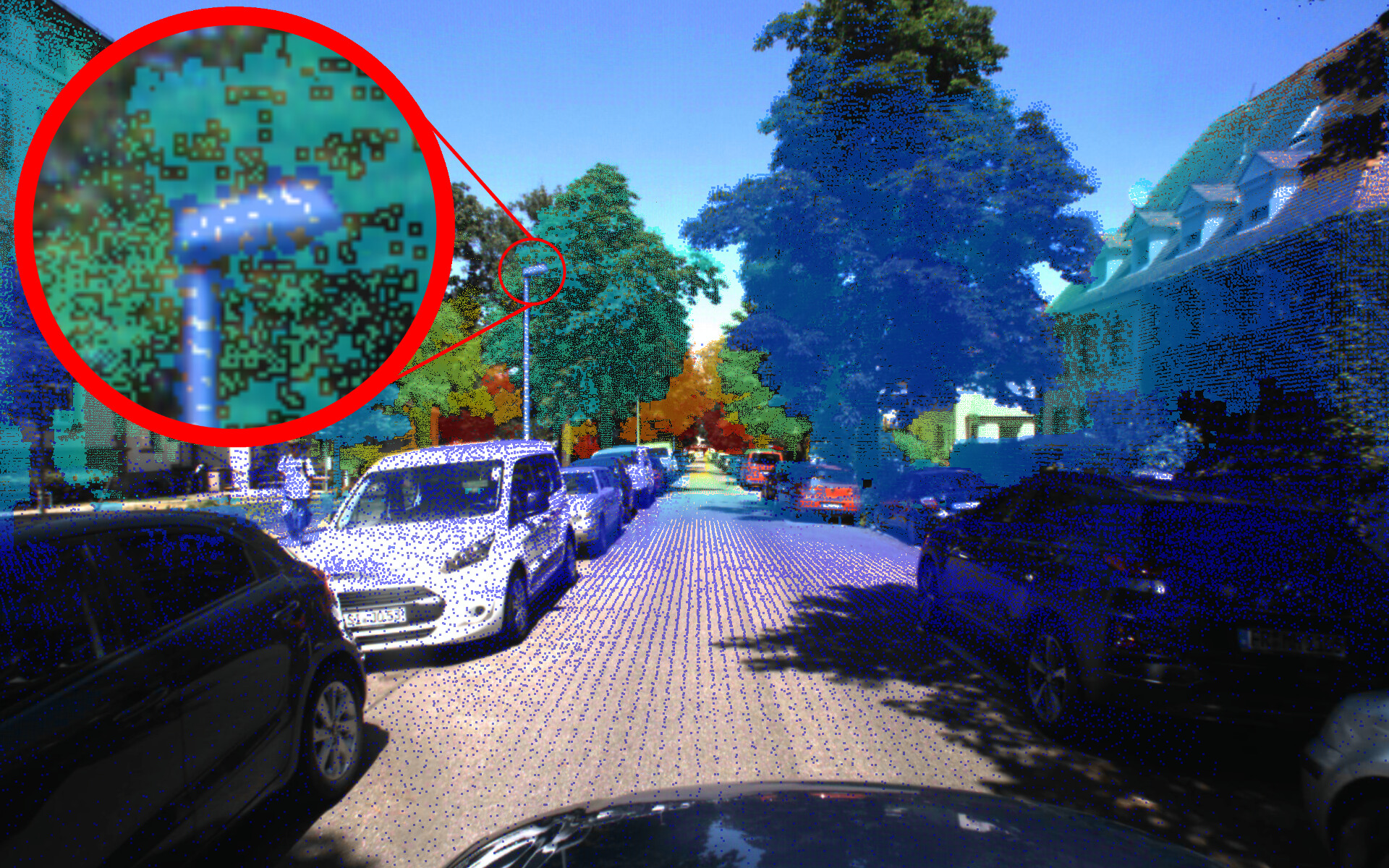} \\
    \end{tabular}}
      \caption{Qualitative results of \cmrnet2{} on the monocular localization task. From left to right: LiDAR image projected in the initial pose, ground truth pose, and pose predicted by \cmrnet2{}. All LiDAR projections are overlaid with the respective RGB image for visualization purposes. From top to bottom: KITTI, Argoverse, Pandaset, and Freiburg-Car datasets.}
      \label{fig:qualitative}
      \vspace{0.2cm}
\end{figure*}

\begin{figure*}[t]
    \centering

    \begin{subfigure}[b]{0.35\textwidth}
         \centering
         \includegraphics[width=\textwidth]{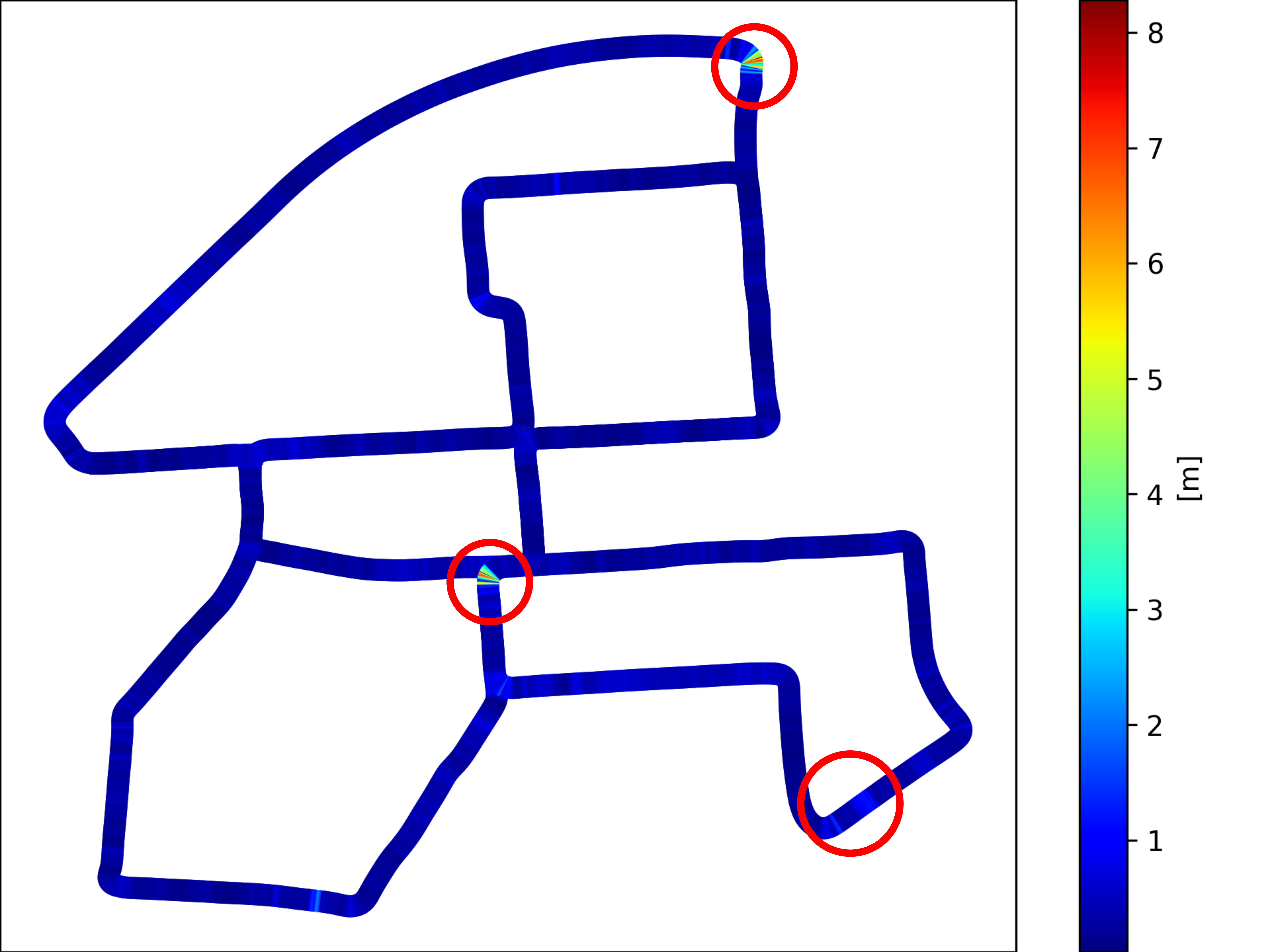}
         \caption{\color{black}Rotation Error}
     \end{subfigure}
     \begin{subfigure}[b]{0.35\textwidth}
         \centering
         \includegraphics[width=\textwidth]{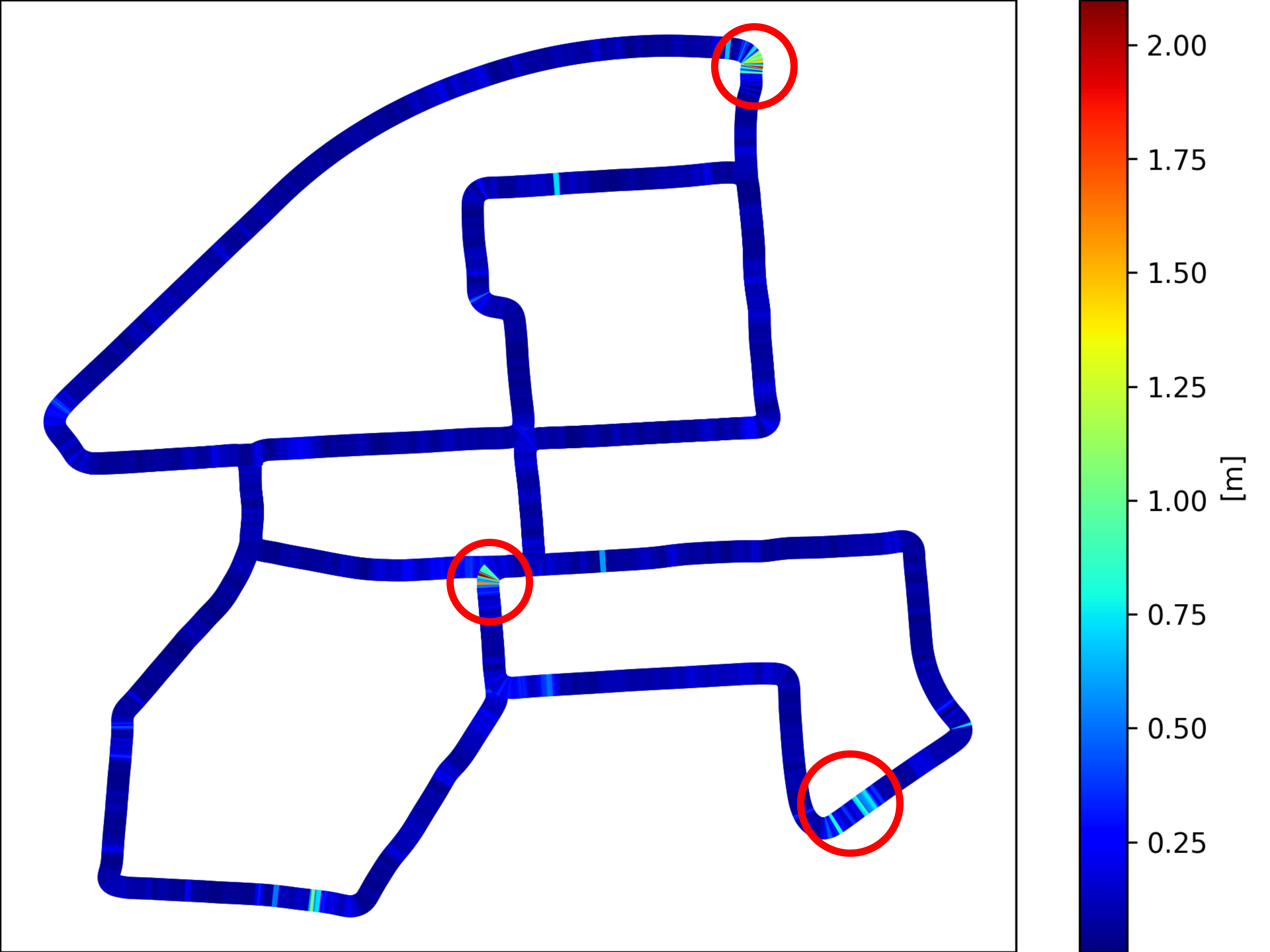}
         \caption{\color{black}Translation Error}
     \end{subfigure}
     \begin{subfigure}[b]{0.27\textwidth}
         \centering
         \includegraphics[width=\linewidth]{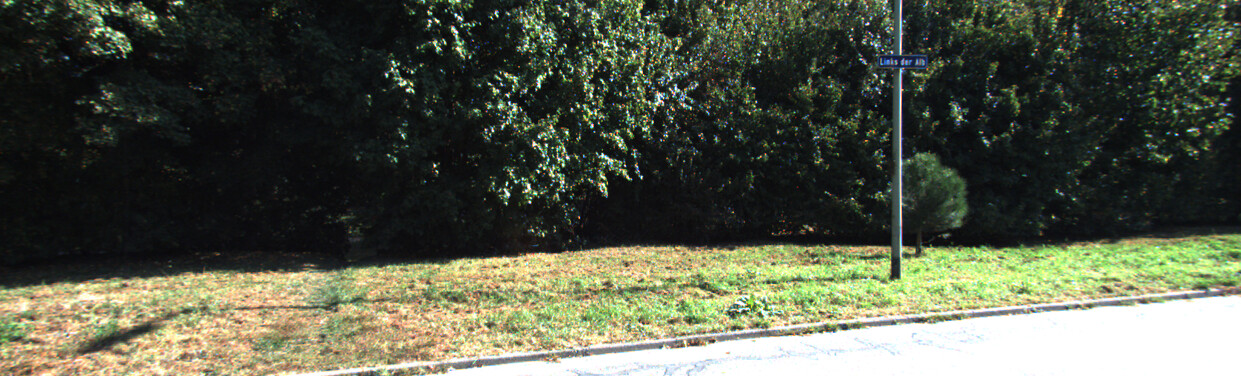}
         
         \vspace{0.1cm}
         
         \includegraphics[width=\linewidth]{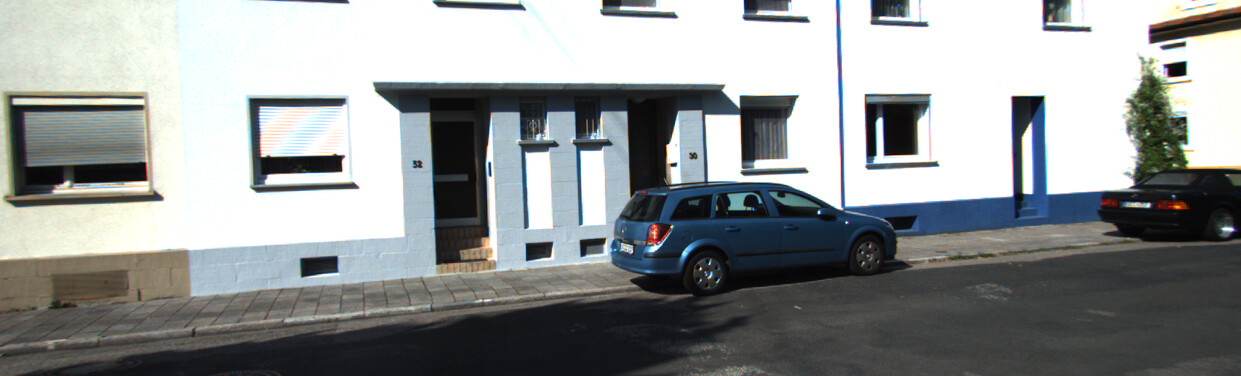}
         
         \vspace{0.1cm}
         
         \includegraphics[width=\linewidth]{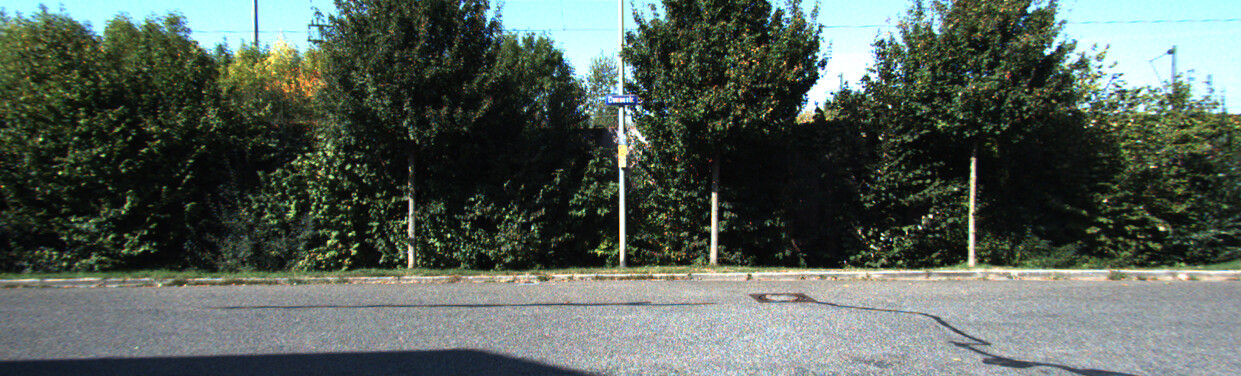}
         \caption{Failure Cases}
     \end{subfigure}
     
    \caption{(a) Rotation and (b) translation errors of CMRNext on the sequence 00 of the KITTI dataset, and (c) visualization of three scenes where CMRNext failed.}
    \label{fig:error_sequence}
\end{figure*}

Camera-dependent methods can only be evaluated on the same dataset they were trained on and cannot generalize to different sensor setups. Therefore, for these methods, a different model is trained for each dataset. HyperMap, PSNet, BEVLOC, POET, and I2P-Net do not provide a publicly available implementation, and therefore, we report the results from their respective papers~\cite{chang2021hypermap,wu2022psnet,chenbevloc,miao2023poses,wang2023end}. For CMRNet, we used the publicly available implementation\footnote{\url{https://github.com/cattaneod/CMRNet}} and we trained a model for each dataset. As our in-house Freiburg dataset is not included in the training set, but only used for evaluating the generalization ability, no camera-dependent method can be evaluated on this dataset.
On the other hand, camera-agnostic methods, such as I2D-Loc, CMRNet++, and our proposed \cmrnet2{}, can be trained on multiple datasets simultaneously, and the same model can be used to evaluate the performance on all four datasets. Although the code I2D-Loc is publicly available\footnote{\url{https://github.com/EasonChen99/I2D-Loc}}, only the training on the KITTI dataset is provided. While trying to train I2D-Loc on multiple datasets, we encountered several issues, and we were unable to reproduce the results reported in the original paper. Therefore, we report the results from the I2D-Loc paper~\cite{chen2022i2d} on KITTI and Argoverse, and we evaluated the pretrained model provided by the authors on Pandaset and Freiburg. All methods except for I2P-Net use an iterative refinement approach, where multiple networks are trained with different initial error ranges. To provide a fair comparison, for these methods, we only report the results of the first iteration, as the number and error ranges of successive iterations vary between methods.

All methods were evaluated in terms of median rotation and median translation errors, defined as:
\begin{eqnarray}
  E_t & =&  \left\lVert t - \tilde{t} \right\rVert_2 , \\
  m & =&  q * \tilde{q}^{-1} , \\
  E_r & =&  atan2\left(\sqrt{m_x^2 + m_y^2 + m_z^2}, m_w\right) ,
\end{eqnarray}
where $t$, $\tilde{t}$, $q$, and $\tilde{q}$ are the ground truth translation, predicted translation, ground truth quaternion, and predicted quaternion, respectively, $*$ and $^{-1}$ are the multiplicative and inverse operations for quaternions.
The results reported in~\cref{tab:localization_comparison} show that our proposed \cmrnet2{} outperforms all existing methods on all datasets, except for the translation error on KITTI, where I2P-Net achieved a slightly smaller error. However, the rotation error of \cmrnet2{} is significantly smaller than that of I2P-Net, Moreover, it is important to note that I2P-Net is trained on KITTI only and therefore more likely to overfit on this dataset. Additionally, \cmrnet2{} employs an iterative refinement approach, thus the localization error can be further decreased, surpassing the performance of I2P-Net already at the second iteration, as discussed in~\cref{sec_result_iterative}.

Notably, \cmrnet2{} is able to effectively generalize to different sensor setups and different cities than those used for training, while existing camera-agnostic methods struggle to do so. As previously mentioned, camera-dependent methods are tied to a specific sensor setup, and therefore, they cannot be evaluated on unseen datasets.
In particular, our method achieves a median translation error of \SI{16}{\centi\meter} and a median rotation error of \ang{0.42} on our in-house dataset, compared to \SI{84}{\centi\meter} and \ang{2.23} achieved by the second-best method.

\subsubsection{Iterative Refinement}
\label{sec_result_iterative}
As discussed in~\cref{sec:iterative}, the performance of our method can be further improved by using an iterative refinement strategy, where multiple instances of \cmrnet2{} are trained using different initial error ranges. More specifically, we train three instances (\cmrnet2{}-1, \cmrnet2{}-2, and \cmrnet2{}-3) with initial error ranges equal to [$\pm$\SI{2}{\metre}, $\pm$\SI{10}{\degree}], [$\pm$\SI{0.2}{\metre}, $\pm$\SI{0.5}{\degree}], and [$\pm$\SI{0.05}{\metre}, $\pm$\SI{0.1}{\degree}], respectively. The results reported in~\cref{tab:localization_lenc} show that the iterative refinement substantially improves the performance of \cmrnet2{} on all datasets, with an average reduction of \SI{37.85}{\percent} in translation error and \SI{14.36}{\percent} in rotation error.

We additionally compare the performance of \cmrnet2{} with camera-dependent and camera-agnostic methods. While HyperMap, BEVLoc, and I2P-Net do not provide iterative refinement, all the other methods use three instances trained with the following error ranges: [$\pm$\SI{2}{\metre}, $\pm$\SI{10}{\degree}], [$\pm$\SI{1}{\metre}, $\pm$\SI{2}{\degree}], and [$\pm$\SI{0.6}{\metre}, $\pm$\SI{2}{\degree}]. For a fair comparison, we additionally evaluate our method using the same error ranges used by the baselines, denoted as CMRNext (same range). For this experiment, we only report the results after the final iteration of all methods. Most baselines report these results only on the KITTI dataset. As shown in~\cref{tab:localization_lenc}, \cmrnet2{} (same range) outperforms all methods, setting the new state-of-the-art for monocular localization in LiDAR maps. By using our optimized initial error ranges in CMRNext (ours), the translation and rotation errors further decrease on most datasets. Specifically, \cmrnet2{} achieves a median translation error of \SI{6}{\centi\meter}, compared to the second-best method I2P-Net, which achieves a median translation error of \SI{7}{\centi\meter}. Moreover, \cmrnet2{} achieves a median rotation error of \ang{0.23}, compared to the second best method, I2D-Loc which achieves a median rotation error of \ang{0.30}.

In~\cref{fig:qualitative}, we present qualitative results of \cmrnet2{} for the monocular localization task on the KITTI, Argoverse, Pandaset, and Freiburg-Car datasets. The first column shows the LiDAR-map projection in the initial pose $H_{map}^{init}$ overlaid with the camera image, the second column shows the projection in the ground truth camera pose $H_{map}^{GT}$, and the third column shows the projection in the pose predicted by \cmrnet2{}. We observe that our method is able to accurately localize the camera in the LiDAR-map, even in challenging scenarios with dynamic objects that are not present in the map. The resulting LiDAR-map projections align well with the camera images.
Finally, in~\cref{fig:error_sequence}, we report the rotation and translation errors of CMRNext on the sequence 00 of the KITTI dataset, alongside images from three scenes where CMRNext achieved higher errors. We observed that the highest errors typically occur when the car is making a turn or when the environment is predominantly filled with trees and lacks structural elements, making feature matching particularly challenging.

\begin{table}
  \centering
  \addtolength{\tabcolsep}{-0.3em}
  \caption{ATE on the sequence 00 of the KITTI dataset.}
  \color{black}
  \label{tab:localization_comparison_slam}
  \begin{threeparttable}
    \begin{tabular}{clcccc}
    \toprule
     && \multicolumn{2}{c}{Translation [cm]} & \multicolumn{2}{c}{Rotation [°]} \\ \cmidrule(lr){3-4} \cmidrule(lr){5-6}
    && Mean & Std & Mean & Std \\ \midrule

   \multirow{5}{*}{\rotatebox[origin=c]{90}{\parbox[c]{1cm}{\centering\scriptsize {Stereo}}}}& S-PTAM~\cite{pire2017s} & 871 & 407 & 1.74 & 1.08 \\
   &VINS-Fusion\textsuperscript{\textdagger}~\cite{qin2018vins} (w/o LC) & \num{1670} & 940 & 4.99 & 2.51 \\
   &VINS-Fusion\textsuperscript{\textdagger}~\cite{qin2018vins} (w/ LC)  & \num{1355} & 710 & 2.23 & 3.02 \\
   &ORB-SLAM3\textsuperscript{\textdagger}~\cite{campos2021orb} (w/o LC) & 328 & 160 & 1.34 & 0.65 \\
   &ORB-SLAM3\textsuperscript{\textdagger}~\cite{campos2021orb} (w/ LC) & 198 & 89 & 1.32 & 0.67 \\ \midrule

   \multirow{4}{*}{\rotatebox[origin=c]{90}{\parbox[c]{1cm}{\centering\scriptsize {Stereo + Map}}}}& Kim \textit{et al.}~\cite{kim2018stereo} & \underline{13} & 11 &  \underline{0.32} & 0.39 \\
    & Zuo \textit{et al.}\cite{zuo2020multimodal} & 47 & -- & 0.87 & --\\ 
    & Ding \textit{et al.}~\cite{ding2019persistent} & 45 & -- & -- & -- \\
    & SemLoc~\cite{liang2022semloc} & 34 & -- & -- & -- \\ \midrule
   
   \multirow{10}{*}{\rotatebox[origin=c]{90}{\parbox[c]{2cm}{\centering\scriptsize {Monocular + Map}}}}&Caselitz \textit{et al.}~\cite{Caselitz_2016} & 30 & 11 & 1.65 & 0.91 \\
   & Sun \textit{et al.}~\cite{sun2019scale} & \textbf{11} & \underline{8} & 1.42 & 1.09 \\
   & Zhang \textit{et al.}~\cite{zhang2023cross} & 58 & -- & -- & -- \\
   & SemLoc~\cite{liang2022semloc} & 149 & -- & -- & -- \\
   &CMRNet\textsuperscript{\textdagger}~\cite{Cattaneo_2019} & \num{16358} & \num{7159} & 126.02 & 36.67 \\
   &I2D-Loc\textsuperscript{\textdagger}~\cite{chen2022i2d} & \num{13142} & \num{8172} & 57.76 & 52.41 \\
   &I2D-VO\textsuperscript{\textdagger}~\cite{yu2024i2d} & 24 & 26 & 0.54 & 0.94 \\
   &I2D-Loc++\textsuperscript{\textdagger}~\cite{yu2024i2d} & \underline{13} & 17 & 0.49 & 0.55 \\
   & CMRNet++~\cite{cattaneo2020cmrnet} & 74 & 35 & 0.42 & \underline{0.25} \\
   & \textbf{\cmrnet2{} (ours)} & \textbf{11} & \textbf{6} & \textbf{0.25} & \textbf{0.15} \\
    \bottomrule
  \end{tabular}
  \begin{tablenotes}[para,flushleft]
       \footnotesize LC represents whether the loop closure detection module was enabled (w/) or disabled (w/o). \textdagger: Results for these methods are taken from~\cite{yu2024i2d}.
     \end{tablenotes}
   \end{threeparttable}
\end{table}

\begin{figure*}[t]
    \centering

    \begin{subfigure}[b]{0.4\textwidth}
         \centering
         \includegraphics[width=\textwidth]{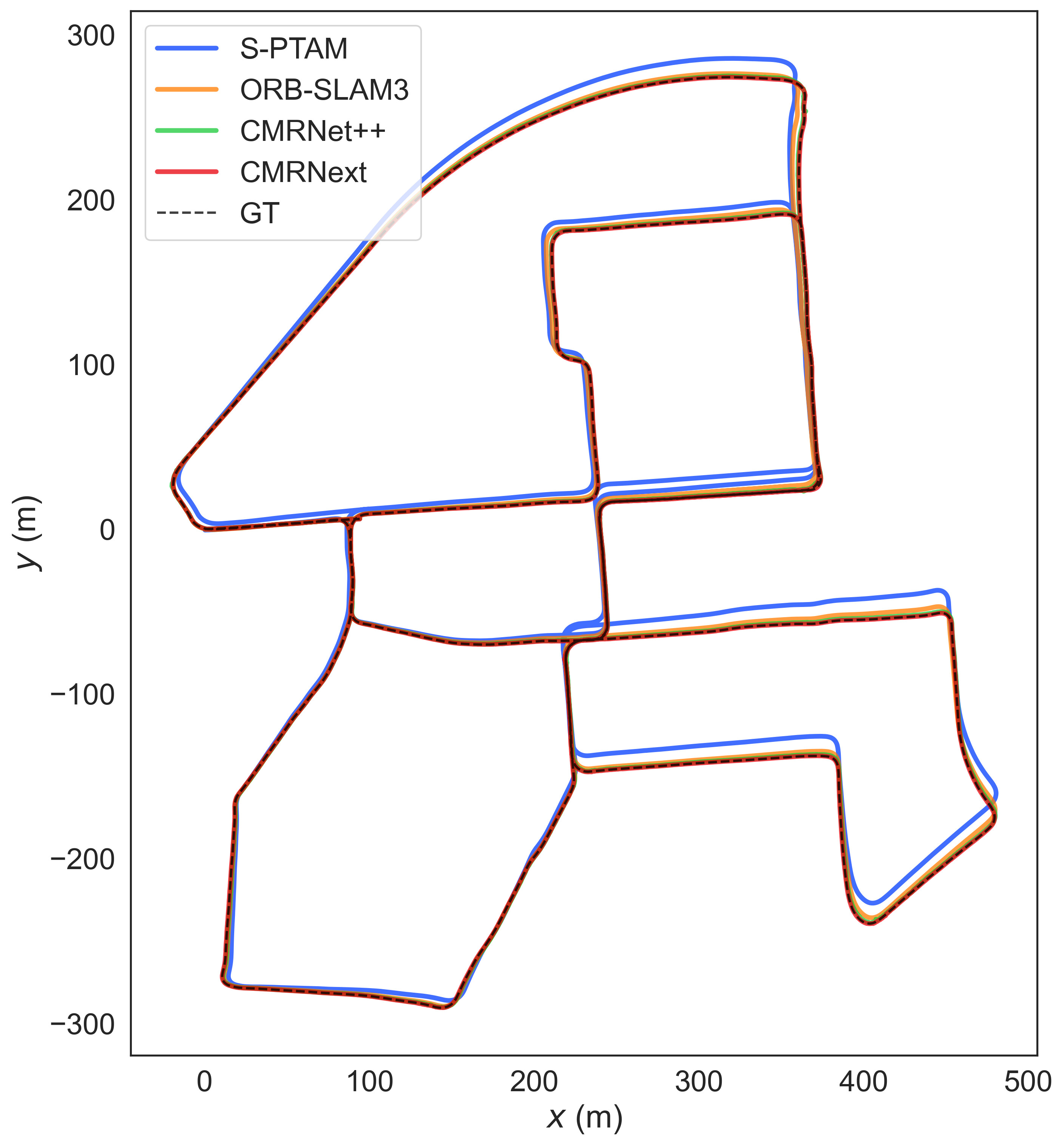}
     \end{subfigure}
     \hspace{3em}
     \begin{subfigure}[b]{0.43\textwidth}
         \centering
         \includegraphics[width=\textwidth]{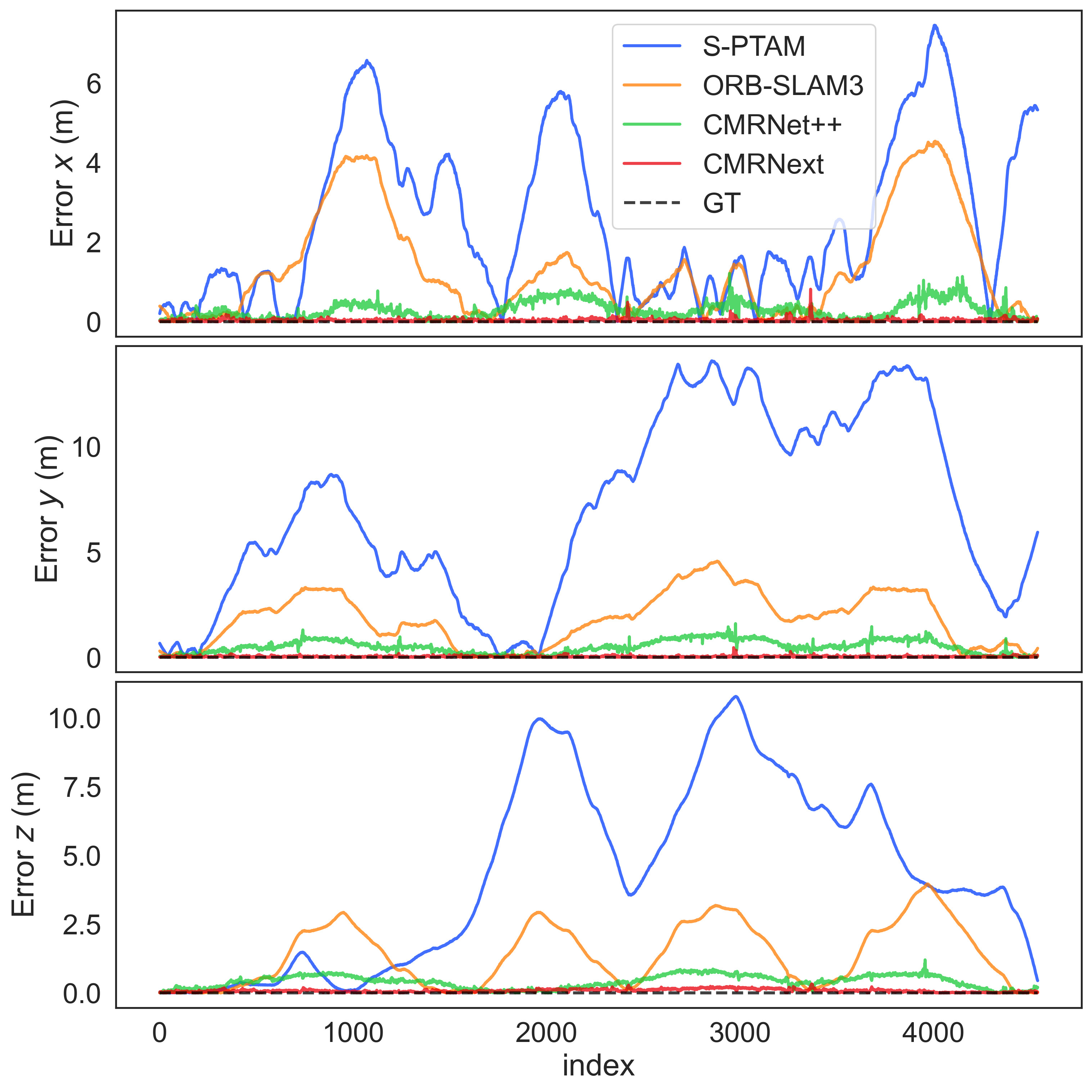}
     \end{subfigure}
     
    \caption{\color{black}Predicted path (left) and translation errors (right) of different methods on the full trajectory localization experiment.}
    \label{fig:odom_path}
\end{figure*}

{\color{black}
\subsection{Full Trajectory Localization}
The experiments reported in~\cref{sec:monocular_loc} assume that an initial pose $H_{map}^{init}$ is provided for every frame of the sequence. However, in real-world scenarios, this initial pose might not always be available, such as in \ac{GNSS}-denied environments. We therefore perform a trajectory-level evaluation, where we provide an initial pose only for the first frame of the trajectory, and then we initialize the successive frames using the prediction of the previous one. This evaluation protocol includes the unique challenge that the drift accumulated over multiple frames, if not corrected, might yield LiDAR-images that have no overlap with the camera images, making the localization infeasible. We compare the performance of \cmrnet2{} with tradition methods for stereo SLAM~\cite{pire2017s,qin2018vins,campos2021orb}, stereo localization in LiDAR maps~\cite{kim2018stereo,zuo2020multimodal,ding2019persistent,liang2022semloc}, monocular localization in LiDAR maps~\cite{Caselitz_2016,sun2019scale,zhang2023cross,liang2022semloc}, as well as learning-based methods~\cite{Cattaneo_2019,chen2022i2d,yu2024i2d,cattaneo2020cmrnet}. We compare all methods using the standard Absolute Trajectory Error (ATE) metric typically employed to evaluate odometry and SLAM approaches. The results reported in~\cref{tab:localization_comparison_slam} show that our \cmrnet2{} outperforms all existing methods achieving a mean ATE of \SI{11}{\centi\metre} and \SI{0.25}{\degree} for the translation and rotation components, respectively. Previous methods CMRNet and I2D-Loc suffer from large drift that results in unrecoverable localization failure. I2D-Loc++ explicitly tackles this problem by employing optical flow between consecutive images and multi-frame bundle adjustment optimization, drastically reducing the localization errors. While our method does not employ any tracking or multi-frame processing, it outperforms I2D-Loc++ and notably achieves the lowest standard deviations among all methods due to its robust camera-LiDAR matching capability. In~\cref{fig:odom_path}, we report the predicted path and the translation errors of CMRNext compared with S-PTAM, ORB-SLAM3, and CMRNet++ on the sequence 00 of the KITTI dataset.}

\subsection{Ablation Study}

We performed an extensive ablation study to evaluate the impact of the different design choices in our proposed method. To avoid computational overhead, all experiments reported in this section were performed only on the first iteration of \cmrnet2{} and evaluated on the validation set of the KITTI, Argoverse, and Pandaset datasets.

In~\cref{tab:localization_ablation_backbone} we evaluate the performance of \cmrnet2{} using different optical flow networks. In particular, we evaluate established optical flow networks, namely PWCNet~\cite{Sun_2018_CVPR}, MaskFlowNet~\cite{zhao2020maskflownet} and RAFT~\cite{teed2020raft}, as well as the recent state-of-the-art methods GMFlow~\cite{xu2023unifying} and FlowFormer~\cite{huang2022flowformer}. Unsurprisingly, RAFT outperforms previous methods PWCNet and MaskFlowNet by a large margin. However, we noticed that both transformer-based methods GMFlow and FlowFormer underperformed compared to RAFT, although they achieved better results on optical flow benchmarks. We hypothesize that this is due to the fact that networks based on Vision Transformers (ViT) lack intrinsic inductive bias~\cite{xu2021vitae} and thus require a large amount of data to train. The vast majority of methods for optical flow estimation are pretrained on large-scale synthetic datasets, such as FlyingChairs and FlyingThings, and therefore ViT-based methods are able to learn this inductive bias. For our cross-modal task, however, such large-scale datasets are not available, and therefore, the CNN-based method RAFT achieves the best performance among all the evaluated networks. Moreover, the positional encoding used in transformers is sensitive to the shape of the input, and during inference, images with the same resolution as the ones used during training should be used for optimal performance. The authors of FlowFormer~\cite{huang2022flowformer} suggested a tiling strategy where the input image is split into multiple tiles, each tile is then independently processed by the network, and the resulting optical flow fields are combined together. While this strategy substantially improves the performance of FlowFormer (see~\cref{tab:localization_ablation_backbone}), it requires multiple forward passes of the network, drastically increasing the computational cost.

\begin{table*}
  \centering
  \caption{Ablation study on the optical flow network architecture.}
  \label{tab:localization_ablation_backbone}
  \begin{threeparttable}
    \begin{tabular}{lp{0.2cm}cccccc}
    \toprule
     && \multicolumn{2}{c}{KITTI} & \multicolumn{2}{c}{Argoverse} & \multicolumn{2}{c}{Pandaset} \\ \cmidrule(lr){3-4} \cmidrule(lr){5-6} \cmidrule(lr){7-8} 
    && Transl. [cm] & Rot. [°] & Transl. [cm] & Rot. [°] & Transl. [cm] & Rot. [°] \\ \midrule

    PWCNet~\cite{Sun_2018_CVPR} && 44.02 & 1.13 & 60.94  & 1.13  & 77.68 & 1.01 \\
    MaskFlowNet-S~\cite{zhao2020maskflownet} && 23.18 & 0.57 & 28.86 & 0.38 & 41.28 & 0.44 \\
    GMFlow~\cite{xu2023unifying} && 27.06 & 0.58 & 87.65 & 1.45 & 61.38 & 0.74 \\
    GMFlow (refine)~\cite{xu2023unifying} && 11.22 & \underline{0.30} & 33.04 & 0.44 & 22.19 & 0.24 \\ 
    FlowFormer~\cite{huang2022flowformer} && 15.25  & 0.36  & 81.17  & 1.02 & 30.37 & 0.28 \\
    FlowFormer (tiled)~\cite{huang2022flowformer} && \underline{10.34} & \underline{0.30} & \underline{16.04} & \underline{0.24} & \underline{17.22} & \underline{0.20} \\
    RAFT~\cite{teed2020raft} && \textbf{9.55} & \textbf{0.29} & \textbf{12.72} & \textbf{0.20} & \textbf{12.43} & \textbf{0.15} \\
    
    \bottomrule
  \end{tabular}
   \end{threeparttable}
\end{table*}

\begin{table*}
  \centering
  \caption{Ablation studies on different components in \cmrnet2{}.}
  \label{tab:localization_ablation}
  \begin{threeparttable}
    \begin{tabular}{ccccp{0.1cm}cccccc}
    \toprule
    \multirow{3}{*}{\shortstack{Positional\\Encoding}} & \multirow{3}{*}{\shortstack{Occlusion\\Filter}} & \multirow{3}{*}{\shortstack{Voxel\\Size [m]}} & \multirow{3}{*}{\shortstack{Context\\Encoder}} && \multicolumn{2}{c}{KITTI} & \multicolumn{2}{c}{Argoverse} & \multicolumn{2}{c}{Pandaset} \\ \cmidrule(lr){6-7} \cmidrule(lr){8-9} \cmidrule(lr){10-11} 
    &&&&& Transl. [cm] & Rot. [°] & Transl. [cm] & Rot. [°] & Transl. [cm] & Rot. [°] \\ \midrule

    \tabnode{\ding{55}} & \multirow{3}{*}{\ding{51}} & \multirow{3}{*}{0.1} & \multirow{3}{*}{RGB} && 12.83 & 0.33 & 23.50 & 0.38 & 24.35 & 0.27 \\
    \tabnode{$m$=6} & &&&& 11.27 & 0.31 & 28.82 & 0.41 & 22.03 & 0.24 \\
    \tabnode{$m$=12} &&& && \textbf{11.02} & \textbf{0.30} & \textbf{23.49} & \textbf{0.30} & \textbf{19.91} & \textbf{0.23} \\
    \midrule
    \multirow{2}{*}{$m$=12} & \tabnode{\ding{55}} &  \multirow{2}{*}{0.1} & \multirow{2}{*}{RGB} && 12.25 & 0.33 & 38.83 & 0.66 & 37.77 & 0.54 \\
    & \tabnode{\ding{51}} & & && \textbf{11.02} & \textbf{0.30} & \textbf{23.49} & \textbf{0.30} & \textbf{19.91} & \textbf{0.23} \\
    \midrule
    \multirow{3}{*}{$m$=12} & \multirow{3}{*}{\ding{51}} & \tabnode{0.5} & \multirow{3}{*}{RGB} && 33.51 & 0.64 & 81.94 & 0.76 & 41.98 & 0.34 \\
    && \tabnode{0.2} &&& 20.15 & 0.43 & 68.86 & 0.66 & 43.67 & 0.41 \\
    && \tabnode{0.1} &&& \textbf{11.02} & \textbf{0.30} & \textbf{23.49} & \textbf{0.30} & \textbf{19.91} & \textbf{0.23} \\
    \midrule
    \multirow{2}{*}{$m$=12} & \multirow{2}{*}{\ding{51}} & \multirow{2}{*}{0.1} & \tabnode{RGB} && {11.02} & {0.30} & {23.49} & {0.30} & {19.91} & {0.23} \\
     &&& \tabnode{LiDAR} && \textbf{9.55} & \textbf{0.29} & \textbf{12.72} & \textbf{0.20} & \textbf{12.43} & \textbf{0.15} \\
    
    \bottomrule
  \end{tabular}
  \begin{tablenotes}[para,flushleft]
       \footnotesize
       Yellow boxes indicate the specific component that is being ablated in each set of rows.
     \end{tablenotes}
   \end{threeparttable}
  \begin{tikzpicture}[overlay]
    \draw [orange] (1.north west) -- (1.north east) -- (3.south east) --
    (3.south west) -- cycle;
    \draw [orange] (4.north west) -- (4.north east) -- (5.south east) --
    (5.south west) -- cycle;
    \draw [orange] (6.north west) -- (6.north east) -- (8.south east) --
    (8.south west) -- cycle;
    \draw [orange] (9.north west) -- (9.north east) -- (10.south east) --
    (10.south west) -- cycle;

  \end{tikzpicture}
\end{table*}

In~\cref{tab:localization_ablation}, we additionally investigate the impact of the positional encoding presented in~\cref{eq:pos_encoding}, the occlusion filter discussed in~\cref{sec:matching}, the voxel size used to discretize the map, and which input to use for the context encoder (camera image or LiDAR image). We first evaluate the effect of the positional encoding by comparing the performance of \cmrnet2{} using no positional encoding, the positional encoding presented in~\cref{eq:pos_encoding} with six Fourier frequencies, and with 12 frequencies. The first set of rows in~\cref{tab:localization_ablation} shows that the positional encoding substantially improves the performance of \cmrnet2{}, with the best results achieved using 12 frequencies, demonstrating that Fourier features are better suited for deep neural network processing than (possibly normalized) coordinates.

Next, in the second set of rows, we verified that the occlusion filter considerably reduces the localization error of \cmrnet2{}, especially on the Argoverse and Pandaset datasets. The camera images of these datasets have a larger field of view and higher resolution than those of the KITTI dataset, and thus, occluded map points are more likely to be projected in the LiDAR images if no occlusion filter is applied, negatively impacting the performance.

Subsequently, we evaluated three different resolutions for the LiDAR maps: \SI{0.5}{\metre}, \SI{0.2}{\metre}, and \SI{0.1}{\metre}. As shown in the third set of rows in~\cref{tab:localization_ablation}, the voxel size has a significant impact on the performance of our method. This result is not surprising, as a coarse resolution would lead to sparse LiDAR-images with non-well-defined edges, hindering the pixel-wise matching with the camera images. Although a resolution lower than \SI{0.1}{\metre} could further improve the performance, computational and storage requirements would increase significantly. Therefore, we chose \SI{0.1}{\metre} as the default resolution for the LiDAR maps, as a good trade-off between performance and computational cost.

Finally, we compared the performance of \cmrnet2{} using the camera image and the LiDAR image as the input for the context encoder. The last two rows of~\cref{tab:localization_ablation} show that the LiDAR context encoder achieves better performance than the camera context encoder.

\subsection{LiDAR-Camera Calibration}

\begin{figure*}
  \centering
      \includegraphics[width=.99\textwidth]{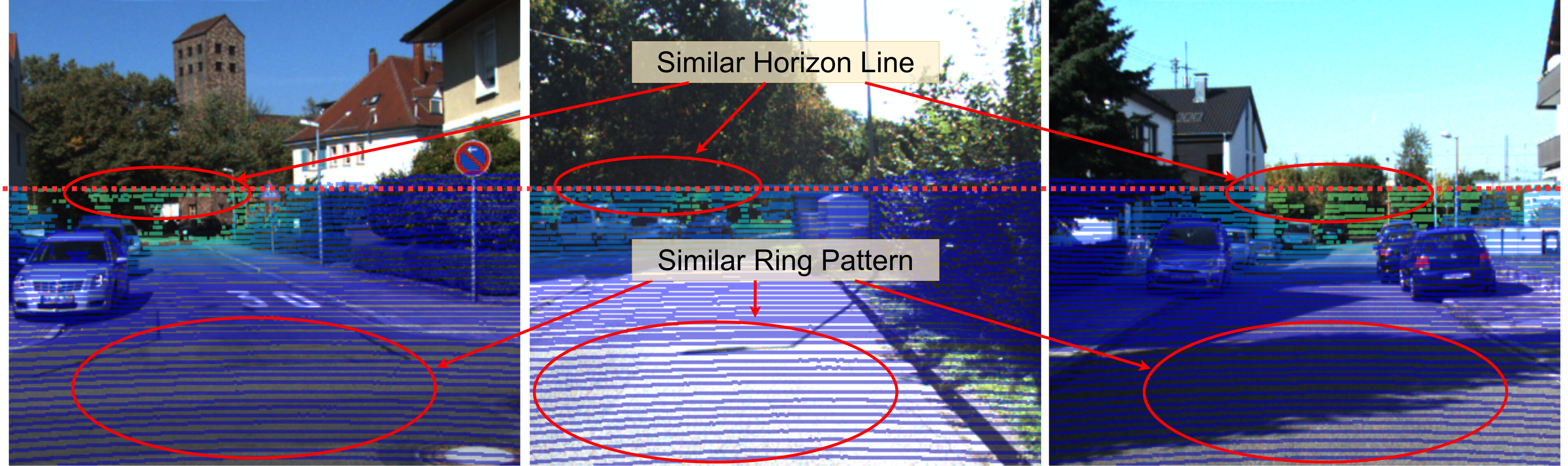}
      \caption{In the LiDAR-camera calibration task, the projection of the LiDAR scan in the camera frame using the ground truth calibration generates recognizable patterns for a specific LiDAR-camera pair. A network trained on a single sensor setup might learn to recognize these patterns, overfitting to the specific sensor setup used for training.}
      \label{fig:extrinsic_similar}
      \vspace{-0.3cm}
  \end{figure*}

We extended our method to the task of LiDAR-camera extrinsic calibration. Although this task shares some similarities with monocular localization in LiDAR maps, there are some key differences. First, in the calibration task, the camera image is matched against a single LiDAR scan recorded at the same time, therefore occlusions and dynamic objects do not hinder the matching process, as both sensors observe the same scene. For this reason, we did not use the occlusion filter for all the experiments reported in this section. Second, for a specific LiDAR-camera pair, the projection of the LiDAR points in the camera frame using the ground truth extrinsic calibration would contain some recognizable pattern induced by the LiDAR sensor specification and placement. For example, the horizon line of the LiDAR scan would be projected on the same row of pixels in the camera image, and the typical ring pattern of mechanical LiDAR sensors would appear very similar on all images, as illustrated in~\cref{fig:extrinsic_similar}. Networks that are trained on a single LiDAR-camera pair could learn to exploit these patterns and thus overfit to the specific extrinsic calibration of the training dataset. This is especially true for camera-dependent methods, as the only way to include multiple extrinsic calibrations without changing the camera is to physically move the sensors, drastically increasing the effort required to collect a training dataset. Most of the existing \ac{DNN}-based methods for LiDAR-camera calibration~\cite{lv2021lccnet,wang2023end,Schneider_2017,8593693,wu2022psnet} are trained and evaluated on the same extrinsic calibration. We believe that this evaluation protocol is misleading, as a network that always predicts the same extrinsic parameters cannot be used in any practical application. Therefore, we evaluate all methods on LiDAR-camera pairs that were not included in the training set. In particular, we used the right camera of the KITTI odometry dataset for evaluation, while only the left camera was included in the training.

\begin{table}
  \centering
  \caption{Comparison of LiDAR-camera extrinsic calibration on the KITTI odometry dataset.}
  \label{tab:calibration_comparison}
  \begin{threeparttable}
    \begin{tabular}{lcccc}
    \toprule
     & \multicolumn{2}{c}{KITTI - left} & \multicolumn{2}{c}{KITTI - right} \\ 
     \cmidrule(lr){2-3} \cmidrule(lr){4-5}
    & Transl.[cm] & Rot.[°] & Transl.[cm] & Rot.[°]\\ \midrule

    CMRNet~\cite{Cattaneo_2019} & 1.57 & 0.10 & 52.92 & 1.49 \\
    LCCNet~\cite{lv2021lccnet} & \textbf{1.01} & 0.12 & 52.51 & 1.47 \\
    \textbf{\cmrnet2{} (ours)} & 1.89 & \textbf{0.08} & \textbf{7.07} & \textbf{0.23} \\
    
    \bottomrule
  \end{tabular}
  \begin{tablenotes}[para,flushleft]
       \footnotesize
       The left camera was included in the training set of all methods, while the right camera was never seen during training.
     \end{tablenotes}
   \end{threeparttable}
\end{table}

We compared the performance of \cmrnet2{} on the LiDAR-camera extrinsic calibration task with our previous methods CMRNet~\cite{Cattaneo_2019}, as well as with the state-of-the-art methods RGGNet~\cite{Yuan2020rggnet} and LCCNet~\cite{lv2021lccnet}. For the latter methods, we used the pretrained models provided by the respective authors\footnote{\url{https://github.com/LvXudong-HIT/LCCNet}\\\url{https://github.com/KleinYuan/RGGNet}}. Other existing methods~\cite{wang2023end,8593693,jing2022dxq,wu2022psnet} do not provide a publicly available implementation and therefore we were unable to evaluate them on the right camera of the KITTI dataset, as discussed above. In~\cref{tab:calibration_comparison}, we report the results of CMRNet, LCCNet, and \cmrnet2{} after the iterative refinement (five iterations for CMRNet and LCCNet, and three iterations for \cmrnet2{}). The results show that although LCCNet achieves the lowest translation error when evaluated on the same camera used for training, it fails to generalize to a different camera, still predicting the extrinsic for the left camera when evaluated on the right one. We observe a similar behavior for our previous work CMRNet, as both CMRNet and LCCNet are camera-dependent and thus more prone to overfitting to the specific extrinsic. Our proposed \cmrnet2{}, on the other hand, outperforms existing methods on the right camera, with a median translation error of \SI{7}{\centi\metre} and a median rotation error of \ang{0.23}, while at the same time achieving the lowest rotation error and comparable translation error on the left camera.

\begin{table}
  \centering
  \caption{Comparison of LiDAR-camera extrinsic calibration on the KITTI raw dataset.}
  \label{tab:rgg_comparison_vert}
  \begin{threeparttable}
    \begin{tabular}{clccc}
    \toprule
     && $\beta$-RegNet~\cite{Schneider_2017} & RGGNet+~\cite{Yuan2020rggnet} & \textbf{\cmrnet2{}} \\ \midrule

     \multirow{4}{*}{\rotatebox[origin=c]{90}{\parbox[c]{1cm}{\centering\scriptsize T1-Left}}} & Transl. $\downarrow$ & - & 11.49 & \textbf{4.19} \\ 
     & Rot. $\downarrow$ & - & 1.29 & \textbf{0.14} \\
     & MSEE $\downarrow$ & 0.048 &  0.021 & \textbf{0.006} \\
     & MRR $\uparrow$ & 52.23\% & 79.02\% & \textbf{95.18\%} \\ \midrule

     \multirow{4}{*}{\rotatebox[origin=c]{90}{\parbox[c]{1cm}{\centering\scriptsize T1-Right}}} & Transl. $\downarrow$ & - & 23.52 & \textbf{5.76} \\ 
     & Rot. $\downarrow$ & - & 3.87 & \textbf{0.19} \\
     & MSEE $\downarrow$ & - &  0.079 & \textbf{0.011} \\
     & MRR $\uparrow$ & - & 12.37\% & \textbf{90.13\%} \\ \midrule

     \multirow{4}{*}{\rotatebox[origin=c]{90}{\parbox[c]{1cm}{\centering\scriptsize T3-Left}}} & Transl. $\downarrow$ & - & 8.52 & \textbf{4.36} \\ 
     & Rot. $\downarrow$ & - & 0.80 & \textbf{0.18} \\
     & MSEE $\downarrow$ & 0.092 & 0.001 & \textbf{0.004} \\
     & MRR $\uparrow$ & -1.89\% & 80.90\% & \textbf{94.39\%} \\ \midrule

     \multirow{4}{*}{\rotatebox[origin=c]{90}{\parbox[c]{1cm}{\centering\scriptsize T3-Right}}} & Transl. $\downarrow$ & - & 31.31 & \textbf{5.61} \\ 
     & Rot. $\downarrow$ & - & 2.12 & \textbf{0.21} \\
     & MSEE $\downarrow$ & - & 0.133 & \textbf{0.006} \\
     & MRR $\uparrow$ & - & -113.67\% & \textbf{89.71\%} \\
    \bottomrule
  \end{tabular}
   \end{threeparttable}
\end{table}

RGGNet~\cite{Yuan2020rggnet} was trained on raw sequences of the KITTI dataset, instead of the odometry sequences and with an initial error up to [$\pm$ \SI{0.3}{\metre}, $\pm$ \ang{15}]. To provide a fair comparison with RGGNet, we trained a single instance of \cmrnet2{} without iterative refinement using the same training set used by RGGNet and we evaluated on both left and right cameras of the T1 and T3 testing datasets defined in~\cite{Yuan2020rggnet}. The testing datasets T2a and T2b contain sequences that were included in the training set and therefore do not provide a fair evaluation, hence we excluded them from the evaluation. $\beta$-RegNet is the reimplementation of RegNet~\cite{Schneider_2017} by the authors of RGGNet. In addition to the median rotation and translation error, following~\cite{Yuan2020rggnet} for this experiment, we also report the mean $se3$ error (MSEE) and mean re-calibration rate (MRR), defined as:
\begin{eqnarray}
    MSEE = \frac{1}{n} \sum_i^n E_i, \\
    MRR = \frac{1}{n} \sum_i^n \left| \frac{\eta_i - E_i}{\eta_i} \right|,
\end{eqnarray}
where $\eta_i$ is the initial error applied to the $i$-th sample, and $E_i$ is the left-invariant Riemannian metric on the SE(3) Lie group between the ground truth and the predicted extrinsic calibration. The results reported in~\cref{tab:rgg_comparison_vert} show that \cmrnet2{} outperforms both $\beta$-RegNet and RGGNet on all testing splits by a large margin. When evaluated on the right camera, \cmrnet2{} achieves an MMR of 90.13\% on the T1 testing split, compared to 12.37\% achieved by RGGNet and 89.71\% MRR on the T3 split, compared to -113.67\% achieved by RGGNet.

\begin{table}
  \centering
  \caption{Extrinsic calibration results of \cmrnet2{} using temporal aggregation on the KITTI odometry dataset.}
  \label{tab:calibration_temporal}
  \begin{threeparttable}
    \begin{tabular}{clcccc}
    \toprule
     && \multicolumn{2}{c}{KITTI - left} & \multicolumn{2}{c}{KITTI - right} \\ 
     \cmidrule(lr){3-4} \cmidrule(lr){5-6}
    n & Measure & Transl. [cm] & Rot. [°] & Transl. [cm] & Rot. [°]\\ \midrule

    \multirow{3}{*}{\rotatebox[origin=c]{90}{\parbox[c]{1cm}{\centering\scriptsize ALL}}} & Mean & 2.04 & 0.04 & 10.94 & 0.28 \\
    & Median & \textbf{0.73} & - & 5.42 & - \\
    & Mode & 0.75 & \textbf{0.03} & \textbf{1.98} & \textbf{0.05}\\ 
    \midrule
    \multirow{3}{*}{\rotatebox[origin=c]{90}{\parbox[c]{1cm}{\centering\scriptsize 2000}}} & Mean & 2.06 & 0.05 & 10.82 & 0.28 \\
    & Median & \textbf{0.78 }& - & 5.45 & - \\
    & Mode & 1.19 & \textbf{0.04} & \textbf{1.99} & \textbf{0.10} \\
    \midrule
    \multirow{3}{*}{\rotatebox[origin=c]{90}{\parbox[c]{1cm}{\centering\scriptsize 1000}}} & Mean & 2.05 & 0.06 & 10.81 & 0.28 \\
    & Median & \textbf{0.79} & - & 5.50 & - \\
    & Mode & 1.21 & \textbf{0.05} & \textbf{2.05} & \textbf{0.11} \\
    \midrule
    \multirow{3}{*}{\rotatebox[origin=c]{90}{\parbox[c]{1cm}{\centering\scriptsize 500}}} & Mean & 2.16 & 0.08 & 10.93 & 0.28 \\
    & Median & \textbf{0.83} & - & 5.55 & - \\
    & Mode & 1.21 & \textbf{0.05} & \textbf{2.00} & \textbf{0.09} \\
    \midrule
    \multirow{3}{*}{\rotatebox[origin=c]{90}{\parbox[c]{1cm}{\centering\scriptsize 250}}} & Mean & 2.40 & 0.10 & 11.04 & 0.29 \\
    & Median & \textbf{0.88 }& - & 5.67 & - \\
    & Mode & 1.26 & \textbf{0.06} & \textbf{2.28} & \textbf{0.11} \\
    \midrule
    \multirow{3}{*}{\rotatebox[origin=c]{90}{\parbox[c]{1cm}{\centering\scriptsize 100}}} & Mean & 3.05 & 0.13 & 11.23 & 0.32 \\
    & Median & \textbf{0.96} & - & 6.17 & - \\
    & Mode & 1.39 & \textbf{0.07} & \textbf{3.84} & \textbf{0.13} \\
    
    \bottomrule
  \end{tabular}
   \end{threeparttable}
\end{table}

\begin{figure*}
  \centering
  \footnotesize
  \setlength{\tabcolsep}{0.05cm}
      {\renewcommand{\arraystretch}{1}
        \begin{tabular}{p{5.9cm}p{5.9cm}p{5.9cm}}

        \multicolumn{1}{c}{Initial Pose} & \multicolumn{1}{c}{Ground Truth} & \multicolumn{1}{c}{\cmrnet2{} (ours)} \\

        \includegraphics[width=\linewidth]{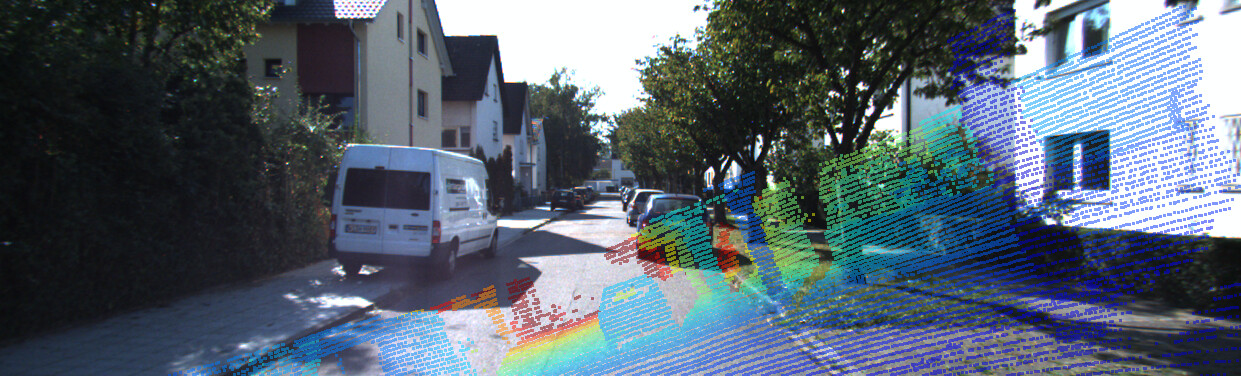} &
        \includegraphics[width=\linewidth]{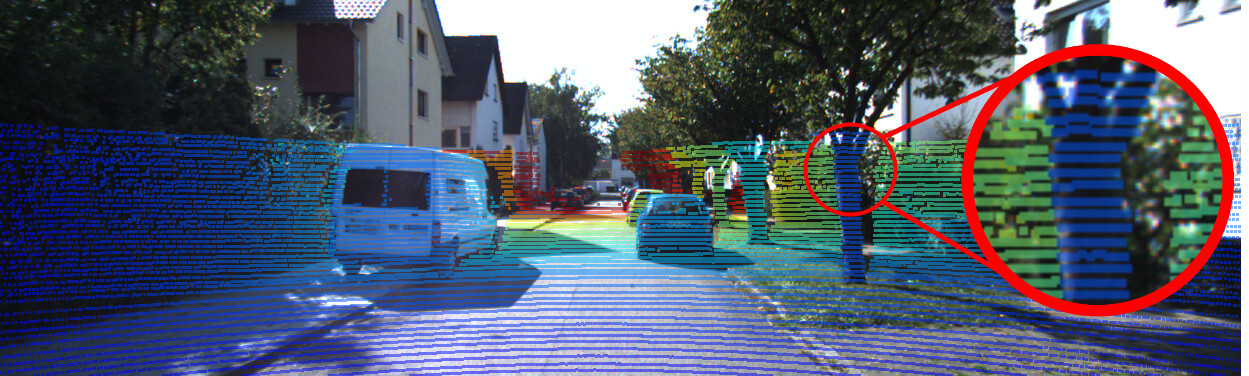} &
        \includegraphics[width=\linewidth]{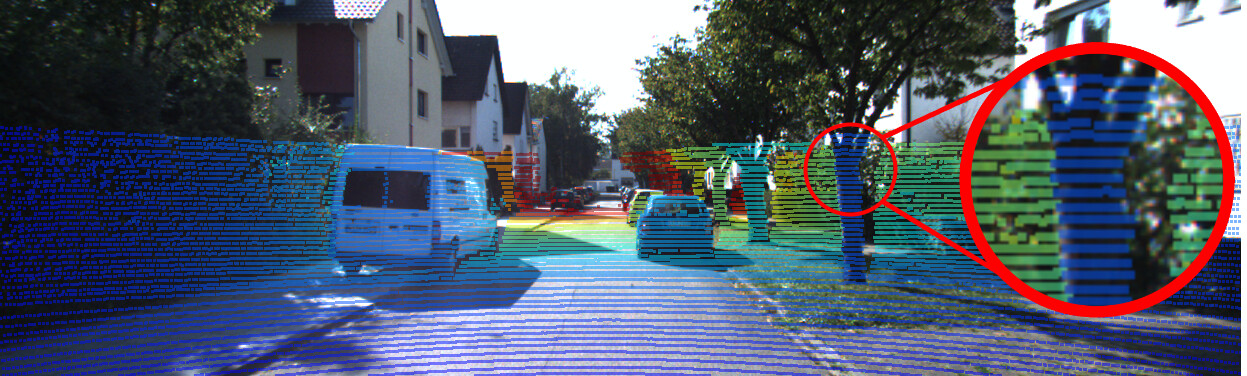} \\

      \includegraphics[width=\linewidth,trim={0 5cm 0 5cm},clip]{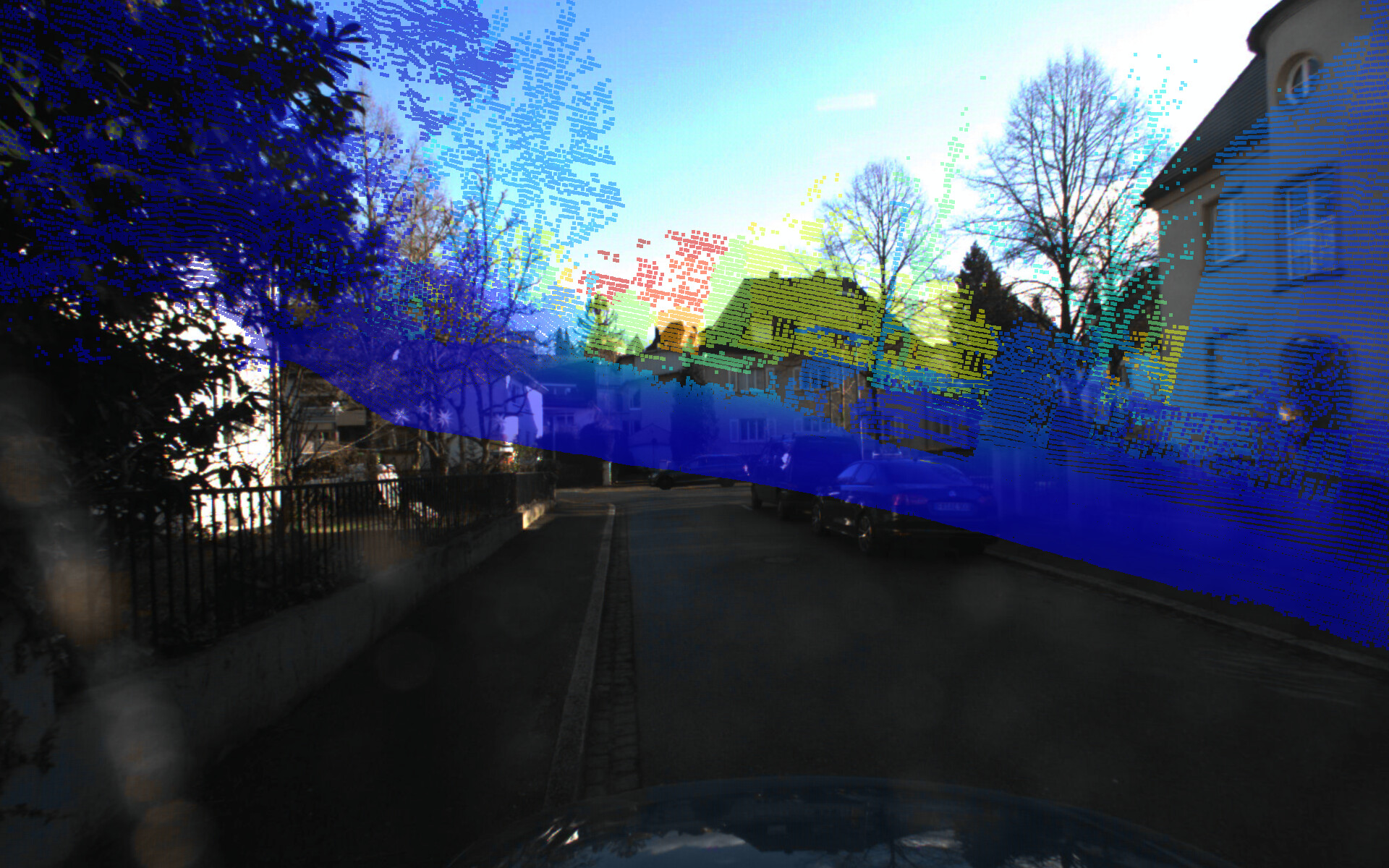} &
      \includegraphics[width=\linewidth,trim={0 5cm 0 5cm},clip]{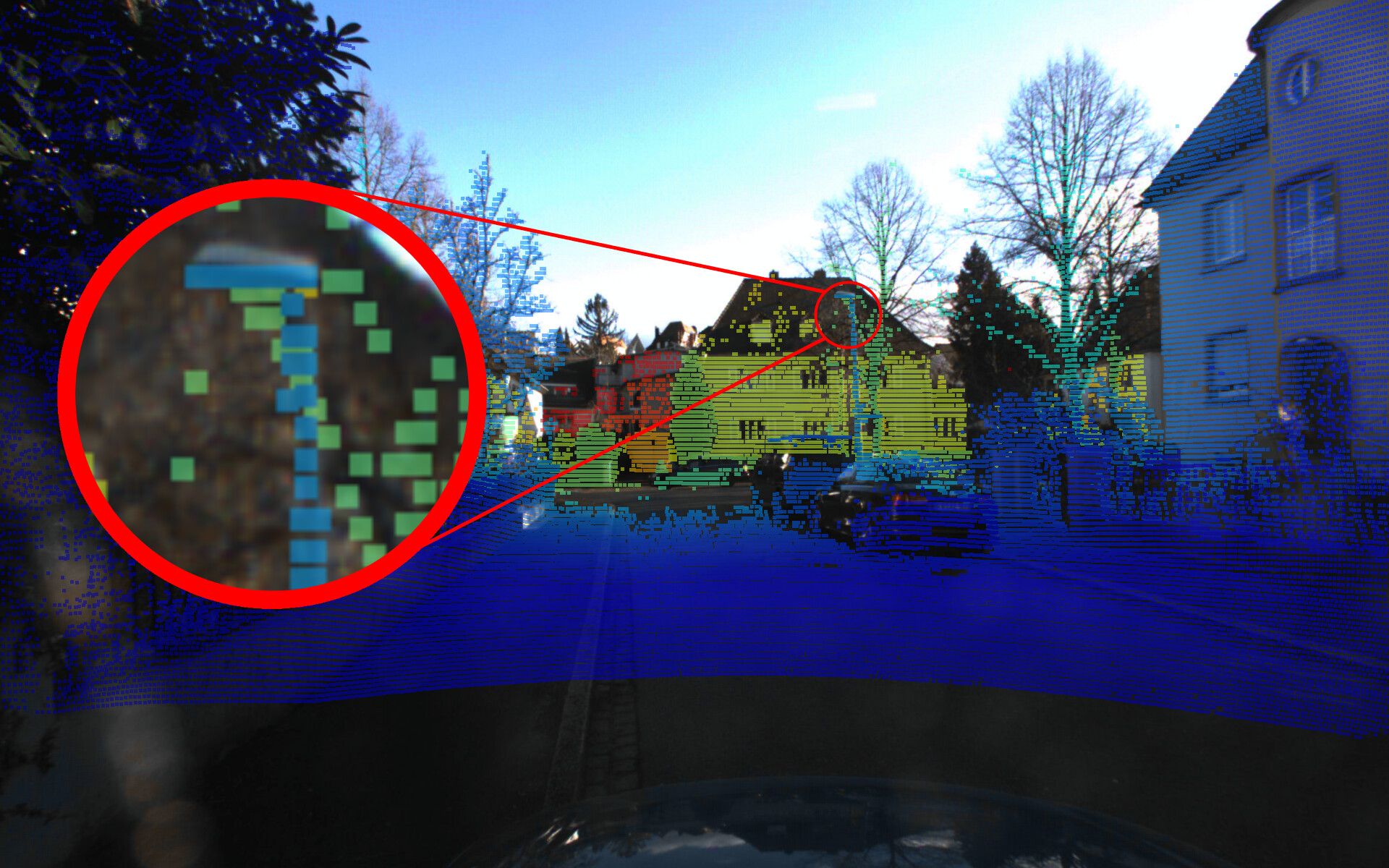} &
      \includegraphics[width=\linewidth,trim={0 5cm 0 5cm},clip]{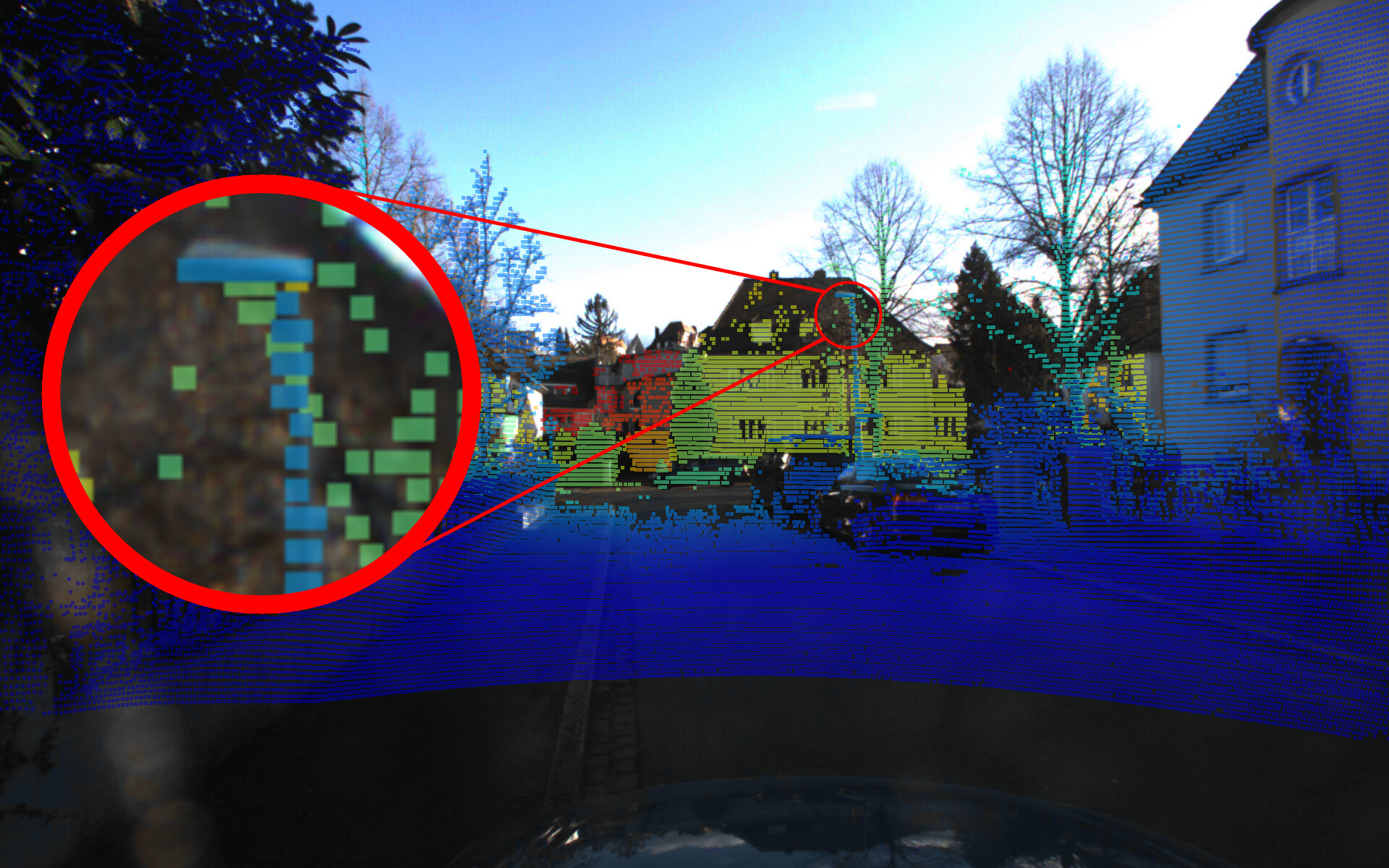} \\

      \includegraphics[width=\linewidth,trim={0 5cm 0 5cm},clip]{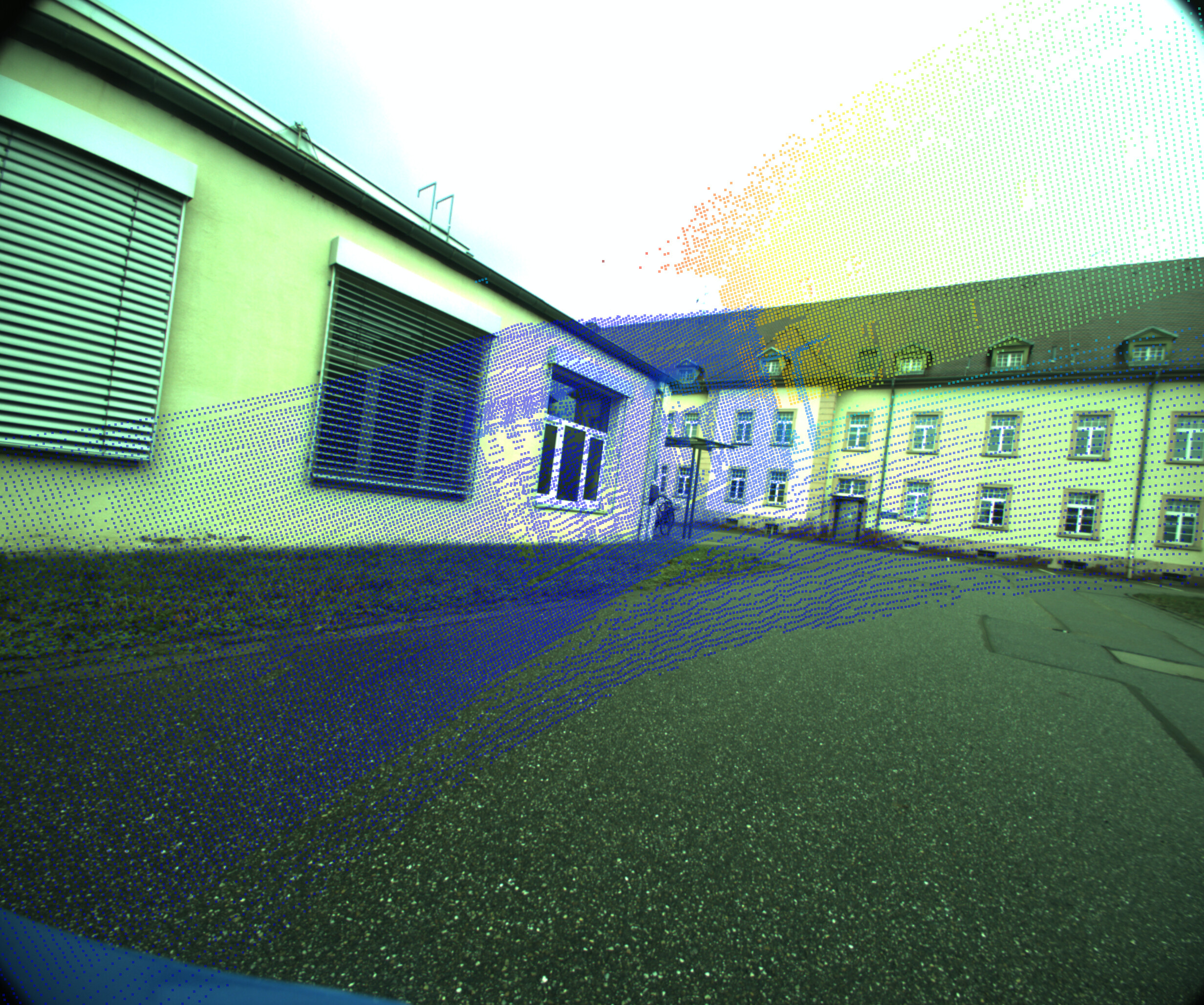} &
      \includegraphics[width=\linewidth,trim={0 5cm 0 5cm},clip]{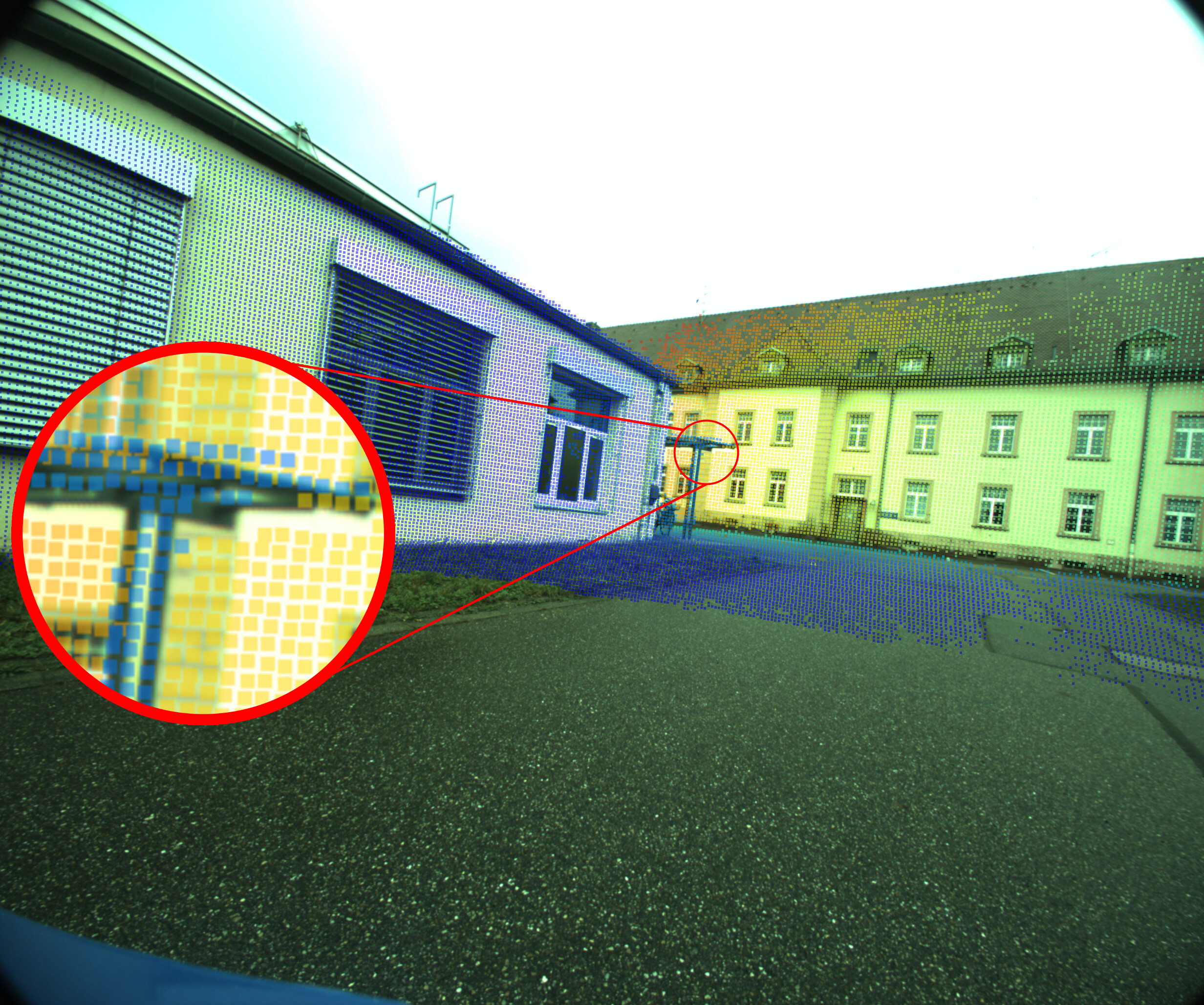} &
      \includegraphics[width=\linewidth,trim={0 5cm 0 5cm},clip]{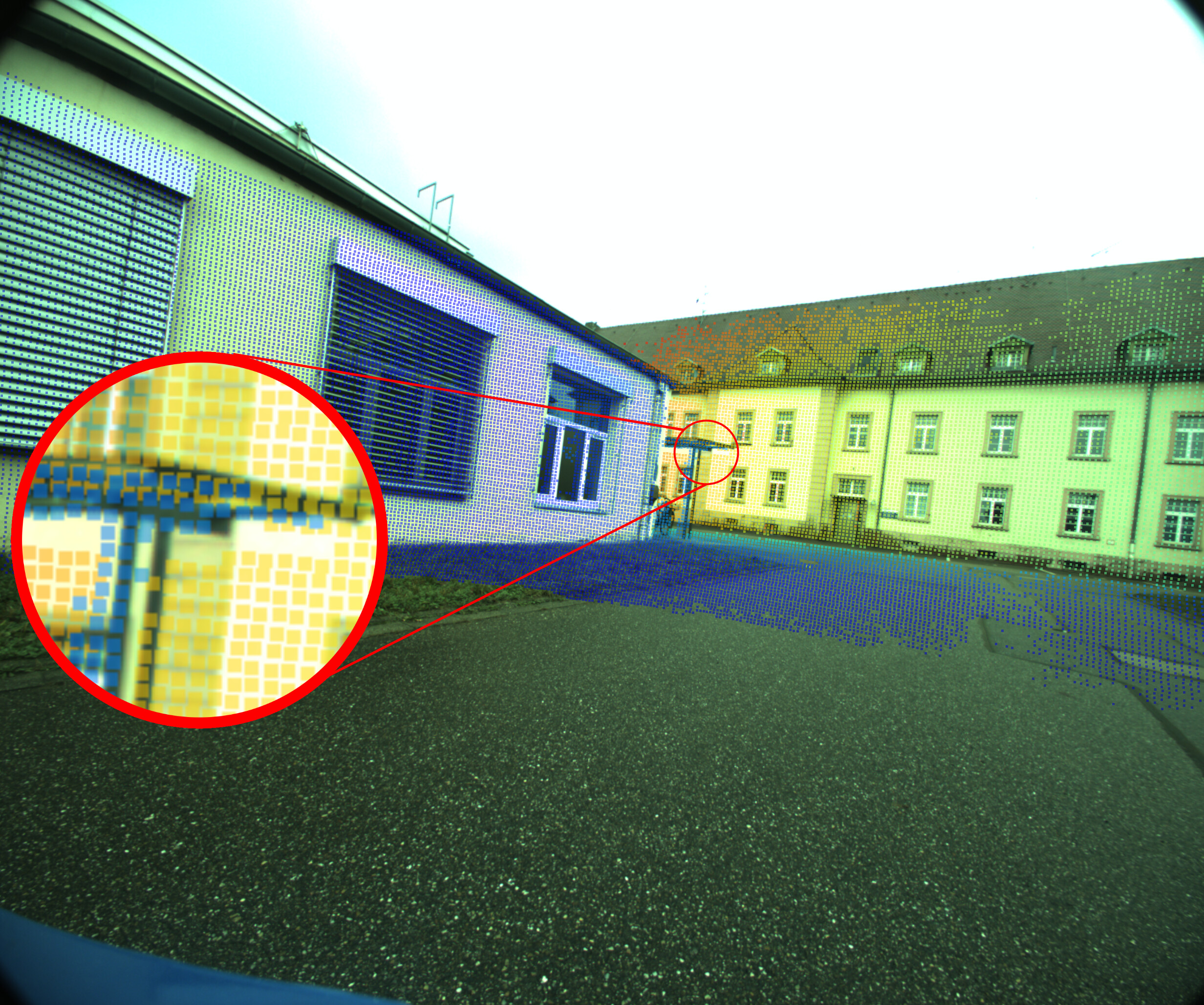} \\

      \includegraphics[width=\linewidth,trim={0 5cm 0 5cm},clip]{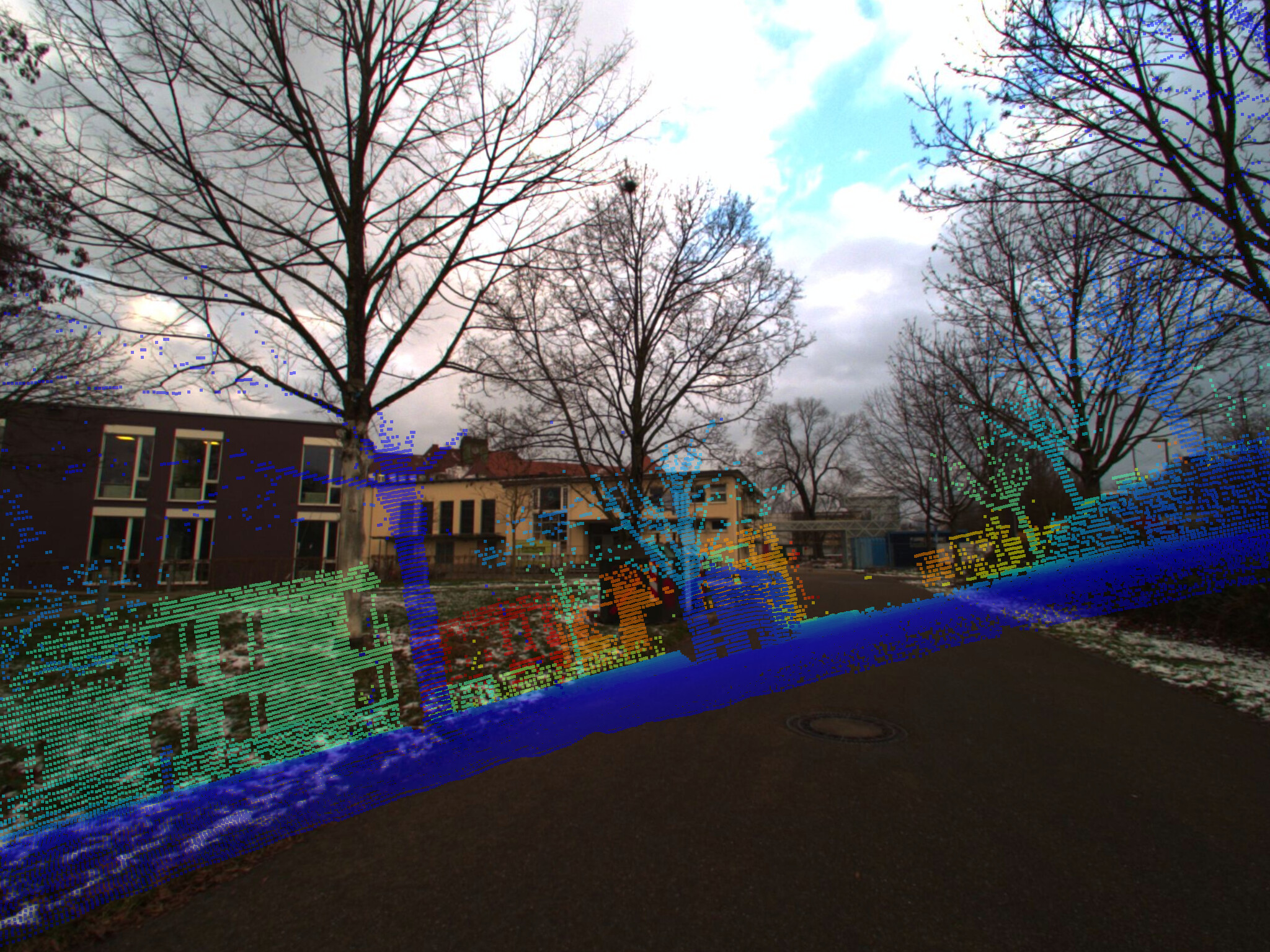} &
      \includegraphics[width=\linewidth,trim={0 5cm 0 5cm},clip]{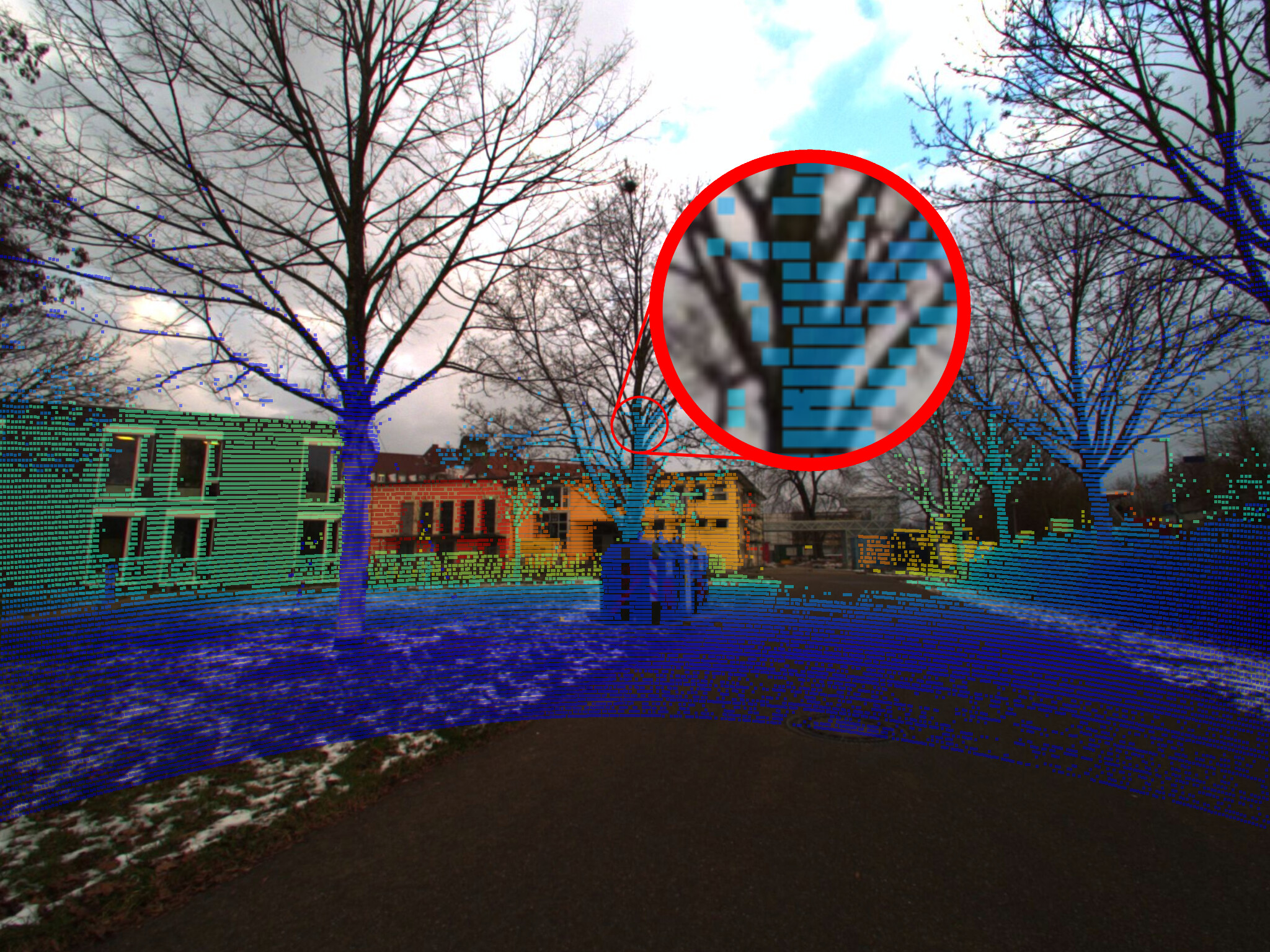} &
      \includegraphics[width=\linewidth,trim={0 5cm 0 5cm},clip]{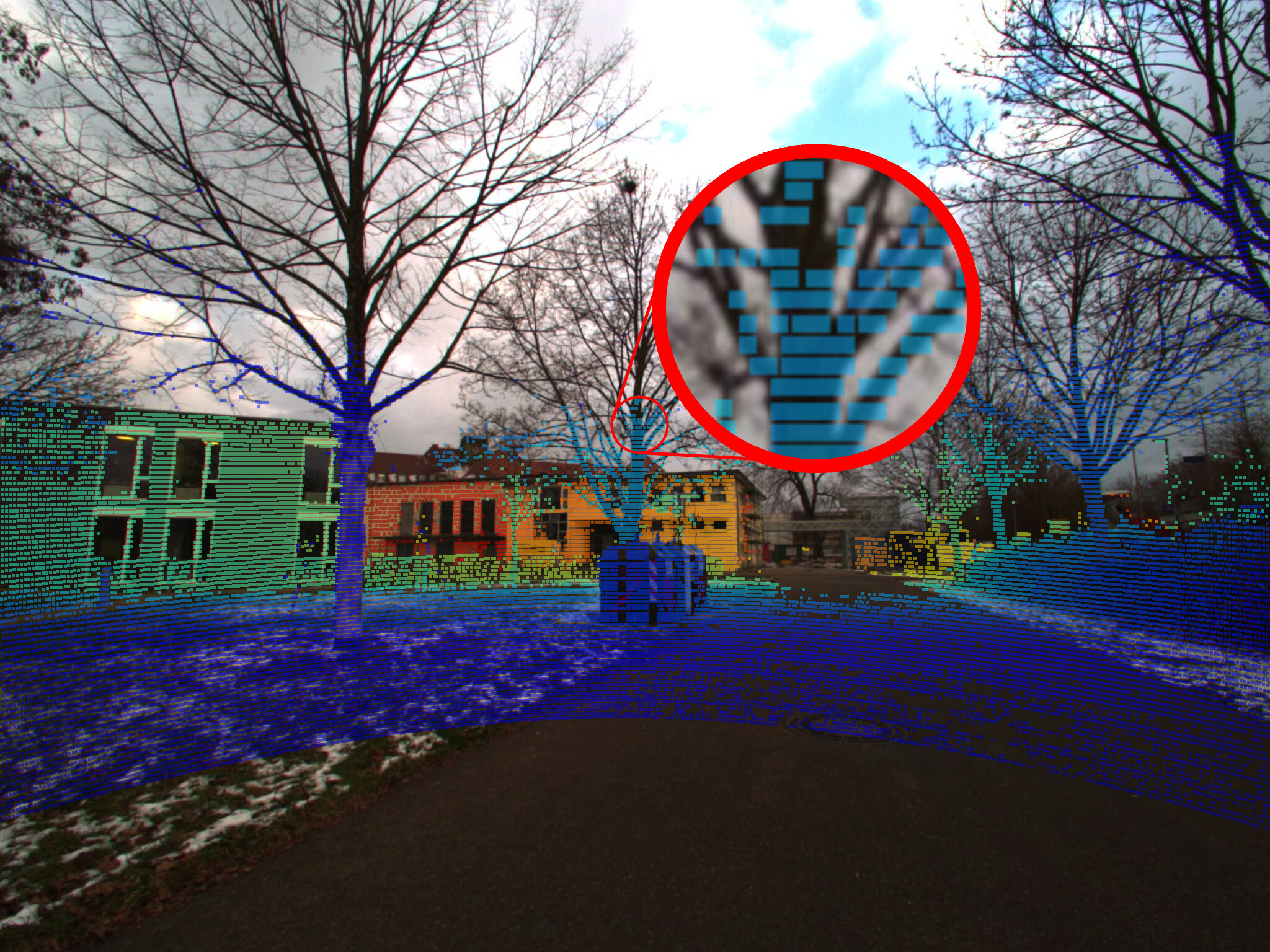} \\
    \end{tabular}}
      \caption{Qualitative results of \cmrnet2{} on the extrinsic calibration task. From left to right: LiDAR image projected in the initial pose, ground truth pose, and pose predicted by \cmrnet2{}. All LiDAR projections are overlaid with the respective RGB image for visualization purpose. From top to bottom: KITTI, Freiburg-Car, Freiburg-Quadruped, and Freiburg-UAV.}
      \label{fig:qualitativecalib}
      \vspace{0.2cm}
\end{figure*}

\subsection{Extrinsic Calibration --- Temporal Aggregation}

Since the extrinsic calibration of a specific sensor setup does not change over the course of a single sequence, we can further refine the calibration predicted by \cmrnet2{} by aggregating the predictions over multiple frames. Extrapolating a single rigid-body transformation from multiple 6-DoF poses is however a non-trivial task. In particular, we aggregate the translation and the rotation components separately, using three different central tendency measures: mean, median, and mode. While for the translation vector, we can use the standard component-wise measures, 3D rotations require special consideration. We compute the mean rotation using the average quaternion as defined in~\cite{markley2007averaging}, which is computed as the eigenvector $q_{mean}$ corresponding to the maximum eigenvalue of the matrix
\begin{equation}
  M = \frac{1}{n} \sum_i^n \mathbf{q}_i \mathbf{q}_i^T ,
\end{equation}
where $n$ is the number of frames used for aggregation and $\mathbf{q}_i$ is the quaternion corresponding to the $i$-th frame. $q_{mean}$ can be computed using the Singular Value Decomposition (SVD) of $M$. For the median rotation, we evaluated the method proposed in~\cite{aftab2014generalized}, which starts from an initial guess, set to the average quaternion defined above, and iteratively updates the guess until convergence. However, due to the iterative nature of the method, the runtime was too high (the evaluation was not complete even after 24 hours), and therefore, we did not include the median rotation in the results. Finally, the mode rotation does not require any special consideration, as the mode of a set of quaternions is simply the quaternion with the highest frequency in the set. For computing the mode of both translation and rotation, we first round all values to a given number of decimals, specifically two for translations and four for rotations.

\begin{table*}
  \centering
  \caption{Zero-shot generalization results using our three in-house platforms on the extrinsic calibration task.}
  \label{tab:calibration_inhouse}
  \begin{threeparttable}
    \begin{tabular}{lp{0.1cm}cccccc}
    \toprule
     && \multicolumn{2}{c}{Freiburg - Car} & \multicolumn{2}{c}{Freiburg - Quadruped} & \multicolumn{2}{c}{Freiburg - UAV} \\ \cmidrule(lr){3-4} \cmidrule(lr){5-6} \cmidrule(lr){7-8} 
    && Transl. [cm] & Rot. [°] & Transl. [cm] & Rot. [°] & Transl. [cm] & Rot. [°] \\ \midrule

    {\color{black}Koide~\textit{et~al.}~\cite{koide2023general} (Multi-frame optimization)} && {\color{black}9.76} & {\color{black}0.27} & {\color{black}16.21} & {\color{black}1.34} & {\color{black}\textbf{1.65}} & {\color{black}0.36} \\ 

    \textbf{\cmrnet2{}} (Single frame) && 12.20 & 0.90 & 23.95 & 1.36 & 12.47 & 0.97 \\
    \textbf{\cmrnet2{}} (Temporal Aggregation) && \textbf{1.89} & \textbf{0.19} & \textbf{9.70} & \textbf{0.21} & 1.84 & \textbf{0.23} \\
    
    \bottomrule
  \end{tabular}
   \end{threeparttable}
\end{table*}

The multi-frames results reported in~\cref{tab:calibration_temporal} show that the performance of all tendency measures decreases as the number of frames used for aggregation decreases, as hypothesized. Moreover, the mode consistently outperforms the mean measures in all experiments. The median aggregation achieves the best translation error on the left camera but decreases significantly on the right camera, where the mode achieves the best results. When using all frames of the testing sequence (4541 frames), the aggregated \textit{mode} extrinsic calibration achieves a translation error of \SI{0.75}{\centi\metre} and a rotation error of \ang{0.03} on the left camera, and \SI{1.98}{\centi\metre} and \ang{0.05} on the right camera, setting the new state of the art for targetless LiDAR-camera extrinsic calibration.

\subsection{Extrinsic Calibration --- Generalization}

We evaluate the generalization ability of \cmrnet2{} on the extrinsic calibration task on our three in-house platforms detailed in~\cref{sec:datasets}: a self-driving perception car, a quadruped robot, and a UAV.
To quantitatively evaluate our proposed method, we first calibrated the extrinsic parameters of each platform using a simple manual calibration procedure. In particular, we recorded a calibration sequence by moving a planar calibration board in front of the platform and then we manually selected the corners of the calibration board in both the camera images and the LiDAR scans. Finally, based on these correspondences, we used PnP with RANSAC, followed by a Levenberg-Marquardt optimization based on the inliers of RANSAC. We compare the translation and rotation errors of \cmrnet2{} on the three platforms against the manual calibration. In this experiment, since the number of frames in each dataset is relatively low (around 20 frames), we use 30 different random initial noise poses for each frame, and we report the median frame-wise errors and the error of the pose computed using temporal aggregation across all frames. {\color{black}We compare our approach with the optimization-based method proposed by Koide~\textit{et al.}~\cite{koide2023general}, which leverages temporal aggregation of point clouds and multi-frame optimization.} The results reported in~\cref{tab:calibration_inhouse} show that our method with temporal aggregation achieves rotation errors lower than \ang{0.23} on all three platforms and translation errors lower than \SI{2}{\centi\metre} on the car and UAV platforms, {\color{black}outperforming Koide~\textit{et al.}~\cite{koide2023general} on almost all metrics}. The translation error on the quadruped is considerably higher than those on the other platforms. However, this higher error might derive from an inaccurate manual calibration, as we have no way to guarantee the accuracy of our manual calibration procedure.

\begin{figure*}
  \centering
      \includegraphics[width=.43\textwidth]{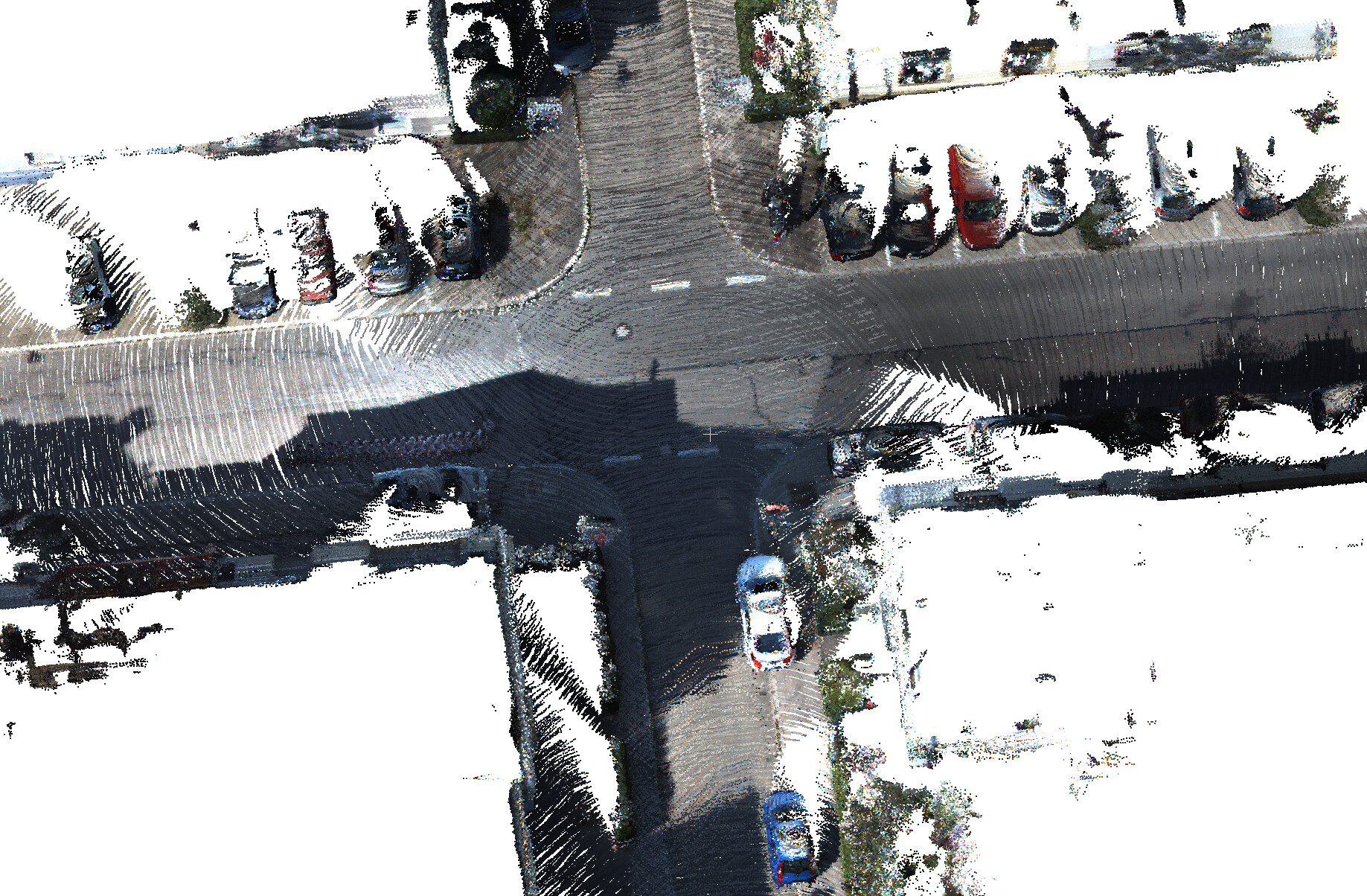}
      \includegraphics[width=.56\textwidth]{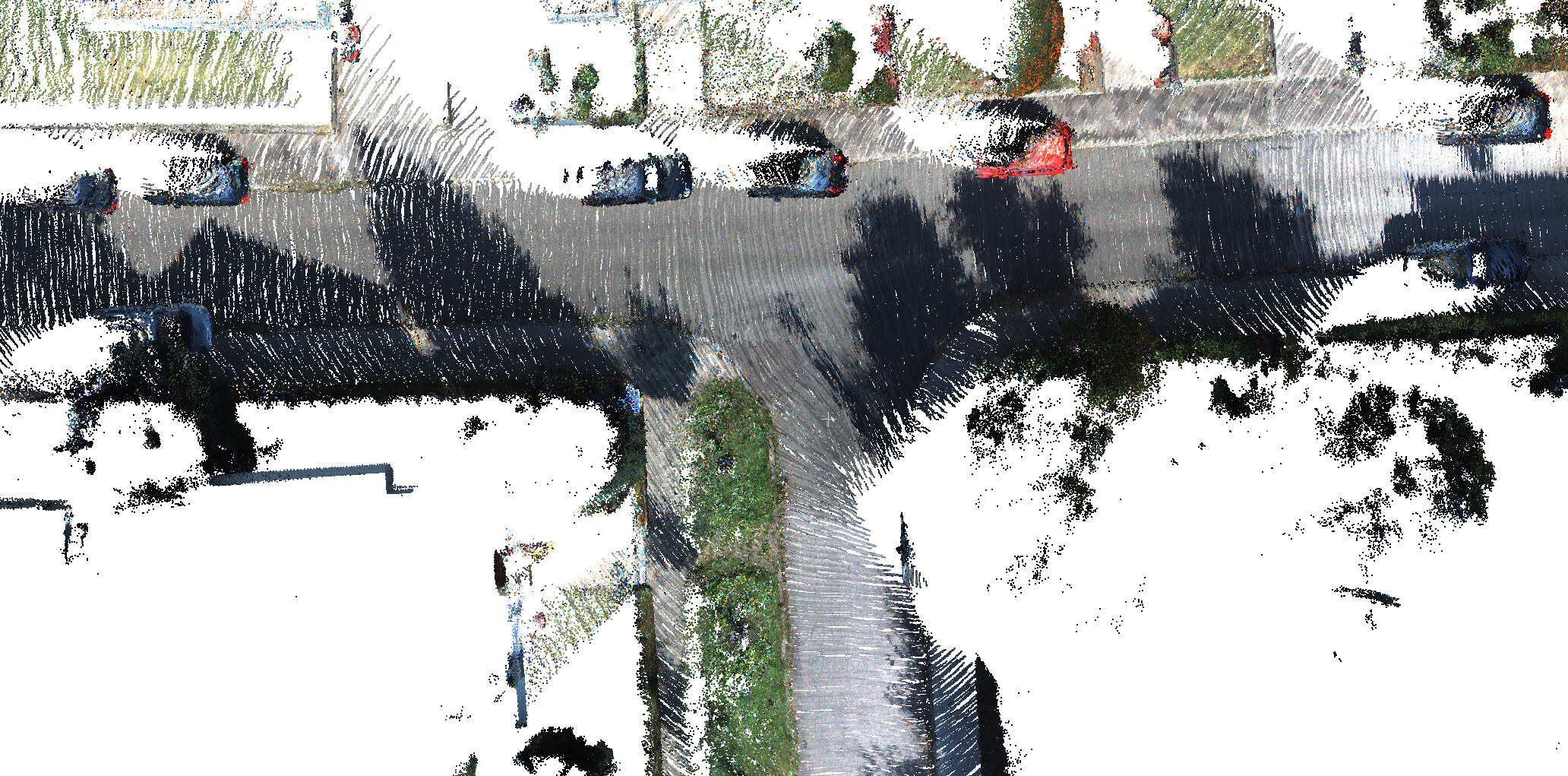}

      \vspace{0.1cm}

      \includegraphics[width=.53\textwidth]{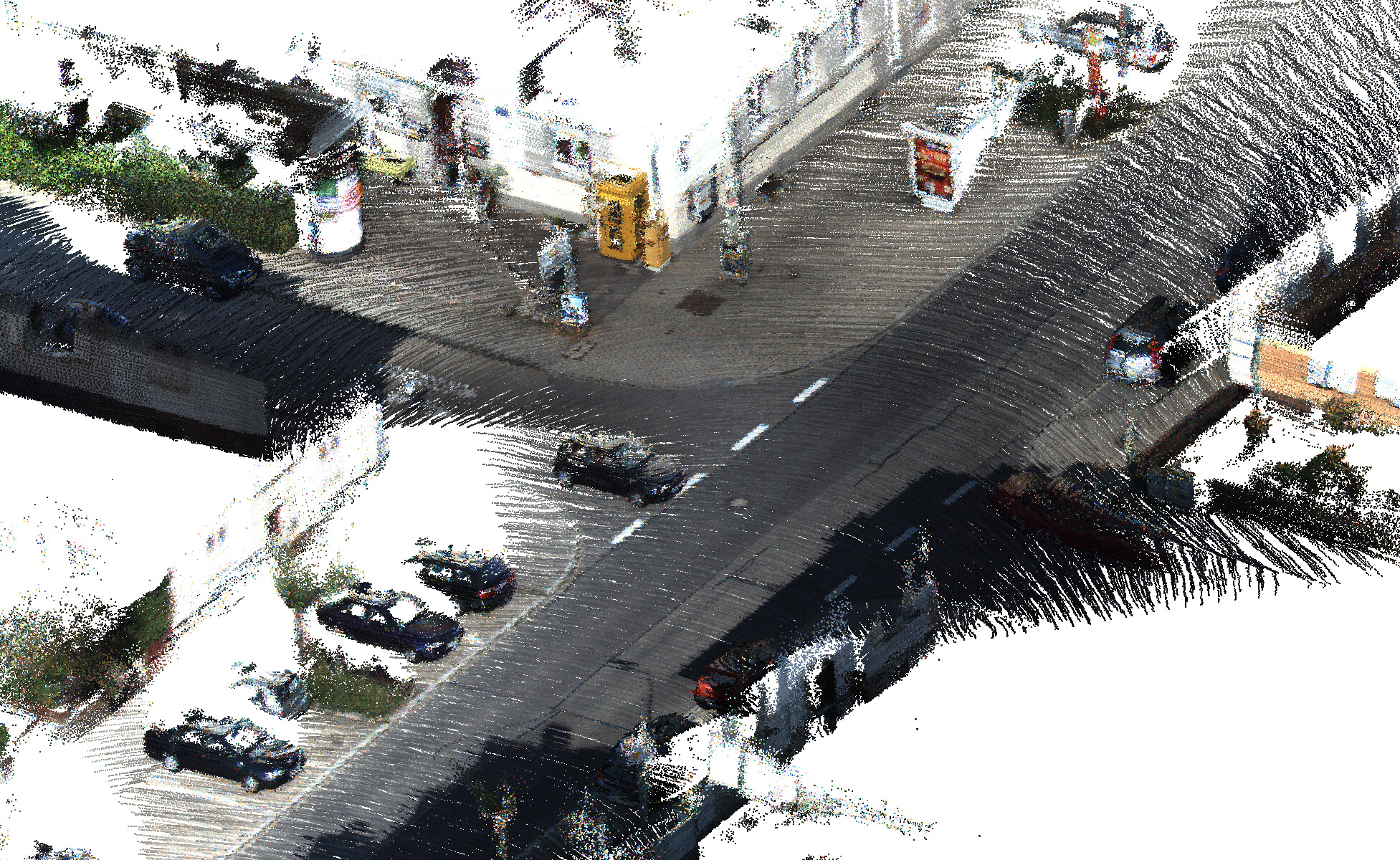}
      \includegraphics[width=.46\textwidth]{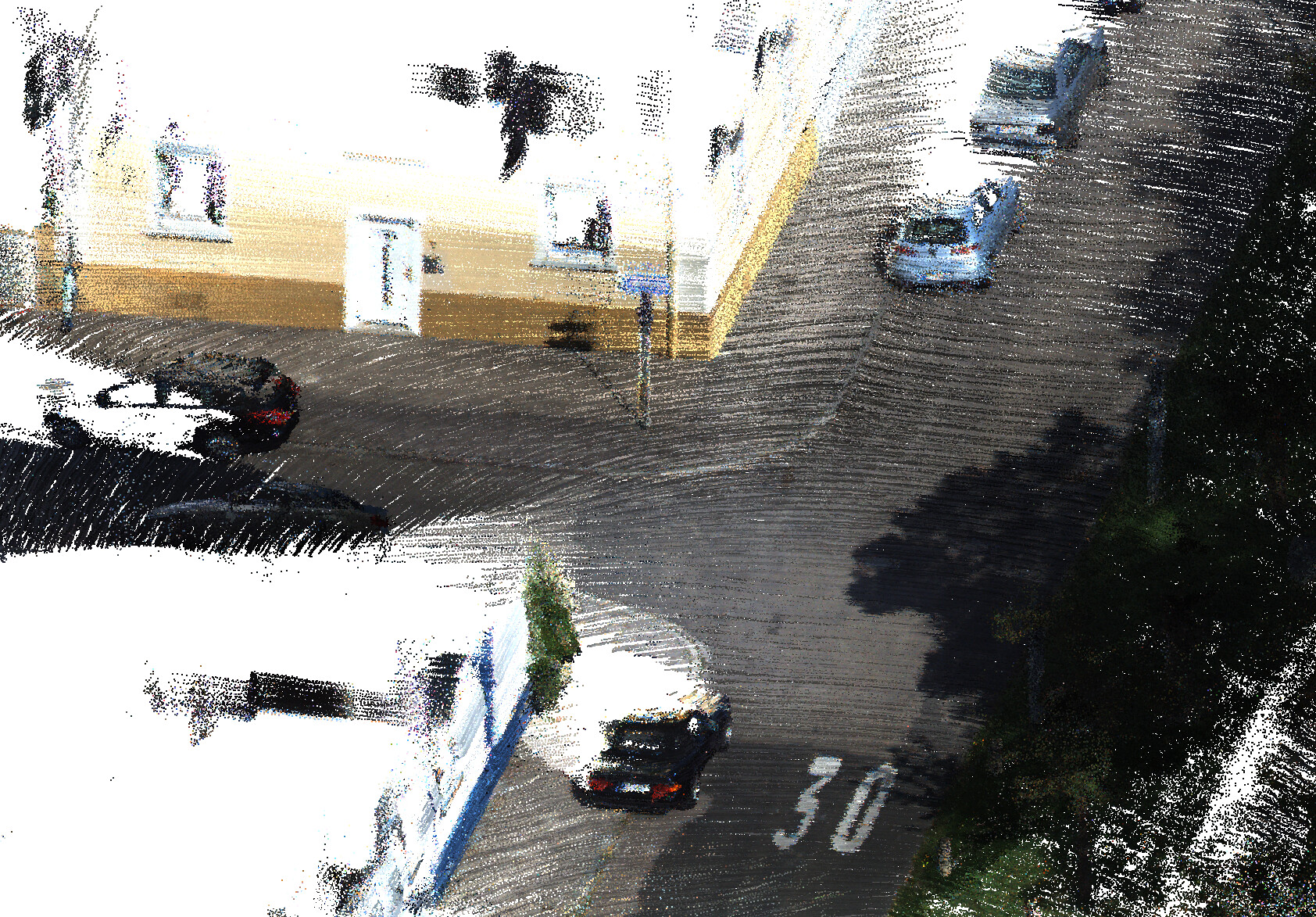}

      \caption{Four areas of the colorized LiDAR map obtained by combining the LiDAR scans and the right camera images of the KITTI dataset using the extrinsic parameters predicted by \cmrnet2{}.}
      \label{fig:color_map}
      \vspace{0.2cm}
\end{figure*}

In~\cref{fig:qualitativecalib}, we present qualitative results of \cmrnet2{} for LiDAR-camera extrinsic calibration on the right camera of the KITTI dataset and our three in-house platforms. Our method is able to accurately calibrate the extrinsic parameters of very different platforms without any retraining or fine-tuning. Additionally, we used the extrinsic parameters predicted by \cmrnet2{} to colorize the LiDAR scans of the KITTI dataset and combine them using the ground truth LiDAR poses to obtain a colorized LiDAR map. \cref{fig:color_map} shows four areas of the resulting LiDAR map colorized using the right camera of the sequence 00.

\subsection{Hyperparameter Optimization}
The default hyperparameter values used for all the experiments reported in previous sections were optimized using the hyperparameter optimization method BOHB~\cite{falkner2018bohb}, which combines Bayesian optimization and Hyperband. In particular, we used the implementation provided with the Weights\&Biases platform~\cite{wandb}. The hyperparameters optimized were the learning rate, the batch size, the number of Fourier frequencies used for the positional encoding, the parameters of the occlusion filter, the optimizer, the learning rate scheduler, and the maximum depth at which a point in the LiDAR image is kept. \cref{tab:hpo} shows the search space used for each hyperparameter and the optimal value found by the hyperparameter optimization.

\begin{table}
  \centering
  \caption{Hyperparameter optimization search space and results.}
  \label{tab:hpo}
  \begin{threeparttable}
    \begin{tabular}{lcc}
    \toprule
    Hyperparameter & Search Space & Optimal Value \\ \midrule

    Learning Rate & [1e-3, 5e-05] & 3e-4 \\
    Batch Size & [1, 8] & 4 \\
    Optimizer & SGD, Adam, AdamW & Adam \\

    Scheduler & MultiStep, OneCycle, Cyclic & OneCycle \\
    Fourier Freq. & [0, 15] & 12 \\
    Filter Size $K_{occ}$ & [5, 18] & 9 \\
    Max Depth & [50, 200] & 160 \\
    
    \bottomrule
  \end{tabular}
   \end{threeparttable}
\end{table}

\subsection{Runtime Analysis}

In this experiment, we compare the runtime of \cmrnet2{} with existing camera-agnostic methods CMRNet++~\cite{cattaneo2020cmrnet} and I2D-Loc~\cite{chen2022i2d}. Additionally, we compare the performance and runtime of different algorithms to solve the PnP problem, most of which are implemented in the OpenCV library~\cite{opencv_library}, including EPnP~\cite{lepetit2009epnp}, SQPnP~\cite{terzakis2020consistently}, and our GPU-based implementation of EPnP + RANSAC. All runtimes were computed only for the first iteration on a workstation with 32GB DDR3 RAM, an Intel i7-4790K CPU, and an NVIDIA RTX 2080 Ti GPU. For the iterative refinement approach described in~\cref{sec:iterative}, the runtime scales linearly with the number of iterations. The results reported in~\cref{tab:runtime} show that \cmrnet2{} with the GPU-based PnP solver achieves the fastest runtime among all methods. Moreover, it is possible to further improve the performance of \cmrnet2{} by performing a Levenberg–Marquardt (LM) optimization on the inliers found by RANSAC at the cost of increased runtime.

\begin{table}
  \centering
  \caption{Runtime analysis on the KITTI dataset.}
  \label{tab:runtime}
  \begin{threeparttable}
    \begin{tabular}{clccc}
    \toprule
    &Pose Estimator & Transl. & Rot. & Runtime [s]\\ \midrule

    &I2D-Loc & 18 & 0.70 & 0.326 \\
    &CMRNet++ & 44 & 1.13 & 1.300 \\
    \midrule
    \multirow{4}{*}{\rotatebox[origin=c]{90}{\parbox[c]{1cm}{\centering\scriptsize \cmrnet2{}}}} &SQPnP + RANSAC & 10 & 0.28 & 0.800 \\
    &EPnP + RANSAC (C) & 10 & 0.29 & 0.957 \\
    &EPnP + RANSAC (G) & 10 & 0.29 & \textbf{0.200} \\
    &EPnP + RANSAC (G) + LM  & \textbf{9} & \textbf{0.27} & 0.629 \\
    
    \bottomrule
  \end{tabular}
  \begin{tablenotes}[para,flushleft]
       \footnotesize
       (C) and (G) represent the CPU- and GPU-based implementation, respectively. LM represents an additional Levenberg–Marquardt optimization on the inliers found by RANSAC. All runtimes are computed for a single iteration.
     \end{tablenotes}
   \end{threeparttable}
\end{table}

\section{\color{black}Conclusions and Limitations}\label{sec:conclusion}
In this paper, we presented \cmrnet2{} for monocular localization in LiDAR-maps and extrinsic LiDAR-camera calibration in the wild. \cmrnet2{} decomposes the LiDAR-camera relative pose estimation problem in two steps. First, an optical flow estimation network predicts dense point-pixel correspondences between the camera image and the point cloud. Subsequently, we use these matches to predict the relative pose between the camera and the LiDAR using robust geometric techniques.
{
\color{black}CMRNext has three primary limitations. First, since it is trained exclusively on autonomous driving datasets, its generalization ability is likely to be limited in other environments, such as off-road or mining environments. Second, while our method is faster than existing camera-agnostic approaches, it is not yet capable of real-time performance. Lastly, similar to most existing methods, the training paradigm requires ground truth poses as well as correct intrinsic camera parameters, which are typically provided in publicly available datasets, but add an additional constraint to the approach.
}
Through extensive experimental evaluations, we demonstrated that \cmrnet2{} sets the new state-of-the-art on both monocular localization in LiDAR-maps and extrinsic calibration tasks. Remarkably, \cmrnet2{} can effectively generalize without any retraining or fine-tuning to different environments and different robotic platforms, including three in-house robots with diverse locomotion: a self-driving perception car, a quadruped robot, and a quadcopter UAV. \cmrnet2{} achieves median translation and rotation errors of \SI{6.21}{\centi\meter} and \ang{0.23}, respectively, for the localization task on the sequence 00 of the KITTI odometry dataset. Additionally, \cmrnet2{} calibrated the LiDAR-camera extrinsic parameters of our in-house platforms with translation and rotation errors as low as \SI{1.84}{\centi\meter} and \ang{0.19}, respectively, demonstrating exceptional zero-shot generalization ability. To foster research in this direction, we have made the code and pre-trained models publicly available at \url{http://cmrnext.cs.uni-freiburg.de}.

\bibliographystyle{IEEEtran}
\bibliography{IEEEabrv,bibliography}

\begin{IEEEbiography}[{\includegraphics[width=1in,height=1.25in,trim={2cm 0 2cm 0},clip,keepaspectratio]{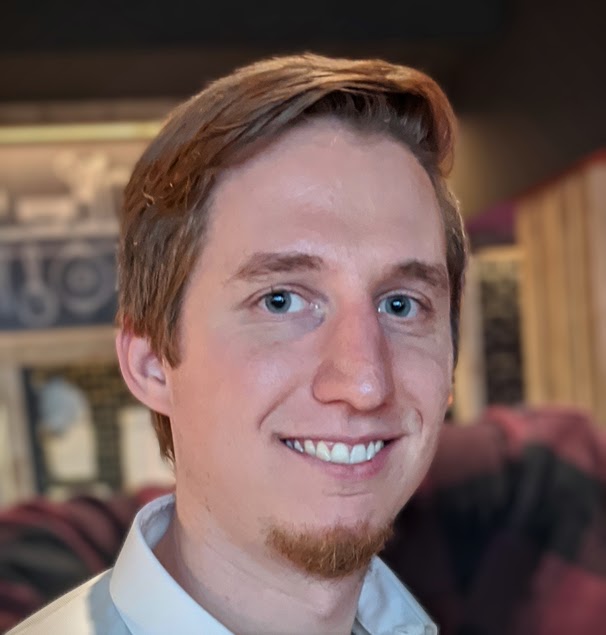}}]{Daniele Cattaneo} received the M.Sc.~degree in Computer Science from the University of Milano-Bicocca, Milan, Italy, in 2016 and the Ph.D.~degree in Computer Science from the same university in 2020.
He is currently a Junior Research Group Leader at the Robot Learning Lab of the University of Freiburg, Germany.
His research interest includes deep learning for robotic perception and localization, with a focus on label-efficient learning, cross-modal matching, domain generalization, sensor fusion, and combining learning-based methods with established geometric and robotic techniques.
\end{IEEEbiography}

\begin{IEEEbiography}[{\includegraphics[width=1in,height=1.25in,clip,keepaspectratio]{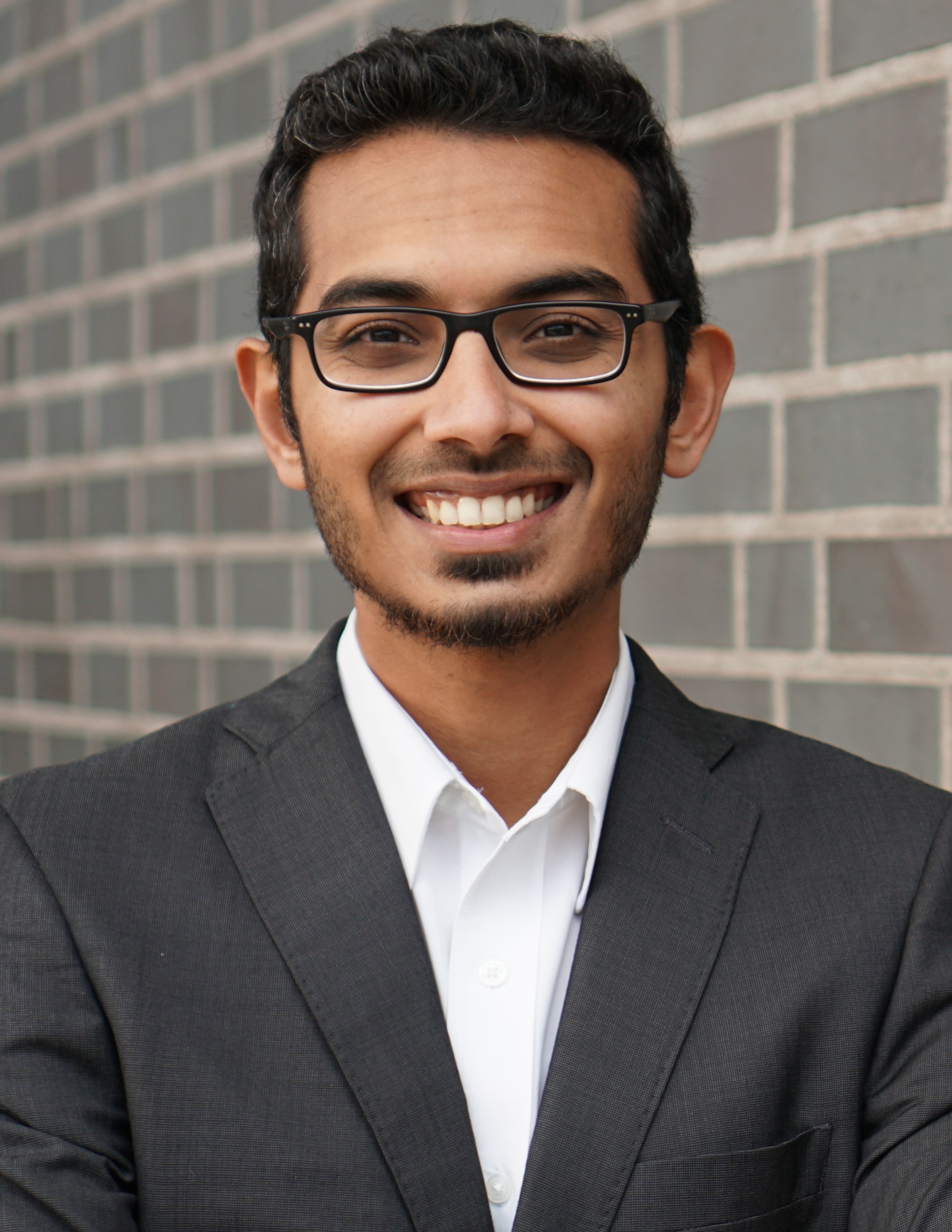}}]{Abhinav Valada}
is a full Professor and Director of the Robot Learning Lab at the University of Freiburg, Germany. He is a member of the Department of Computer Science, a principal investigator at the BrainLinks-BrainTools Center, and a founding faculty of the European Laboratory for Learning and Intelligent Systems (ELLIS) unit at Freiburg. He received his Ph.D.~in Computer Science from the University of Freiburg in 2019 and his M.S.~degree in Robotics from Carnegie Mellon University in 2013. His research lies at the intersection of robotics, machine learning, and computer vision with a focus on tackling fundamental robot perception, state estimation, and planning problems using learning approaches in order to enable robots to reliably operate in complex and diverse domains. Abhinav Valada is a Scholar of the ELLIS Society, a DFG Emmy Noether Fellow, and co-chair of the IEEE RAS TC on Robot Learning.
\end{IEEEbiography}

\end{document}